\documentclass{article}

\PassOptionsToPackage{numbers,compress}{natbib}

\usepackage[main,final]{neurips_2026}

\usepackage[utf8]{inputenc}
\usepackage[T1]{fontenc}
\usepackage{hyperref}
\usepackage{url}
\usepackage{booktabs}
\usepackage{amsfonts}
\usepackage{nicefrac}
\usepackage{microtype}
\usepackage[table]{xcolor}
\usepackage{graphicx}
\usepackage{multirow}
\usepackage{makecell}
\usepackage{bm}
\usepackage{amsmath,amssymb,mathtools}
\usepackage{subcaption}
\usepackage{pgfplots}
\usepgfplotslibrary{groupplots,statistics}
\pgfplotsset{compat=1.18}
\usepackage{etoolbox}
\usepackage{tikz}
\usetikzlibrary{positioning,calc,fit,arrows.meta,decorations.pathreplacing,backgrounds}

\definecolor{famA}{gray}{0.85}
\definecolor{famB}{gray}{0.75}
\definecolor{famC}{gray}{0.65}

\newcommand{\mstd}[2]{%
  #1%
  \ifstrequal{#2}{-}{}{%
    \,\scalebox{0.6}{$\pm$#2}%
  }%
}

\title{GATE-AD: Graph Attention Network Encoding for Few-Shot Industrial Visual Anomaly Detection}

\author{
  Aggelos Psiris \\
  Harokopio University of Athens \\
  Athens, Greece \\
  \texttt{aggelospsiris@hua.gr}
  \And
  Yannis Panagakis \\
  National and Kapodistrian University of Athens \\
  Archimedes, Athena Research Center \\
  Athens, Greece \\
  \texttt{yannisp@di.uoa.gr}
  \AND
  Maria Vakalopoulou \\
  CentraleSupelec, University Paris Saclay \\
  Archimedes, Athena Research Center \\
  Paris, France / Athens, Greece \\
  \texttt{maria.vakalopoulou@centralesupelec.fr}
  \And
  Georgios Th. Papadopoulos \\
  Harokopio University of Athens \\
  Archimedes, Athena Research Center \\
  Athens, Greece \\
  \texttt{g.th.papadopoulos@hua.gr}
}

\begin{document}

\maketitle

\begin{abstract}
Few-Shot Industrial Visual Anomaly Detection (FS-IVAD) is a critical task in modern manufacturing, where automated product inspection systems must identify rare defects using only a handful of normal, defect-free training samples. Existing approaches typically model normality by assessing each patch largely through its individual appearance, e.g., through similarity to stored normal features, reconstruction from support samples, or distance from a learned normal representation. This paper introduces GATE-AD, whose key contribution is to instead model few-shot normality through local contextual relations: a patch is considered normal when it can be consistently recovered from its learned dependencies with its spatial neighbors. To realize this formulation, frozen self-supervised ViT patch tokens are organized on a spatial neighbor graph and processed by a Graph Attention Network (GAT), while representation alignment and a Scaled Cosine Error (SCE) objective enable reconstruction-based anomaly scoring. Node masking and attention dropout further regularize learning in the low-shot regime. On MVTec AD, VisA, and MPDD, GATE-AD achieves strong detection and localization performance across $1$- to $8$-shot settings, attaining the highest image-level AUROC in $10$ of $12$ settings and the highest image-level AP in $8$ of $12$ settings under a unified evaluation protocol, while also achieving the lowest inference latency among the compared methods.  In order to facilitate reproducibility and further research, the source code of GATE-AD is available at https://github.com/gthpapadopoulos/GATE-AD.
\end{abstract}

\section{Introduction}
\label{sec:intro}

Industrial Visual Anomaly Detection (IVAD) comprises a cornerstone of modern quality assurance pipelines, enabling the automated identification of defects (such as scratches, dents, and structural failures) in manufactured products \citep{li2025survey,liu2024deep}. In practice, production lines change frequently and defective samples are rare by nature, making the collection of large labeled datasets prohibitively expensive. This motivates the \textbf{Few-Shot Industrial Visual Anomaly Detection (FS-IVAD)} setting, where a model must detect and localize anomalies using only a small number, typically 1 to 8, of defect-free reference images per product category, with no abnormal samples available during training~\citep{Cui2023,jeong2023winclip}.

The core challenge in FS-IVAD is learning a compact, robust representation of \emph{normality} from very limited data, such that deviations from it can be reliably identified at inference. This is challenging due to: a) \textbf{Data scarcity}, where normality must be captured from as few as a single image, b) \textbf{Subtle variance}, where genuine structural failures must be distinguished from natural appearance variation, and c) \textbf{Domain shift}, where the learned appearance patterns of normality must generalize across unseen product defects \citep{jeong2023winclip,damm2025anomalydino,fang2023fastrecon}.

Existing approaches fall broadly into three main categories: a) \emph{Memory bank methods} \citep{cohen2020sub,roth2022towards,damm2025anomalydino,xie2023pushing} retrieve anomaly scores by comparing test patches against a stored bank of normal features, but their memory requirements and nearest-neighbor search cost generally increase with the size of the stored support representation; b) \emph{Vision-language methods}~\citep{jeong2023winclip,li2024promptad,lv2025oneforall,chen2023april} leverage vision-text representations (e.g., CLIP-based) to align visual patches with textual descriptions of normality, but suffer from a granularity gap, since Vision-Language Models (VLMs) are trained on image-level supervision while industrial defects are fine-grained and local; c) \emph{Reconstruction-based methods}~\citep{fang2023fastrecon,li2024one,tian2025fastref} train a model to reconstruct normal patterns and score anomalies by residual error, but risk the identity-mapping problem and tend to overfit to the limited normal samples available. The current work argues that the key limitation shared across all FS-IVAD categories is the failure to explicitly model \textbf{local relational structure} among image patches, i.e., the contextual consistency between a patch and its spatial neighbors: in defect-free images, each patch is statistically predictable from its immediate surroundings, while industrial defects (e.g., scratches, cracks, contamination, missing parts, etc.) disrupt precisely this neighbor-to-neighbor consistency, often without modifying the global image statistics. 


To address the key challenge mentioned above, a novel approach to reconstruction-based FS-IVAD is proposed, termed \textbf{GATE-AD}. The framework casts few-shot normality modeling as a masked, representation-aligned relational reconstruction problem over frozen self-supervised ViT patch tokens, extracted using DINOv$2$~\citep{oquab2023dinov2}/ DINOv$3$~\citep{simeoni2025dinov3}. A fundamental consideration behind the proposed design is that learning a relational normality prior requires each patch representation to be inferred from its local spatial context, while preserving the rich semantic representation of the frozen backbone. To this end, the patch tokens are organized on a fixed spatial neighbor graph, and the individual components of GATE-AD serve complementary roles within this formulation: a) the GAT encoder performs feature-dependent message passing over the predefined spatial neighborhood, assigning non-uniform importance to neighboring patches and thereby modeling their local dependencies~\citep{velivckovic2017graph}; b) GraphMAE-style node masking~\citep{hou2022graphmae} prevents trivial copying of the input features and forces the GAT encoder to infer masked patch representations from their surrounding context; c) the representation-alignment head provides the shared latent space in which the graph-conditioned prediction is compared with the corresponding frozen ViT target, thereby providing the reconstruction signal used to train the GAT encoder; and d) the Scaled Cosine Error (SCE) down-weights well-reconstructed patches and places greater emphasis on harder reconstruction errors, while attention dropout regularizes the learned neighborhood dependencies in the low-shot regime. At inference, patches that cannot be consistently reconstructed from their spatial context produce high SCE reconstruction errors and are identified as anomalous. The key contributions of GATE-AD are:

\begin{itemize}
    \item GATE-AD is, to the best of the authors' knowledge, the first FS-IVAD framework to \textbf{reformulate normality as a relational reconstruction problem}, where a patch is considered normal if its representation can be recovered from its learned dependencies with the spatial neighborhood, rather than solely from its similarity to stored normal features.

    \item To realize this relational formulation, GATE-AD introduces a \textbf{masked graph reconstruction scheme} that learns normal spatial dependencies by predicting patch representations from their local neighborhood, with representation alignment providing a shared latent space between the graph-conditioned prediction and the corresponding frozen DINO target.

    \item Extensive evaluation on MVTec AD, VisA, and MPDD demonstrates \textbf{strong and consistent performance across $1$ to $8$ shot settings}, with GATE-AD achieving the highest image-level AUROC in $10/12$ settings and the highest image-level AP in $8/12$ settings under a unified evaluation protocol, while also attaining the lowest inference latency, $22.5\%$ lower than the next fastest compared method.
\end{itemize}

The remainder of the paper is organized as follows: Section \ref{sec:related} outlines the relevant FS-IVAD literature; Section \ref{sec:method} describes the proposed GATE-AD approach; Section \ref{sec:experiments} provides experimental evaluation results and comparisons with literature methods; Section \ref{sec:conclusion} concludes the paper.

\section{Related Work}
\label{sec:related}

\paragraph{Visual feature embeddings and memory banks.}
Feature embedding-based methods rely on mapping an input image to a high-dimensional feature space, where anomalous samples correspond to low-density areas or lie far from normal ones. PatchCore \citep{roth2022towards} establishes patch-level memory-bank matching, AnomalyDINO \citep{damm2025anomalydino} strengthens it with DINOv2 features and foreground masking, and VisionAD \citep{wang2025search} further enriches the normal memory through augmentation and multi-layer features. DFM \citep{wu2025dfm} jointly optimizes the feature extractor and matching head, while SubspaceAD \citep{lendering2026subspacead} models normal variation through a PCA subspace. Other approaches use graph representations to enhance feature modeling or matching \citep{xie2023pushing,gu2025univad,tao2025kernel}. Common challenges include modeling complex intra-class variation from very few normal samples, memory-dependent inference costs for retrieval-based methods, and sensitivity to spatial misalignment.

\paragraph{Vision-language model reasoning and adaptation.}
Vision-Language Models (VLMs) support linguistic descriptions that often encapsulate high-level priors for modeling abnormalities. WinCLIP+ \citep{jeong2023winclip} combines state words and prompt templates with multi-scale features. FADE \citep{li2024fade} adapts CLIP to extract multi-scale, image-patch embeddings better aligned with language. PromptAD \citep{li2024promptad} employs a dual-branch vision-language decoder for both normality and abnormality modeling. Lv et al. \citep{lv2025oneforall} learn a class-shared prompt generator. InCTRL \citep{zhu2024toward} learns an in-context residual learning model, while AnomalyR1 \citep{chao2025anomalyr1} autonomously processes image and domain-knowledge inputs. Despite ``open-vocabulary'' capabilities, VLM-based approaches can suffer from the vision-language granularity gap, limited domain specificity, and sensitivity to prompt design.

\paragraph{Reconstruction-based frameworks and prototype refinement.}
Reconstruction-based approaches detect anomalies through discrepancies between query representations and their estimated normal counterparts. FastRecon \citep{fang2023fastrecon} employs a regression scheme with distribution regularization. FastRef \citep{tian2025fastref} relies on characteristic transfer from query features to prototypes and anomaly suppression through prototype alignment. One-to-Normal \citep{li2024one} performs a personalized one-to-normal transformation of query images. FoundAD \citep{zhai2025foundation} learns a non-linear projection operator on the natural image manifold. Key limitations include the risk of reconstructing both normal and abnormal samples equally well, loss of reconstruction detail (often) larger than the defects themselves, tendency to overfit to normal samples, and reduced ability to handle high intra-class variance.

\section{Proposed approach}
\label{sec:method}

\subsection{Overview}
\label{subsec:overview}

\begin{figure}[t!]
  \centering
  \includegraphics[width=0.95\textwidth]{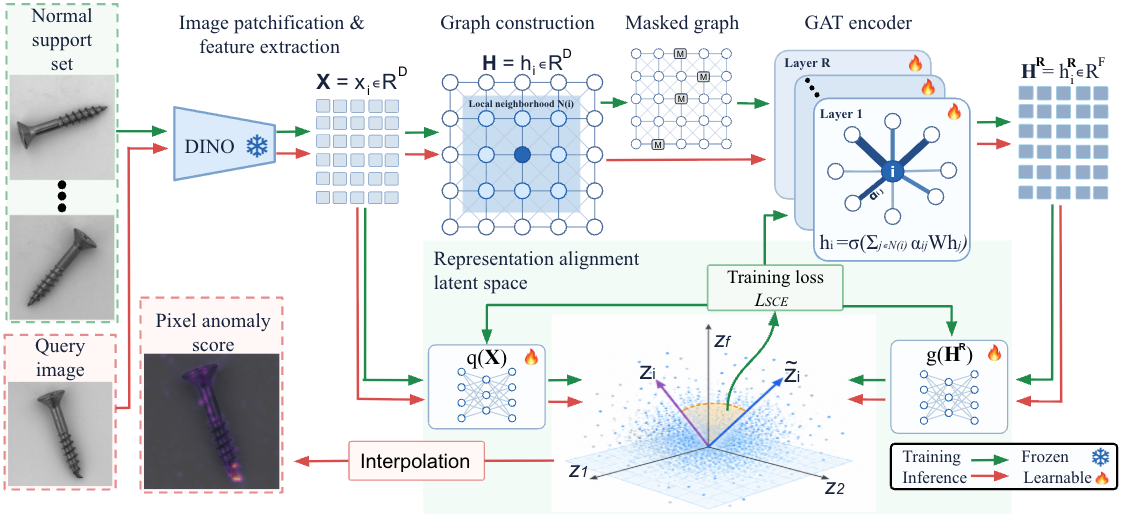}
  \caption{General architecture of the proposed masked, representation-aligned GAT encoding scheme for FS-IVAD.}
  \label{fig:gatead_overview}
\end{figure}

The proposed GATE-AD framework models few-shot normality through masked, representation-aligned graph reconstruction. Normal patches are expected to be predictable from their local spatial context, while anomalies violate these dependencies. A masked GAT encoder models such relations, and representation alignment provides the latent space where graph-conditioned predictions are compared with the corresponding frozen ViT features. The general architecture of GATE-AD is illustrated in Fig.~\ref{fig:gatead_overview}, while its individual components are detailed in the followings.

\subsection{Image patchification and feature extraction}
\label{subsec:feat}
Given an input image $I_{mg} \in \mathbb{R}^{H \times W \times 3}$, it is resized so that the smaller dimension equals $\beta$ pixels. A dense local feature token $\mathbf{x}_i \in \mathbb{R}^D$, $i \in [1,N]$, is then extracted for each $16{\times}16$ / $14{\times}14$ pixel image patch (standard for DINOv3/DINOv2) using a frozen ViT backbone (DINOv3 \citep{simeoni2025dinov3}/DINOv2 \citep{oquab2023dinov2}). Following common practices, the $\mathbf{x}_i$ representation is obtained by averaging the patch token across the last $\zeta$ transformer layers \citep{damm2025anomalydino, zhai2025foundation}.

\subsection{Masked graph attention network encoder}
\label{subsec:GAT}
\paragraph{Encoder type selection.}
A key component of reconstruction-based FS-IVAD is the encoder used to model normal appearance patterns. In GATE-AD, a GAT encoder \citep{velivckovic2017graph} operates on the patch-level representations $\mathbf{X}=\{\mathbf{x}_i\}_{i=1}^{N}$ and models their local dependencies. This choice is motivated by the observation that anomalous patches may violate the consistency between local features and their surrounding context \citep{bae2023pni}. In contrast to standard graph convolution~\citep{kipf2016semi}, GAT assigns feature-dependent weights to neighboring nodes and therefore performs selective local aggregation. Transformer attention~\citep{vaswani2017attention} provides a similar adaptive mechanism over global pairwise interactions, with quadratic complexity in the number of patches, whereas GAT restricts message passing to the fixed sparse local graph, yielding linear complexity for a fixed neighborhood size and a favorable accuracy-efficiency trade-off for FS-IVAD.


\paragraph{Encoder network architecture.}
The GAT encoder takes as input the patch-level representation $\mathbf{X}$, with each token $\mathbf{x}_i\in\mathbb{R}^D$ defining a node at the corresponding location of the ViT patch grid. Graph edges are determined solely by spatial proximity, defining a local neighborhood $\mathcal{N}(i)$ of size $K$ for each node, whose effect is examined in Appendix~\ref{subsec:graph_neighborhood}; no feature-space neighbor search is performed. Setting $\mathbf{h}_i^0=\mathbf{x}_i$, each GAT layer aggregates information over $\mathcal{N}(i)$ and updates the node representation. Consequently, after $r$ layers, $\mathbf{h}_i^r$ depends on the information contained in the $r$-hop spatial neighborhood of node $i$~\citep{velivckovic2017graph,nikolentzos2020k}.

\paragraph{Graph attentional layer.}
The input to a single attention layer is the set of node features $\mathbf{H}=\{\mathbf{h}_i\}_{i=1}^{N}$, $\mathbf{h}_i\in\mathbb{R}^F$ ($\mathbf{h}_i\equiv\mathbf{x}_i$ at the first layer), and the output is a new set $\mathbf{H'}=\{\mathbf{h}'_i\}_{i=1}^{N}$, $\mathbf{h}'_i\in\mathbb{R}^{F'}$. A shared learnable linear transformation $\mathbf{W}\in\mathbb{R}^{F'\times F}$ is applied to every node, while a shared self-attention mechanism $\alpha:\mathbb{R}^{F'}\times\mathbb{R}^{F'}\to\mathbb{R}$ is applied to neighboring nodes:
\begin{equation}
e_{ij} = \alpha (\mathbf{W}\mathbf{h}_i , \mathbf{W}\mathbf{h}_j ), ~j \in \mathcal{N}(i),
\label{eq:attention_coefficients}
\end{equation}
which indicates the importance of node $j$'s features to node $i$. Following common practice, $\alpha$ is implemented as a single-layer feedforward network parametrized by $\mathbf{a}\in\mathbb{R}^{2F'}$ with a LeakyReLU non-linearity:
\begin{equation}
\alpha_{ij} = \frac{\exp\big( \mathrm{LeakyReLU} \big(\mathbf{a}^T [\mathbf{W}\mathbf{h}_i \|\ \mathbf{W}\mathbf{h}_j]\big)\big)}
{\sum_{m\in \mathcal{N}(i)} \exp\big( \mathrm{LeakyReLU} \big(\mathbf{a}^T [\mathbf{W}\mathbf{h}_i \|\ \mathbf{W}\mathbf{h}_m]\big)\big)},
\label{eq:gat_attention_coefficients}
\end{equation}
where $\|$ denotes vector concatenation. The normalized coefficients $\alpha_{ij}$ are then used to compute a non-linear combination of the neighborhood features that yields the output $\mathbf{h}'_i = \sigma \big(\sum_{j \in \mathcal{N}(i)} \alpha_{ij} \mathbf{W}\mathbf{h}_j \big)$, where $\sigma(\cdot)$ is the ELU activation.

\paragraph{Representation alignment.}
Repeated message passing in GNNs may lead to over-smoothing, whereby node representations become increasingly similar \citep{khoshraftar2024survey}. To limit this effect and retain consistency with the original visual representation, a Representation Alignment (RA) module is introduced. In particular, the output of the last GAT layer $\mathbf{H}^R$ is aligned with the corresponding frozen ViT features $\mathbf{X}$ in a common latent space $\mathbf{Z}\in\mathbb{R}^f$ ($f\leq F$). The two representations are projected through lightweight learnable transformations:
\begin{equation}
\mathbf{z}_i = q(\mathbf{x}_i),~
\tilde{\mathbf{z}}_i = g(\mathbf{h}^R_i),~~
\mathbf{z}_i,\tilde{\mathbf{z}}_i \in\mathbb{R}^{f},
\label{eq:latent_proj}
\end{equation}
where $g(\cdot)$ denotes a Linear$\to$ReLU$\to$Linear MLP and $q(\cdot)$ a single linear layer. During training, a subset of nodes $\mathcal{M}\subseteq\{1,\ldots,N\}$ is selected for feature masking, as detailed below. The Scaled Cosine Error (SCE) \citep{hou2022graphmae} is then used to measure the discrepancy between the graph-conditioned representation $\tilde{\mathbf{z}}_i$ and its corresponding ViT target $\mathbf{z}_i$, supervising the reconstruction of masked nodes:
\begin{equation}
\mathcal{L}_{SCE} = \frac{1}{|\mathcal{M}|}\sum_{i\in\mathcal{M}}
\left(1-\frac{\mathbf{z}_i^T \tilde{\mathbf{z}}_i}{\|\mathbf{z}_i\|\,\|\tilde{\mathbf{z}}_i\|}\right)^{\gamma},~ \gamma\ge 1 .
\label{eq:sce}
\end{equation}
For $\gamma>1$, SCE reduces the contribution of well-reconstructed nodes and emphasizes harder reconstruction cases. This is useful in the few-shot setting, where easily reconstructed normal patches may otherwise dominate the objective. The choice of SCE and $\gamma$ is examined in Appendix~A.8 against MSE and plain cosine error.

\paragraph{Feature masking.}
To prevent direct reproduction of the input features and enforce the use of local context, node feature masking is employed following GraphMAE \citep{hou2022graphmae}. Specifically, at each training iteration, a fraction $\mu$ of the graph nodes is selected to form the masked set $\mathcal{M}$, and their corresponding features are replaced by a learnable mask token. The GAT is therefore required to recover the masked representations from the information propagated through their neighboring nodes, encouraging the learning of local dependencies between patches.

\paragraph{Graph regularization.}
Since FS-IVAD operates under limited normal supervision, the learned attention coefficients may over-adapt to particular neighborhood relations. To regularize the aggregation process, attention dropout is applied to the coefficients $\alpha_{ij}$ during training \citep{velivckovic2017graph}. Specifically, each coefficient is independently masked with probability $\delta$, reducing the dependence on individual edges and encouraging information to be aggregated from multiple neighboring nodes.

\subsection{Inference and anomaly scoring}
\label{subsec:inference}

At inference, a query image $I_{mg}$ is processed by the frozen backbone, yielding the patch-level representation $\mathbf{X}=\{\mathbf{x}_i\}_{i=1}^{N}$. The corresponding nodes are arranged on the same spatial graph and passed through the trained GAT encoder, producing the final representation $\mathbf{H}^R$. 
At inference, the anomaly score $s_i$ of each patch is computed using the SCE reconstruction error in Eq.~\eqref{eq:sce}, evaluated over all nodes. For image-level scoring, the top-$k$ patch scores are selected and averaged, where $k$ is determined from a fixed ratio $\epsilon$ of the total number of patches. For pixel-level scoring, the patch scores $\{s_i\}_{i=1}^{N}$ are arranged according to their spatial locations to form a coarse anomaly map, which is Gaussian-smoothed \citep{damm2025anomalydino} and bilinearly interpolated to the original image resolution.

\section{Experimental results}
\label{sec:experiments}

\subsection{Setup}
\label{ssec:setup}

\paragraph{Datasets.}
Three widely used public benchmarks for high-resolution FS-IVAD are employed: MVTec AD \citep{bergmann2021mvtec} ($3,629$ defect-free training and $1,725$ test images, $5$ texture and $10$ object categories), VisA \citep{zou2022spot} ($10,821$ images: $9,621$ normal and $1,200$ anomalous, $12$ object categories in $3$ domains), and MPDD \citep{jezek2021deep} ($888$ defect-free training and $458$ test images, $6$ object categories from real-world manufacturing).

\paragraph{Experimental protocol and metrics.}
GATE-AD is evaluated under the $k$-shot setting, where $k$ normal samples are provided per class for training and evaluation is performed on the corresponding test set. Image-level performance is measured using AUROC and Average Precision (AP), while pixel-level performance is measured using AUROC and the Area Under the Per-Region Overlap curve (AUPRO). Unless otherwise stated, experiments are repeated over three fixed support-set seeds, and results are reported as mean $\pm$ standard deviation.

\paragraph{Implementation and evaluation configurations.}
GATE-AD employs an input resolution of $\beta=512$, with patch representations obtained by averaging the last $\zeta=8$ backbone layers. The GAT encoder consists of $R=3$ attentional layers with hidden dimension $F=256$, while the latent dimension is set to $f=256$. The graph topology is fixed across images, with each patch connected to its $K=8$ spatially adjacent neighbors on the ViT patch lattice; no feature-space nearest-neighbor search is performed. The SCE scaling factor is $\gamma=2$, the node masking ratio is $\mu=0.2$, and the attention dropout rate is $\delta=0.3$. The dataset-specific backbones and pooling rules, as well as a
detailed hyper-parameter analysis, are provided in Appendix~\ref{sec:hparam}.

Two experimental configurations are used for different purposes. For the controlled cross-method comparison of Section~\ref{ssec:comparative_evaluation}, GATE-AD and the re-evaluated baselines use the same DINOv3 ViT-B/16 backbone, a common top-$1\%$ image-level aggregation rule, and identical support images for each dataset, shot, and support-set seed. For the component and architectural analyses of Section~\ref{ssec:ablation}, GATE-AD instead uses the dataset-specific configurations selected through the hyper-parameter analysis of Appendix~\ref{sec:hparam}; within each benchmark, all remaining settings are kept fixed across matched variants so that only the component or architectural choice under examination is changed.

The model is implemented in PyTorch Geometric and optimized with Adam at a learning rate of $3{\times}10^{-4}$ for a maximum of $2{,}000$ epochs with early stopping. A training-budget analysis is provided in Appendix~\ref{subsec:epoch_budget}. Rotation-based augmentation is applied only to categories that exhibit pose variation, following AnomalyDINO~\citep{damm2025anomalydino}. All experiments were conducted on an NVIDIA RTX~6000 Ada Generation GPU.

\subsection{Comparative evaluation}
\label{ssec:comparative_evaluation}

\begin{table}[!t]
\centering
\scriptsize
\setlength{\tabcolsep}{1.3pt}
\renewcommand{\arraystretch}{0.92}
\caption{FS-IVAD detection results on the MVTec AD, VisA, and MPDD benchmarks. Methods marked with $^{*}$ are re-evaluated under the unified experimental protocol.}
\label{tab:mvtec_visa_mpdd_allinone}
\resizebox{\textwidth}{!}{%
\begin{tabular}{@{}l*{12}{c}@{}}
\toprule
\multicolumn{1}{c}{\multirow{5}{*}{Method}}
& \multicolumn{4}{c}{MVTec AD} & \multicolumn{4}{c}{VisA} & \multicolumn{4}{c}{MPDD} \\
\cmidrule(lr){2-5}\cmidrule(lr){6-9}\cmidrule(lr){10-13}
&
\multicolumn{2}{c}{Image} & \multicolumn{2}{c}{Pixel}
& \multicolumn{2}{c}{Image} & \multicolumn{2}{c}{Pixel}
& \multicolumn{2}{c}{Image} & \multicolumn{2}{c}{Pixel} \\
\cmidrule(lr){2-3}\cmidrule(lr){4-5}
\cmidrule(lr){6-7}\cmidrule(lr){8-9}
\cmidrule(lr){10-11}\cmidrule(lr){12-13}
&
AUROC & AP & AUROC & AUPRO
& AUROC & AP & AUROC & AUPRO
& AUROC & AP & AUROC & AUPRO \\

\midrule
\multicolumn{13}{c}{\textbf{1-shot}} \\
\midrule
\rowcolor{famA} SPADE~\citep{cohen2020sub}\,\scalebox{0.8}{(arXiv'20)} & \mstd{82.9}{2.6} & \mstd{91.7}{1.2} & \mstd{92.0}{0.3} & \mstd{85.7}{0.7} & \mstd{80.7}{5.0} & \mstd{82.3}{4.3} & \mstd{96.2}{0.4} & \mstd{85.7}{1.1} & \mstd{-}{-} & \mstd{-}{-} & \mstd{-}{-} & \mstd{-}{-} \\
\rowcolor{famA} PatchCore~\citep{roth2022towards}\,\scalebox{0.8}{(CVPR'22)} & \mstd{83.4}{3.0} & \mstd{92.2}{1.5} & \mstd{92.0}{1.0} & \mstd{79.7}{2.0} & \mstd{79.9}{2.9} & \mstd{82.8}{2.3} & \mstd{95.4}{0.6} & \mstd{80.5}{2.5} & \mstd{-}{-} & \mstd{-}{-} & \mstd{-}{-} & \mstd{-}{-} \\
\rowcolor{famA} AnomalyDINO$^{*}$~\citep{damm2025anomalydino}\,\scalebox{0.8}{(WACV'25)} & \underline{\mstd{96.2}{0.8}} & \mstd{98.0}{0.4} & \textbf{\mstd{97.2}{0.0}} & \textbf{\mstd{94.6}{0.1}} & \mstd{90.3}{0.5} & \mstd{91.3}{0.5} & \mstd{97.5}{0.1} & \mstd{91.1}{0.5} & \mstd{69.2}{1.5} & \textbf{\mstd{72.5}{1.3}} & \textbf{\mstd{97.3}{0.1}} & \underline{\mstd{92.0}{0.1}} \\
\rowcolor{famA} VisionAD$^{*}$~\citep{wang2025search}\,\scalebox{0.8}{(arXiv'25)} & \mstd{96.1}{0.3} & \underline{\mstd{98.3}{0.1}} & \mstd{97.0}{0.1} & \underline{\mstd{93.9}{0.1}} & \underline{\mstd{92.9}{0.2}} & \underline{\mstd{94.0}{0.2}} & \underline{\mstd{97.7}{0.1}} & \underline{\mstd{93.0}{0.5}} & \mstd{69.0}{0.8} & \mstd{72.0}{0.4} & \mstd{97.0}{0.2} & \mstd{91.6}{0.3} \\
\rowcolor{famA} SubspaceAD$^{*}$~\citep{lendering2026subspacead}\,\scalebox{0.8}{(CVPR'26)} & \mstd{95.8}{0.8} & \mstd{98.0}{0.3} & \underline{\mstd{97.0}{0.1}} & \mstd{93.6}{0.1} & \mstd{89.3}{1.2} & \mstd{90.7}{0.8} & \mstd{97.5}{0.1} & \mstd{92.5}{0.5} & \underline{\mstd{69.2}{0.9}} & \mstd{72.1}{1.3} & \underline{\mstd{97.2}{0.1}} & \mstd{91.9}{0.4} \\
\rowcolor{famB} WinCLIP+~\citep{jeong2023winclip}\,\scalebox{0.8}{(CVPR'23)} & \mstd{93.1}{2.0} & \mstd{96.5}{0.9} & \mstd{95.2}{0.5} & \mstd{87.1}{1.2} & \mstd{83.8}{4.0} & \mstd{85.1}{4.0} & \mstd{96.4}{0.4} & \mstd{85.1}{2.1} & \mstd{-}{-} & \mstd{-}{-} & \mstd{-}{-} & \mstd{-}{-} \\
\rowcolor{famB} APRIL-GAN~\citep{chen2023april}\,\scalebox{0.8}{(CVPRW'23)} & \mstd{92.0}{0.3} & \mstd{95.8}{0.2} & \mstd{95.1}{0.1} & \mstd{90.6}{0.2} & \mstd{91.2}{0.8} & \mstd{93.3}{0.8} & \mstd{96.0}{0.0} & \mstd{90.0}{0.1} & \mstd{-}{-} & \mstd{-}{-} & \mstd{-}{-} & \mstd{-}{-} \\
\rowcolor{famB} PromptAD~\citep{li2024promptad}\,\scalebox{0.8}{(WACV'24)} & \mstd{94.6}{1.7} & \mstd{97.1}{1.0} & \mstd{95.9}{0.5} & \mstd{87.9}{1.0} & \mstd{86.9}{2.3} & \mstd{88.4}{2.6} & \mstd{96.7}{0.4} & \mstd{85.1}{2.5} & \mstd{-}{-} & \mstd{-}{-} & \mstd{-}{-} & \mstd{-}{-} \\
\rowcolor{famB} IIPAD~\citep{lv2025oneforall}\,\scalebox{0.8}{(ICLR'25)} & \mstd{94.2}{-} & \mstd{97.2}{-} & \mstd{96.4}{-} & \mstd{89.8}{-} & \mstd{85.4}{-} & \mstd{87.5}{-} & \mstd{96.9}{-} & \mstd{87.3}{-} & \mstd{-}{-} & \mstd{-}{-} & \mstd{-}{-} & \mstd{-}{-} \\
\rowcolor{famC} FoundAD$^{*}$~\citep{zhai2025foundation}\,\scalebox{0.8}{(ICLR'26)} & \mstd{94.6}{0.8} & \mstd{97.1}{0.4} & \mstd{96.7}{0.1} & \mstd{93.1}{0.1} & \mstd{89.2}{1.3} & \mstd{90.6}{1.0} & \mstd{97.0}{0.1} & \mstd{91.0}{0.4} & \mstd{63.1}{0.6} & \mstd{68.1}{1.1} & \mstd{96.9}{0.2} & \mstd{91.2}{0.4} \\
\rowcolor{famC} GATE-AD$^{*}$ & \textbf{\mstd{97.1}{0.2}} & \textbf{\mstd{98.3}{0.1}} & \mstd{96.9}{0.0} & \mstd{93.7}{0.1} & \textbf{\mstd{93.5}{0.1}} & \textbf{\mstd{94.3}{0.2}} & \textbf{\mstd{97.7}{0.0}} & \textbf{\mstd{93.2}{0.2}} & \textbf{\mstd{70.7}{1.5}} & \underline{\mstd{72.3}{1.0}} & \mstd{97.1}{0.1} & \textbf{\mstd{92.0}{0.4}} \\
\midrule
\multicolumn{13}{c}{\textbf{2-shot}} \\
\midrule
\rowcolor{famA} SPADE~\citep{cohen2020sub}\,\scalebox{0.8}{(arXiv'20)} & \mstd{81.0}{2.0} & \mstd{90.6}{0.8} & \mstd{91.2}{0.4} & \mstd{83.9}{0.7} & \mstd{79.5}{4.0} & \mstd{82.0}{3.3} & \mstd{95.6}{0.4} & \mstd{84.1}{1.6} & \mstd{-}{-} & \mstd{-}{-} & \mstd{-}{-} & \mstd{-}{-} \\
\rowcolor{famA} PatchCore~\citep{roth2022towards}\,\scalebox{0.8}{(CVPR'22)} & \mstd{86.3}{3.3} & \mstd{93.8}{1.7} & \mstd{93.3}{0.6} & \mstd{82.3}{1.3} & \mstd{81.6}{4.0} & \mstd{84.8}{3.2} & \mstd{96.1}{0.5} & \mstd{82.6}{2.3} & \mstd{-}{-} & \mstd{-}{-} & \mstd{-}{-} & \mstd{-}{-} \\
\rowcolor{famA} AnomalyDINO$^{*}$~\citep{damm2025anomalydino}\,\scalebox{0.8}{(WACV'25)} & \mstd{96.7}{1.2} & \mstd{98.2}{0.7} & \textbf{\mstd{97.5}{0.1}} & \textbf{\mstd{95.3}{0.2}} & \mstd{91.1}{0.5} & \mstd{91.8}{0.4} & \mstd{97.9}{0.1} & \mstd{92.2}{0.6} & \underline{\mstd{73.5}{3.0}} & \mstd{75.4}{2.0} & \mstd{97.4}{0.1} & \mstd{92.6}{0.3} \\
\rowcolor{famA} VisionAD$^{*}$~\citep{wang2025search}\,\scalebox{0.8}{(arXiv'25)} & \underline{\mstd{97.3}{0.5}} & \underline{\mstd{98.2}{0.2}} & \mstd{97.3}{0.1} & \underline{\mstd{94.6}{0.1}} & \underline{\mstd{92.5}{0.6}} & \underline{\mstd{93.4}{0.5}} & \mstd{97.8}{0.1} & \mstd{93.4}{0.5} & \mstd{73.5}{2.7} & \underline{\mstd{76.4}{2.6}} & \mstd{97.3}{0.1} & \mstd{92.5}{0.1} \\
\rowcolor{famA} SubspaceAD$^{*}$~\citep{lendering2026subspacead}\,\scalebox{0.8}{(CVPR'26)} & \mstd{96.0}{0.7} & \mstd{98.1}{0.3} & \mstd{97.1}{0.1} & \mstd{93.8}{0.1} & \mstd{90.4}{1.2} & \mstd{91.4}{0.8} & \mstd{97.6}{0.1} & \mstd{92.8}{0.4} & \mstd{71.2}{2.5} & \mstd{73.9}{2.4} & \mstd{97.4}{0.1} & \mstd{92.6}{0.4} \\
\rowcolor{famB} WinCLIP+~\citep{jeong2023winclip}\,\scalebox{0.8}{(CVPR'23)} & \mstd{94.4}{1.3} & \mstd{97.0}{0.7} & \mstd{96.0}{0.3} & \mstd{88.4}{0.9} & \mstd{84.6}{2.4} & \mstd{85.8}{2.7} & \mstd{96.8}{0.3} & \mstd{86.2}{1.4} & \mstd{-}{-} & \mstd{-}{-} & \mstd{-}{-} & \mstd{-}{-} \\
\rowcolor{famB} APRIL-GAN~\citep{chen2023april}\,\scalebox{0.8}{(CVPRW'23)} & \mstd{92.4}{0.3} & \mstd{96.0}{0.2} & \mstd{95.5}{0.0} & \mstd{91.3}{0.1} & \mstd{92.2}{0.3} & \mstd{94.2}{0.3} & \mstd{96.2}{0.0} & \mstd{90.1}{0.1} & \mstd{-}{-} & \mstd{-}{-} & \mstd{-}{-} & \mstd{-}{-} \\
\rowcolor{famB} PromptAD~\citep{li2024promptad}\,\scalebox{0.8}{(WACV'24)} & \mstd{95.7}{1.5} & \mstd{97.9}{0.7} & \mstd{96.2}{0.3} & \mstd{88.5}{0.8} & \mstd{88.3}{2.0} & \mstd{96.1}{0.6} & \mstd{97.1}{0.3} & \mstd{85.8}{2.1} & \mstd{-}{-} & \mstd{-}{-} & \mstd{-}{-} & \mstd{-}{-} \\
\rowcolor{famB} IIPAD~\citep{lv2025oneforall}\,\scalebox{0.8}{(ICLR'25)} & \mstd{95.7}{-} & \mstd{97.9}{-} & \mstd{96.7}{-} & \mstd{90.3}{-} & \mstd{86.7}{-} & \mstd{88.6}{-} & \mstd{97.2}{-} & \mstd{87.9}{-} & \mstd{-}{-} & \mstd{-}{-} & \mstd{-}{-} & \mstd{-}{-} \\
\rowcolor{famC} RegAD~\citep{huang2022registration}\,\scalebox{0.8}{(ECCV'22)} & \mstd{85.7}{-} & \mstd{-}{-} & \mstd{94.6}{-} & \mstd{-}{-} & \mstd{-}{-} & \mstd{-}{-} & \mstd{-}{-} & \mstd{-}{-} & \mstd{63.4}{-} & \mstd{-}{-} & \mstd{93.2}{-} & \mstd{-}{-} \\
\rowcolor{famC} FastRecon~\citep{fang2023fastrecon}\,\scalebox{0.8}{(ICCV'23)} & \mstd{91.0}{-} & \mstd{-}{-} & \mstd{95.9}{-} & \mstd{-}{-} & \mstd{-}{-} & \mstd{-}{-} & \mstd{-}{-} & \mstd{-}{-} & \mstd{73.7}{-} & \mstd{-}{-} & \mstd{97.0}{-} & \mstd{-}{-} \\
\rowcolor{famC} FoundAD$^{*}$~\citep{zhai2025foundation}\,\scalebox{0.8}{(ICLR'26)} & \mstd{96.2}{0.7} & \mstd{98.0}{0.4} & \underline{\mstd{97.4}{0.1}} & \mstd{94.2}{0.1} & \mstd{91.4}{1.0} & \mstd{91.6}{0.6} & \underline{\mstd{97.9}{0.0}} & \underline{\mstd{93.5}{0.3}} & \mstd{70.8}{5.5} & \mstd{73.2}{4.6} & \underline{\mstd{97.5}{0.1}} & \underline{\mstd{93.0}{0.4}} \\
\rowcolor{famC} GATE-AD$^{*}$ & \textbf{\mstd{97.6}{0.2}} & \textbf{\mstd{98.5}{0.1}} & \mstd{97.3}{0.0} & \mstd{94.3}{0.1} & \textbf{\mstd{93.0}{0.2}} & \textbf{\mstd{93.7}{0.3}} & \textbf{\mstd{98.0}{0.0}} & \textbf{\mstd{93.6}{0.2}} & \textbf{\mstd{80.9}{5.4}} & \textbf{\mstd{81.4}{5.0}} & \textbf{\mstd{97.6}{0.1}} & \textbf{\mstd{93.9}{0.4}} \\
\midrule
\multicolumn{13}{c}{\textbf{4-shot}} \\
\midrule
\rowcolor{famA} SPADE~\citep{cohen2020sub}\,\scalebox{0.8}{(arXiv'20)} & \mstd{84.8}{2.5} & \mstd{92.5}{1.2} & \mstd{92.7}{0.3} & \mstd{87.0}{0.5} & \mstd{81.7}{3.4} & \mstd{83.4}{2.7} & \mstd{96.6}{0.3} & \mstd{87.3}{0.8} & \mstd{-}{-} & \mstd{-}{-} & \mstd{-}{-} & \mstd{-}{-} \\
\rowcolor{famA} PatchCore~\citep{roth2022towards}\,\scalebox{0.8}{(CVPR'22)} & \mstd{88.8}{2.6} & \mstd{94.5}{1.5} & \mstd{94.3}{0.5} & \mstd{84.3}{1.6} & \mstd{85.3}{2.1} & \mstd{87.5}{2.1} & \mstd{96.8}{0.3} & \mstd{84.9}{1.4} & \mstd{-}{-} & \mstd{-}{-} & \mstd{-}{-} & \mstd{-}{-} \\
\rowcolor{famA} AnomalyDINO$^{*}$~\citep{damm2025anomalydino}\,\scalebox{0.8}{(WACV'25)} & \underline{\mstd{98.0}{0.1}} & \underline{\mstd{98.8}{0.1}} & \underline{\mstd{97.8}{0.1}} & \textbf{\mstd{95.7}{0.2}} & \mstd{92.3}{0.4} & \mstd{92.9}{0.3} & \mstd{98.2}{0.0} & \mstd{93.0}{0.2} & \mstd{76.0}{2.8} & \mstd{77.9}{2.4} & \underline{\mstd{97.7}{0.1}} & \mstd{93.3}{0.3} \\
\rowcolor{famA} VisionAD$^{*}$~\citep{wang2025search}\,\scalebox{0.8}{(arXiv'25)} & \textbf{\mstd{98.0}{0.3}} & \textbf{\mstd{99.2}{0.1}} & \mstd{97.5}{0.1} & \underline{\mstd{94.9}{0.1}} & \underline{\mstd{94.4}{0.4}} & \textbf{\mstd{95.1}{0.4}} & \underline{\mstd{98.4}{0.0}} & \underline{\mstd{94.6}{0.2}} & \underline{\mstd{76.3}{2.6}} & \underline{\mstd{79.2}{3.0}} & \mstd{97.7}{0.2} & \mstd{93.3}{0.3} \\
\rowcolor{famA} SubspaceAD$^{*}$~\citep{lendering2026subspacead}\,\scalebox{0.8}{(CVPR'26)} & \mstd{96.5}{0.3} & \mstd{98.5}{0.1} & \mstd{97.2}{0.1} & \mstd{93.9}{0.1} & \mstd{92.0}{0.8} & \mstd{92.4}{0.5} & \mstd{97.6}{0.0} & \mstd{93.0}{0.1} & \mstd{70.8}{1.6} & \mstd{74.7}{2.2} & \mstd{97.5}{0.1} & \mstd{92.8}{0.3} \\
\rowcolor{famB} WinCLIP+~\citep{jeong2023winclip}\,\scalebox{0.8}{(CVPR'23)} & \mstd{95.2}{1.3} & \mstd{97.3}{0.6} & \mstd{96.2}{0.3} & \mstd{89.0}{0.8} & \mstd{87.3}{1.8} & \mstd{88.8}{1.8} & \mstd{97.2}{0.2} & \mstd{87.6}{0.9} & \mstd{-}{-} & \mstd{-}{-} & \mstd{-}{-} & \mstd{-}{-} \\
\rowcolor{famB} APRIL-GAN~\citep{chen2023april}\,\scalebox{0.8}{(CVPRW'23)} & \mstd{92.8}{0.2} & \mstd{96.3}{0.1} & \mstd{95.9}{0.0} & \mstd{91.8}{0.1} & \mstd{92.6}{0.4} & \mstd{94.5}{0.3} & \mstd{96.2}{0.0} & \mstd{90.2}{0.1} & \mstd{-}{-} & \mstd{-}{-} & \mstd{-}{-} & \mstd{-}{-} \\
\rowcolor{famB} PromptAD~\citep{li2024promptad}\,\scalebox{0.8}{(WACV'24)} & \mstd{96.6}{0.9} & \mstd{98.5}{0.5} & \mstd{96.5}{0.2} & \mstd{90.5}{0.7} & \mstd{89.1}{1.7} & \mstd{90.8}{1.3} & \mstd{97.4}{0.3} & \mstd{86.2}{1.7} & \mstd{-}{-} & \mstd{-}{-} & \mstd{-}{-} & \mstd{-}{-} \\
\rowcolor{famB} IIPAD~\citep{lv2025oneforall}\,\scalebox{0.8}{(ICLR'25)} & \mstd{96.1}{-} & \mstd{98.1}{-} & \mstd{97.0}{-} & \mstd{91.2}{-} & \mstd{88.3}{-} & \mstd{89.6}{-} & \mstd{97.4}{-} & \mstd{88.3}{-} & \mstd{-}{-} & \mstd{-}{-} & \mstd{-}{-} & \mstd{-}{-} \\
\rowcolor{famC} RegAD~\citep{huang2022registration}\,\scalebox{0.8}{(ECCV'22)} & \mstd{88.2}{-} & \mstd{-}{-} & \mstd{95.8}{-} & \mstd{-}{-} & \mstd{-}{-} & \mstd{-}{-} & \mstd{-}{-} & \mstd{-}{-} & \mstd{68.3}{-} & \mstd{-}{-} & \mstd{93.9}{-} & \mstd{-}{-} \\
\rowcolor{famC} FastRecon~\citep{fang2023fastrecon}\,\scalebox{0.8}{(ICCV'23)} & \mstd{94.2}{-} & \mstd{-}{-} & \mstd{97.0}{-} & \mstd{-}{-} & \mstd{-}{-} & \mstd{-}{-} & \mstd{-}{-} & \mstd{-}{-} & \mstd{79.9}{-} & \mstd{-}{-} & \mstd{97.6}{-} & \mstd{-}{-} \\
\rowcolor{famC} FoundAD$^{*}$~\citep{zhai2025foundation}\,\scalebox{0.8}{(ICLR'26)} & \mstd{97.4}{0.5} & \mstd{98.8}{0.2} & \textbf{\mstd{97.9}{0.1}} & \mstd{94.8}{0.2} & \mstd{93.2}{0.5} & \mstd{92.8}{0.4} & \textbf{\mstd{98.6}{0.0}} & \textbf{\mstd{95.2}{0.1}} & \mstd{74.3}{2.5} & \mstd{77.4}{2.0} & \textbf{\mstd{97.9}{0.2}} & \underline{\mstd{94.0}{0.6}} \\
\rowcolor{famC} GATE-AD$^{*}$ & \mstd{97.7}{0.1} & \mstd{98.6}{0.1} & \mstd{97.6}{0.0} & \mstd{94.5}{0.0} & \textbf{\mstd{94.5}{0.1}} & \underline{\mstd{94.7}{0.3}} & \mstd{98.3}{0.0} & \mstd{94.5}{0.1} & \textbf{\mstd{79.0}{3.0}} & \textbf{\mstd{82.0}{3.8}} & \mstd{97.7}{0.1} & \textbf{\mstd{94.1}{0.4}} \\
\midrule
\multicolumn{13}{c}{\textbf{8-shot}} \\
\midrule
\rowcolor{famA} AnomalyDINO$^{*}$~\citep{damm2025anomalydino}\,\scalebox{0.8}{(WACV'25)} & \textbf{\mstd{98.5}{0.1}} & \underline{\mstd{99.3}{0.1}} & \underline{\mstd{98.0}{0.0}} & \textbf{\mstd{96.2}{0.1}} & \mstd{93.0}{0.3} & \mstd{93.2}{0.3} & \mstd{98.4}{0.0} & \mstd{93.7}{0.2} & \mstd{78.2}{2.4} & \mstd{79.6}{2.6} & \underline{\mstd{98.0}{0.0}} & \mstd{93.8}{0.2} \\
\rowcolor{famA} VisionAD$^{*}$~\citep{wang2025search}\,\scalebox{0.8}{(arXiv'25)} & \underline{\mstd{98.3}{0.2}} & \textbf{\mstd{99.4}{0.1}} & \mstd{97.7}{0.0} & \underline{\mstd{95.4}{0.1}} & \underline{\mstd{95.1}{0.2}} & \underline{\mstd{95.3}{0.2}} & \underline{\mstd{98.6}{0.0}} & \underline{\mstd{95.2}{0.2}} & \underline{\mstd{80.1}{1.4}} & \underline{\mstd{82.3}{1.8}} & \mstd{97.9}{0.1} & \underline{\mstd{93.9}{0.1}} \\
\rowcolor{famA} SubspaceAD$^{*}$~\citep{lendering2026subspacead}\,\scalebox{0.8}{(CVPR'26)} & \mstd{96.8}{0.5} & \mstd{98.8}{0.2} & \mstd{97.2}{0.0} & \mstd{94.1}{0.2} & \mstd{92.6}{0.6} & \mstd{93.1}{0.7} & \mstd{97.6}{0.0} & \mstd{93.1}{0.2} & \mstd{70.8}{2.6} & \mstd{74.3}{1.5} & \mstd{97.6}{0.1} & \mstd{92.7}{0.5} \\
\rowcolor{famC} RegAD~\citep{huang2022registration}\,\scalebox{0.8}{(ECCV'22)} & \mstd{91.2}{-} & \mstd{-}{-} & \mstd{96.7}{-} & \mstd{-}{-} & \mstd{-}{-} & \mstd{-}{-} & \mstd{-}{-} & \mstd{-}{-} & \mstd{71.9}{-} & \mstd{-}{-} & \mstd{95.1}{-} & \mstd{-}{-} \\
\rowcolor{famC} FastRecon~\citep{fang2023fastrecon}\,\scalebox{0.8}{(ICCV'23)} & \mstd{95.2}{-} & \mstd{-}{-} & \mstd{97.3}{-} & \mstd{-}{-} & \mstd{-}{-} & \mstd{-}{-} & \mstd{-}{-} & \mstd{-}{-} & \mstd{82.5}{-} & \mstd{-}{-} & \mstd{97.9}{-} & \mstd{-}{-} \\
\rowcolor{famC} FoundAD$^{*}$~\citep{zhai2025foundation}\,\scalebox{0.8}{(ICLR'26)} & \mstd{97.9}{0.3} & \mstd{99.0}{0.0} & \textbf{\mstd{98.1}{0.0}} & \mstd{95.3}{0.1} & \mstd{94.1}{0.1} & \mstd{93.5}{0.1} & \textbf{\mstd{98.9}{0.0}} & \textbf{\mstd{96.0}{0.1}} & \mstd{78.5}{2.4} & \mstd{80.5}{1.5} & \textbf{\mstd{98.3}{0.1}} & \textbf{\mstd{94.8}{0.1}} \\
\rowcolor{famC} GATE-AD$^{*}$ & \mstd{97.8}{0.1} & \mstd{98.7}{0.0} & \mstd{97.7}{0.0} & \mstd{94.7}{0.0} & \textbf{\mstd{95.3}{0.1}} & \textbf{\mstd{95.4}{0.3}} & \mstd{98.4}{0.0} & \mstd{94.6}{0.1} & \textbf{\mstd{80.8}{3.0}} & \textbf{\mstd{82.8}{2.1}} & \mstd{97.7}{0.1} & \mstd{93.7}{0.4} \\
\bottomrule
\end{tabular}%
}
\end{table}

\begin{figure}[tb]
  \centering
  \includegraphics[width=0.80\linewidth]{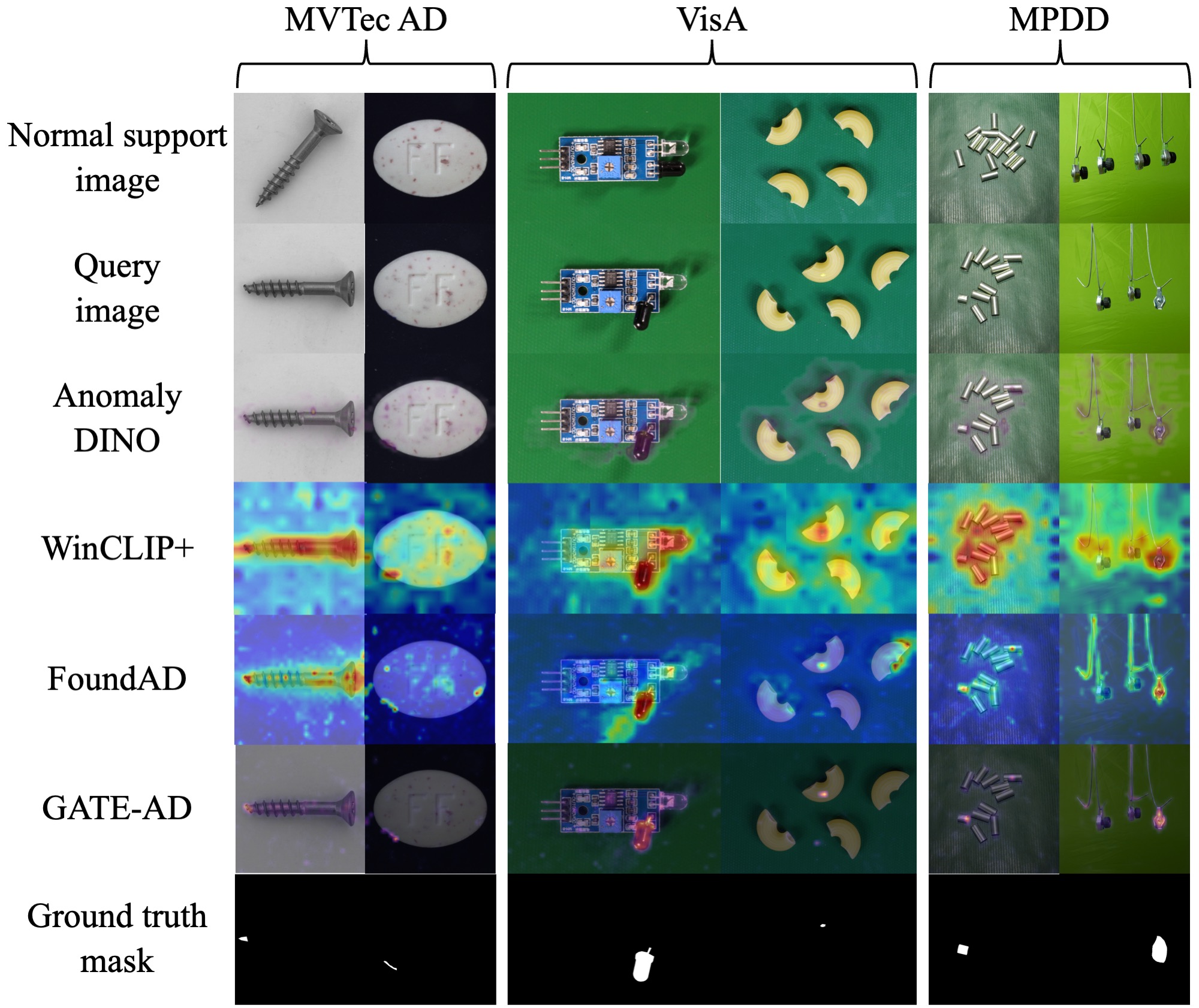}
  \caption{Indicative $1$-shot defect detection results. GATE-AD localizes more accurately and robustly defects, across multiple pose, placement, and abnormality type settings.}
  \label{fig:qualitative_examples}
\end{figure}

\paragraph{Baselines.}
GATE-AD is compared with representative FS-IVAD methods spanning three main families: a) feature-based methods, including SPADE~\citep{cohen2020sub}, PatchCore~\citep{roth2022towards}, AnomalyDINO$^{*}$~\citep{damm2025anomalydino}, VisionAD$^{*}$~\citep{wang2025search}, and SubspaceAD$^{*}$~\citep{lendering2026subspacead}; b) vision-language methods, including WinCLIP+~\citep{jeong2023winclip}, PromptAD~\citep{li2024promptad}, IIPAD~\citep{lv2025oneforall}, and APRIL-GAN~\citep{chen2023april}; and c) reconstruction- and projection-based methods, including FastRecon~\citep{fang2023fastrecon}, FoundAD$^{*}$~\citep{zhai2025foundation}, and RegAD~\citep{huang2022registration}. Results without $^{*}$ are taken from the literature, while methods marked with $^{*}$ are additionally re-evaluated under the unified experimental protocol defined in Section~\ref{ssec:setup}, following the few-shot sampling procedure of AnomalyDINO~\citep{damm2025anomalydino}. These methods are selected because public implementations are available and permit evaluation under the common protocol without altering their core anomaly-detection mechanism. Table~\ref{tab:mvtec_visa_mpdd_allinone} reports the $1$/$2$/$4$/$8$-shot results on MVTec AD, VisA, and MPDD. Indicative $1$-shot detection results are shown in Fig.~\ref{fig:qualitative_examples}.

\paragraph{Discussion.}\
From Table~\ref{tab:mvtec_visa_mpdd_allinone}, the following observations can be made: a) GATE-AD \textbf{matches or exceeds the state of the art in most image-level settings}, achieving the highest image AUROC in $10/12$ and AP in $8/12$ cases under the unified protocol, while remaining competitive at the pixel level. b) It shows a clear \textbf{advantage in the very low-shot regime} ($1/2$-shots), where normal evidence is most limited. c) GATE-AD maintains \textbf{stable performance across shot counts and support selections}, with the performance gap decreasing as additional normal references become available. d) The gains are particularly evident on \textbf{MPDD}, the most challenging manufacturing benchmark considered. The above observations are consistent with the proposed normality-learning scheme: masking enforces neighborhood-based reconstruction, attention dropout regularizes local dependencies, and representation alignment supervises recovery in a compact latent space rather than directly in the original DINO representation. Together, these mechanisms enable relational normality to be learned effectively from very few normal samples. Detailed object-level results, product-agnostic defect-type analysis, qualitative examples, and failure cases are provided in Appendix~\ref{sec:perclass}, with the defect-type analysis showing stronger localization for local surface defects than for broader structural defects.

\subsection{Ablation study}
\label{ssec:ablation}

\begin{table*}[t!]
\centering
\scriptsize
\setlength{\tabcolsep}{2.0pt}
\renewcommand{\arraystretch}{0.92}

\caption{Component ablation of GATE-AD on MVTec AD, VisA, and MPDD
(\%, mean$\pm$std).}
\label{tab:component_ablation}

\begin{tabular}{@{}lcccccccccc@{}}
\toprule
\multirow{2}{*}{\textbf{Variant}}
& \multicolumn{4}{c}{\textbf{Components}}
& \multicolumn{2}{c}{\textbf{MVTec AD}}
& \multicolumn{2}{c}{\textbf{VisA}}
& \multicolumn{2}{c}{\textbf{MPDD}} \\
\cmidrule(lr){2-5}
\cmidrule(lr){6-7}
\cmidrule(lr){8-9}
\cmidrule(lr){10-11}
& \textbf{GAT}
& \textbf{RA}
& \textbf{Mask.}
& \textbf{Drop.}
& \textbf{I-AUROC}
& \textbf{P-AUROC}
& \textbf{I-AUROC}
& \textbf{P-AUROC}
& \textbf{I-AUROC}
& \textbf{P-AUROC} \\
\midrule

\multicolumn{11}{c}{\textbf{1-shot}} \\
\midrule

DINO 1-NN
& -- & -- & -- & --
& \mstd{96.2}{0.7} & \mstd{95.9}{0.1}
& \mstd{87.4}{1.1} & \mstd{97.1}{0.1}
& \mstd{78.0}{2.8} & \mstd{96.9}{0.1} \\

DINO + RA
& -- & \checkmark & -- & --
& \mstd{96.0}{0.1} & \mstd{96.1}{0.0}
& \mstd{88.7}{0.3} & \mstd{97.0}{0.1}
& \mstd{79.3}{4.2} & \mstd{97.2}{0.3} \\

GATE-AD w/o RA
& \checkmark & -- & \checkmark & \checkmark
& \mstd{94.1}{0.5} & \mstd{95.1}{0.1}
& \mstd{80.5}{0.3} & \mstd{96.0}{0.2}
& \mstd{55.7}{1.1} & \mstd{94.4}{0.4} \\

GATE-AD w/o mask. \& drop.
& \checkmark & \checkmark & -- & --
& \mstd{96.0}{0.5} & \mstd{95.7}{0.1}
& \mstd{91.1}{0.9} & \mstd{95.5}{0.4}
& \mstd{81.2}{0.7} & \mstd{95.9}{0.3} \\

GATE-AD w/o dropout
& \checkmark & \checkmark & \checkmark & --
& \mstd{96.3}{0.5} & \mstd{96.0}{0.1}
& \mstd{91.9}{0.3} & \mstd{96.9}{0.4}
& \mstd{82.5}{0.4} & \mstd{96.3}{0.3} \\

GATE-AD w/o masking
& \checkmark & \checkmark & -- & \checkmark
& \mstd{96.8}{0.5} & \mstd{95.9}{0.0}
& \mstd{91.6}{0.9} & \mstd{97.2}{0.4}
& \mstd{82.4}{0.7} & \mstd{96.4}{0.3} \\

GATE-AD
& \checkmark & \checkmark & \checkmark & \checkmark
& \textbf{\mstd{97.7}{0.2}} & \textbf{\mstd{96.3}{0.1}}
& \textbf{\mstd{93.5}{0.2}} & \textbf{\mstd{97.7}{0.1}}
& \textbf{\mstd{84.2}{0.9}} & \textbf{\mstd{97.5}{0.1}} \\

\midrule
\multicolumn{11}{c}{\textbf{2-shot}} \\
\midrule

DINO 1-NN
& -- & -- & -- & --
& \mstd{96.7}{0.9} & \mstd{96.5}{0.1}
& \mstd{89.4}{1.1} & \mstd{97.5}{0.1}
& \mstd{81.1}{1.4} & \mstd{97.1}{0.1} \\

DINO + RA
& -- & \checkmark & -- & --
& \mstd{96.6}{0.0} & \mstd{96.6}{0.1}
& \mstd{90.2}{0.2} & \mstd{97.9}{0.0}
& \mstd{81.1}{2.2} & \mstd{97.0}{0.2} \\

GATE-AD w/o RA
& \checkmark & -- & \checkmark & \checkmark
& \mstd{94.8}{0.2} & \mstd{95.7}{0.1}
& \mstd{82.5}{1.4} & \mstd{96.4}{0.2}
& \mstd{58.2}{1.0} & \mstd{94.6}{0.5} \\

GATE-AD w/o mask. \& drop.
& \checkmark & \checkmark & -- & --
& \mstd{95.3}{1.0} & \mstd{96.1}{0.2}
& \mstd{91.1}{0.1} & \mstd{95.7}{0.2}
& \mstd{82.2}{0.6} & \mstd{96.2}{0.3} \\

GATE-AD w/o dropout
& \checkmark & \checkmark & \checkmark & --
& \mstd{97.0}{0.9} & \mstd{96.6}{0.4}
& \mstd{92.4}{0.4} & \mstd{96.6}{0.5}
& \mstd{83.5}{0.8} & \mstd{96.7}{0.5} \\

GATE-AD w/o masking
& \checkmark & \checkmark & -- & \checkmark
& \mstd{96.7}{1.1} & \mstd{96.4}{0.3}
& \mstd{92.2}{0.1} & \mstd{96.4}{0.2}
& \mstd{83.4}{0.6} & \mstd{96.4}{0.5} \\

GATE-AD
& \checkmark & \checkmark & \checkmark & \checkmark
& \textbf{\mstd{98.0}{0.2}} & \textbf{\mstd{96.8}{0.1}}
& \textbf{\mstd{92.8}{0.3}} & \textbf{\mstd{97.9}{0.1}}
& \textbf{\mstd{84.3}{0.5}} & \textbf{\mstd{97.5}{0.1}} \\

\midrule
\multicolumn{11}{c}{\textbf{4-shot}} \\
\midrule

DINO 1-NN
& -- & -- & -- & --
& \mstd{97.6}{0.8} & \textbf{\mstd{96.9}{0.1}}
& \mstd{91.7}{1.1} & \mstd{97.9}{0.0}
& \mstd{83.6}{1.7} & \mstd{97.3}{0.2} \\

DINO + RA
& -- & \checkmark & -- & --
& \mstd{97.6}{0.0} & \mstd{96.2}{0.0}
& \mstd{91.0}{0.3} & \mstd{97.2}{0.0}
& \mstd{83.4}{3.2} & \mstd{97.5}{0.2} \\

GATE-AD w/o RA
& \checkmark & -- & \checkmark & \checkmark
& \mstd{95.6}{0.4} & \mstd{96.2}{0.2}
& \mstd{87.9}{1.3} & \mstd{96.8}{0.3}
& \mstd{58.6}{3.2} & \mstd{95.1}{0.4} \\

GATE-AD w/o mask. \& drop.
& \checkmark & \checkmark & -- & --
& \mstd{96.9}{0.4} & \mstd{96.5}{0.1}
& \mstd{94.2}{0.2} & \mstd{97.1}{0.2}
& \mstd{83.6}{0.3} & \mstd{96.9}{0.1} \\

GATE-AD w/o dropout
& \checkmark & \checkmark & \checkmark & --
& \mstd{97.4}{0.4} & \mstd{96.6}{0.1}
& \mstd{94.2}{0.4} & \mstd{97.2}{0.1}
& \mstd{83.9}{0.4} & \mstd{97.2}{0.2} \\

GATE-AD w/o masking
& \checkmark & \checkmark & -- & \checkmark
& \mstd{97.3}{0.4} & \mstd{96.7}{0.1}
& \textbf{\mstd{94.4}{0.1}} & \mstd{98.0}{0.2}
& \mstd{83.8}{0.3} & \mstd{97.2}{0.2} \\

GATE-AD
& \checkmark & \checkmark & \checkmark & \checkmark
& \textbf{\mstd{97.9}{0.2}} & \textbf{\mstd{96.9}{0.1}}
& \textbf{\mstd{94.4}{0.2}} & \textbf{\mstd{98.3}{0.1}}
& \textbf{\mstd{84.1}{0.9}} & \textbf{\mstd{97.6}{0.1}} \\

\midrule
\multicolumn{11}{c}{\textbf{8-shot}} \\
\midrule

DINO 1-NN
& -- & -- & -- & --
& \textbf{\mstd{98.3}{0.4}} & \textbf{\mstd{97.2}{0.0}}
& \mstd{93.0}{0.2} & \mstd{98.2}{0.0}
& \mstd{86.0}{1.9} & \mstd{97.6}{0.1} \\

DINO + RA
& -- & \checkmark & -- & --
& \mstd{97.7}{0.0} & \textbf{\mstd{97.2}{0.0}}
& \mstd{92.7}{0.3} & \mstd{98.2}{0.0}
& \mstd{85.5}{3.1} & \mstd{97.6}{0.2} \\

GATE-AD w/o RA
& \checkmark & -- & \checkmark & \checkmark
& \mstd{95.8}{0.6} & \mstd{95.5}{0.2}
& \mstd{90.5}{0.3} & \mstd{97.1}{0.3}
& \mstd{64.5}{1.3} & \mstd{95.5}{0.3} \\

GATE-AD w/o mask. \& drop.
& \checkmark & \checkmark & -- & --
& \mstd{97.3}{0.4} & \mstd{96.7}{0.3}
& \mstd{94.5}{0.3} & \mstd{97.0}{0.1}
& \mstd{87.3}{0.4} & \mstd{97.3}{0.2} \\

GATE-AD w/o dropout
& \checkmark & \checkmark & \checkmark & --
& \mstd{97.5}{0.4} & \mstd{96.8}{0.2}
& \mstd{95.1}{0.4} & \mstd{97.5}{0.2}
& \mstd{87.5}{0.4} & \mstd{97.6}{0.2} \\

GATE-AD w/o masking
& \checkmark & \checkmark & -- & \checkmark
& \mstd{97.7}{0.5} & \mstd{96.8}{0.1}
& \mstd{95.1}{0.2} & \mstd{98.2}{0.1}
& \mstd{87.6}{0.4} & \mstd{97.7}{0.2} \\

GATE-AD
& \checkmark & \checkmark & \checkmark & \checkmark
& \mstd{98.2}{0.1} & \mstd{97.0}{0.1}
& \textbf{\mstd{95.2}{0.2}} & \textbf{\mstd{98.3}{0.1}}
& \textbf{\mstd{87.8}{0.5}} & \textbf{\mstd{97.8}{0.0}} \\

\bottomrule
\end{tabular}
\end{table*}

\paragraph{Component analysis.}
In order to isolate the contribution of the main components of GATE-AD, a `DINO 1-NN' baseline is first defined, following the patch-level nearest-neighbor formulation of AnomalyDINO~\citep{damm2025anomalydino}, where each query patch is scored according to its nearest normal DINO feature, without graph construction or representation alignment. Starting from this baseline, the following variants are considered: a) `DINO + RA', where the same DINO features are projected to the representation-alignment latent space and scored using the SCE objective, without relational modeling; b) `GATE-AD w/o RA', where the graph encoder is retained but RA is removed and the SCE objective is directly estimated between the final GAT representation and the input feature tokens; c) `GATE-AD w/o masking \& dropout', where both low-shot regularization mechanisms are removed; d--e) variants where attention dropout or feature masking is independently discarded; and f) the complete `GATE-AD' configuration. All variants follow the dataset-specific ablation protocol defined in Section~\ref{ssec:setup}, with the remaining configuration kept fixed within each benchmark. Table~\ref{tab:component_ablation} reports I-AUROC and P-AUROC across all datasets and shot settings.
\paragraph{Discussion.}\
From Table~\ref{tab:component_ablation}, the following key observations can be made: a) Frozen DINO features already provide a \textbf{strong normality representation}, as demonstrated by the competitive DINO 1-NN baseline. However, introducing RA directly on the same features does not consistently improve nearest-neighbor matching, with DINO + RA performing below DINO 1-NN in several settings. Hence, \textbf{representation alignment alone is not sufficient to explain the gains of GATE-AD}. b) The complete GATE-AD formulation clearly improves over the non-graph variants, with the largest gains observed in the $1$-shot regime, where normal evidence is most limited. This supports the central premise of the proposed approach: rather than characterizing a patch only through its similarity to stored normal features, normality is learned through its \textbf{consistency with the surrounding spatial context}. c) The substantial degradation observed when RA is removed from the graph-based formulation further shows that \textbf{relational modeling and representation alignment are complementary}. In particular, the GAT estimates a context-dependent representation from the local neighborhood, while RA defines the latent space in which this prediction is compared with the corresponding DINO target. d) Feature masking and attention dropout provide further improvements, particularly in the $1/2$-shot regime. Masking \textbf{enforces neighborhood-based reconstruction} instead of direct feature copying, whereas attention dropout \textbf{regularizes the learned local dependencies} and reduces over-reliance on individual edges; removing both generally leads to a larger performance reduction. Overall, these results indicate that the effectiveness of GATE-AD does not originate from any individual component in isolation, but from their joint formulation for learning \textbf{relational normality from very few normal samples}. A query-conditioned representation analysis in Appendix~\ref{subsec:representation_analysis} further illustrates how the GAT forms a context-conditioned prediction and how its disagreement with the projected DINO target localizes anomalous regions.

\paragraph{Architectural design analysis.}
Following the component analysis, the main architectural choices of GATE-AD are further examined along two complementary directions. First, the choice of relational encoder is investigated by replacing the GAT layers with either conventional Graph Convolutional Network (GCN) layers or global Transformer Encoder (TE) blocks, while retaining the remaining formulation unchanged. Second, the reconstruction formulation is examined through: a) `GATE-AD AE', where RA is replaced by an MLP graph decoder and the SCE objective is estimated between the decoder output and the input feature representation; and b) `MAE', which follows a conventional masked-autoencoding formulation~\citep{he2022masked} and directly reconstructs the input feature tokens. All variants follow the same dataset-specific ablation protocol as the component analysis, with only the architectural choice under examination being modified. Minor numerical variations with respect to the reference GATE-AD results may arise from independent experimental repetitions. Table~\ref{tab:architectural_ablation} reports I-AUROC and P-AUROC across all datasets and shot settings.

\begin{table*}[t!]
\centering
\scriptsize
\setlength{\tabcolsep}{2.6pt}
\renewcommand{\arraystretch}{0.94}

\caption{Architectural design analysis on the MVTec AD, VisA, and MPDD benchmarks (\%, mean$\pm$std). Best results are shown in bold.}
\label{tab:architectural_ablation}

\begin{tabular}{@{}cl cc cc cc@{}}
\toprule
& &
\multicolumn{2}{c}{\textbf{MVTec AD}} &
\multicolumn{2}{c}{\textbf{VisA}} &
\multicolumn{2}{c}{\textbf{MPDD}} \\
\cmidrule(lr){3-4}
\cmidrule(lr){5-6}
\cmidrule(lr){7-8}

\textbf{Shot} &
\textbf{Variant} &
\textbf{I-AUROC} & \textbf{P-AUROC} &
\textbf{I-AUROC} & \textbf{P-AUROC} &
\textbf{I-AUROC} & \textbf{P-AUROC} \\
\midrule

\multirow{5}{*}{\textbf{1}}
& GATE-AD (GAT)
& \textbf{\mstd{97.7}{0.2}} & \textbf{\mstd{96.3}{0.1}}
& \textbf{\mstd{93.5}{0.2}} & \textbf{\mstd{97.7}{0.1}}
& \textbf{\mstd{84.3}{1.1}} & \textbf{\mstd{97.4}{0.2}} \\

& GATE-AD (GCN)
& \mstd{97.3}{0.2} & \mstd{95.9}{0.2}
& \mstd{92.7}{0.4} & \mstd{97.4}{0.1}
& \mstd{82.1}{0.9} & \mstd{95.7}{0.1} \\

& Transformer Encoder (TE)
& \mstd{97.1}{0.2} & \mstd{95.7}{0.1}
& \mstd{93.4}{0.2} & \mstd{97.5}{0.1}
& \mstd{82.4}{0.9} & \mstd{96.3}{0.1} \\

\addlinespace[1pt]

& GATE-AD AE
& \mstd{93.2}{1.0} & \mstd{95.3}{0.1}
& \mstd{80.8}{0.5} & \mstd{96.2}{0.1}
& \mstd{59.2}{2.5} & \mstd{94.2}{0.3} \\

& MAE
& \mstd{66.6}{5.0} & \mstd{62.0}{10.1}
& \mstd{55.0}{1.6} & \mstd{71.2}{0.6}
& \mstd{45.0}{2.7} & \mstd{87.2}{6.2} \\

\midrule

\multirow{5}{*}{\textbf{2}}
& GATE-AD (GAT)
& \textbf{\mstd{98.0}{0.2}} & \textbf{\mstd{96.8}{0.1}}
& \textbf{\mstd{92.8}{0.3}} & \textbf{\mstd{97.9}{0.1}}
& \textbf{\mstd{84.5}{0.7}} & \textbf{\mstd{97.4}{0.2}} \\

& GATE-AD (GCN)
& \mstd{97.6}{0.2} & \mstd{96.4}{0.2}
& \textbf{\mstd{92.8}{0.5}} & \mstd{97.6}{0.1}
& \mstd{82.9}{0.8} & \mstd{96.5}{0.1} \\

& Transformer Encoder (TE)
& \mstd{97.1}{1.0} & \mstd{96.5}{0.2}
& \mstd{92.1}{1.0} & \mstd{97.5}{0.2}
& \mstd{83.4}{1.0} & \mstd{97.0}{0.1} \\

\addlinespace[1pt]

& GATE-AD AE
& \mstd{94.2}{0.3} & \mstd{95.1}{0.1}
& \mstd{82.3}{1.2} & \mstd{96.5}{0.2}
& \mstd{57.8}{1.1} & \mstd{94.4}{0.3} \\

& MAE
& \mstd{66.2}{3.2} & \mstd{64.4}{5.0}
& \mstd{62.2}{0.7} & \mstd{71.4}{0.5}
& \mstd{49.4}{3.9} & \mstd{86.3}{4.7} \\

\midrule

\multirow{5}{*}{\textbf{4}}
& GATE-AD (GAT)
& \textbf{\mstd{97.9}{0.2}} & \textbf{\mstd{96.9}{0.1}}
& \textbf{\mstd{94.4}{0.2}} & \textbf{\mstd{98.3}{0.1}}
& \textbf{\mstd{84.5}{0.7}} & \textbf{\mstd{97.5}{0.2}} \\

& GATE-AD (GCN)
& \mstd{97.7}{0.2} & \mstd{96.7}{0.1}
& \mstd{94.0}{0.2} & \mstd{98.1}{0.1}
& \mstd{83.3}{0.8} & \mstd{97.3}{0.1} \\

& Transformer Encoder (TE)
& \mstd{97.4}{0.6} & \mstd{96.8}{0.2}
& \mstd{94.1}{0.6} & \mstd{98.2}{0.1}
& \mstd{83.6}{0.4} & \textbf{\mstd{97.5}{0.1}} \\

\addlinespace[1pt]

& GATE-AD AE
& \mstd{94.9}{0.4} & \mstd{95.7}{0.1}
& \mstd{87.8}{0.3} & \mstd{97.0}{0.1}
& \mstd{59.8}{3.4} & \mstd{95.0}{0.2} \\

& MAE
& \mstd{70.2}{0.8} & \mstd{66.2}{3.6}
& \mstd{64.3}{0.7} & \mstd{72.5}{0.6}
& \mstd{44.5}{4.7} & \mstd{84.9}{3.3} \\

\midrule

\multirow{5}{*}{\textbf{8}}
& GATE-AD (GAT)
& \textbf{\mstd{98.2}{0.1}} & \textbf{\mstd{97.0}{0.1}}
& \textbf{\mstd{95.2}{0.2}} & \textbf{\mstd{98.3}{0.1}}
& \textbf{\mstd{87.3}{0.8}} & \textbf{\mstd{97.8}{0.1}} \\

& GATE-AD (GCN)
& \mstd{98.1}{0.2} & \mstd{96.9}{0.2}
& \mstd{95.1}{0.2} & \textbf{\mstd{98.3}{0.1}}
& \mstd{87.2}{0.7} & \mstd{97.7}{0.1} \\

& Transformer Encoder (TE)
& \mstd{97.7}{0.3} & \mstd{96.9}{0.2}
& \mstd{94.5}{0.5} & \textbf{\mstd{98.3}{0.1}}
& \mstd{87.1}{0.4} & \mstd{97.7}{0.1} \\

\addlinespace[1pt]

& GATE-AD AE
& \mstd{95.9}{0.5} & \mstd{95.6}{0.1}
& \mstd{89.7}{0.4} & \mstd{97.2}{0.2}
& \mstd{65.2}{0.9} & \mstd{95.4}{0.2} \\

& MAE
& \mstd{72.2}{0.1} & \mstd{66.2}{0.1}
& \mstd{62.3}{0.1} & \mstd{70.8}{1.2}
& \mstd{46.4}{2.0} & \mstd{77.3}{10.7} \\

\bottomrule
\end{tabular}
\end{table*}

\paragraph{Discussion.}\
From Table~\ref{tab:architectural_ablation}, the following key observations can be made: a) GAT, GCN, and TE achieve relatively close performance on MVTec AD and VisA, where the considered relational encoders operate close to the performance ceiling. Hence, the results do not indicate that \textbf{GAT alone drives the performance gains}; instead, they further support the importance of relational modeling within the proposed formulation. b) GAT shows its clearest advantage in the more challenging 1-shot MPDD setting. Unlike the global TE with $O(N^2d)$ pairwise attention, GAT operates over the fixed $K=8$ local graph with $O(KNd)$ complexity, i.e., linear in the number of patches for fixed $K$. Together with Appendix~\ref{sec:efficiency}, where TE has substantially more parameters and FLOPs and requires approximately an order of magnitude longer per-class training, this supports GAT as the more favorable accuracy-efficiency choice, while the comparable accuracy of the relational encoders on MVTec AD and VisA further supports the underlying relational formulation. c) Regarding the reconstruction formulation, replacing RA with an explicit graph decoder (GATE-AD AE) leads to a substantial reduction in image-level detection performance, particularly on VisA and MPDD, indicating that \textbf{representation-aligned latent recovery is more effective than direct feature reconstruction} in the considered FS-IVAD setting. d) The considerably lower performance of MAE further shows that conventional masked reconstruction of the input feature tokens is not sufficient for effective normality modeling from very few normal samples. Overall, these results support the adopted architectural formulation, where \textbf{sparse relational encoding is combined with latent representation alignment} rather than conventional autoencoding reconstruction.

\subsection{Time performance}

\paragraph{Reported performance and discussion.}\
Table~\ref{tab:runtime_same_gpu} reports the inference efficiency of GATE-AD and representative FS-IVAD methods under the $1$-shot setting on MVTec AD. All methods are evaluated on the same NVIDIA RTX~6000 GPU with batch size $1$ and their official configurations; latency, throughput, and peak memory are averaged over three repetitions after $25$ warm-up images per class. GATE-AD achieves the \textbf{lowest inference latency} ($118.02\pm9.60$ ms/image) and the \textbf{highest throughput} ($8.47$ images/s), corresponding to approximately $22.5\%$ lower latency and $28.9\%$ higher throughput than the next-fastest method, FastRecon. Its peak memory usage is $1157.7$ MiB, remaining below AnomalyDINO, FoundAD, and APRIL-GAN. This efficiency is consistent with the sparse GAT design, whose computational cost after the frozen backbone scales linearly with the number of patches for a fixed neighborhood size. 

\begin{table}[t]
\centering
\scriptsize
\setlength{\tabcolsep}{6pt}
\renewcommand{\arraystretch}{1.0}
\caption{Same-GPU inference efficiency on MVTec AD under the $1$-shot setting.}
\label{tab:runtime_same_gpu}
\begin{tabular}{lccc}
\toprule
\textbf{Method} &
\textbf{Latency (ms/image)} &
\textbf{Throughput (images/s)} &
\textbf{Peak memory (MiB)} \\
\midrule
AnomalyDINO \citep{damm2025anomalydino}
& $233.58\pm119.72$ & $4.28$ & $2197.7$ \\
FoundAD \citep{zhai2025foundation}
& $253.90\pm9.98$ & $3.94$ & $1319.8$ \\
FastRecon \citep{fang2023fastrecon}
& $152.32\pm36.18$ & $6.57$ & $\mathbf{434.9}$ \\
WinCLIP+ \citep{jeong2023winclip}
& $391.82\pm96.43$ & $2.55$ & $646.7$ \\
APRIL-GAN \citep{chen2023april}
& $625.28\pm248.79$ & $1.60$ & $2799.9$ \\
GATE-AD
& $\mathbf{118.02\pm9.60}$ & $\mathbf{8.47}$ & $1157.7$ \\
\bottomrule
\end{tabular}
\end{table}

\section{Conclusion}
\label{sec:conclusion}

This paper introduced GATE-AD, a Few-Shot Industrial Visual Anomaly Detection (FS-IVAD) framework that models normality through masked, representation-aligned relational reconstruction over frozen ViT patch tokens. Rather than assessing patches independently, GATE-AD reconstructs each patch from its local spatial context using a GAT encoder, with representation alignment, SCE-based scoring, masking, and attention dropout supporting robust learning in the low-shot regime. Extensive evaluation on MVTec AD, VisA, and MPDD demonstrates strong detection and localization performance across $1$- to $8$-shot settings, achieving the highest image-level AUROC in $10/12$ settings and AP in $8/12$ settings under the unified evaluation protocol, while also attaining the lowest inference latency among the compared methods. Ablation studies further show that the gains arise from the joint formulation of relational modeling, representation alignment, and low-shot regularization. Future work will extend the current local formulation toward multi-scale or global relational reasoning for broader structural anomalies and continual adaptation.

\begin{ack}
\textbf{Funding.}
This work has received funding from the European Union's
Horizon Europe research and innovation programme under
Grant Agreement No. 101189557, project TORNADO
(foundaTion mOdels for Robots that haNdle smAll,
soft and Deformable Objects).

The views and opinions expressed in this paper are
those of the authors only and do not necessarily
reflect those of the European Union or the
European Commission.

\textbf{Competing interests.}
The authors declare no competing interests.
\end{ack}

\bibliographystyle{plainnat}
\bibliography{bibliography}

\appendix


\section{Hyper-parameter tuning}
\label{sec:hparam}

This section provides a detailed analysis of the staged hyper-parameter tuning conducted to determine suitable and robust configurations for GATE-AD. A sequential tuning strategy is followed, where each hyper-parameter is examined while the remaining settings are kept fixed. Specifically, the analysis considers: a) input image resolution, b) backbone network, c) number of last transformer layers, d) graph neighborhood size, e) number of attentional layers, f) cardinality of the GAT layers, g) cardinality of the latent representation, h) objective function, i) masking ratio, j) dropout rate, and k) anomalous patches ratio. Most analyses are conducted on MVTec AD and VisA, while the backbone and graph neighborhood size are additionally examined on MPDD. A separate training-budget analysis is also conducted on MVTec AD under the $1$-shot setting. The resulting dataset-specific configurations are used for the component and architectural analyses of Section~\ref{ssec:ablation}; the controlled cross-method comparison follows the unified protocol defined in Section~\ref{ssec:setup}.

In the dataset-specific configurations, the backbone is DINOv2 ViT-L/14 on MVTec AD (with $\beta = 488$), DINOv3 ViT-B/16 on VisA, and DINOv3 ViT-L/16 on MPDD (see Appendix~\ref{subsec:backbone_robustness}). The image-level score uses $\epsilon = 0.025$ on MVTec AD and $\epsilon = 0.01$ on VisA, following the top-$k$ convention of FoundAD~\citep{zhai2025foundation}. On MPDD, max pooling is used, following RegAD~\citep{huang2022registration}, GraphCore~\citep{xie2023pushing}, and FastRecon~\citep{fang2023fastrecon}. Pixel-level metrics are computed directly from the anomaly maps and therefore do not depend on the pooling rule. Since the benchmarks provide no separate validation split, the dataset-specific settings are selected on the benchmark test sets. For this reason, all cross-method comparisons follow the unified protocol.

\subsection{Input image resolution}
\label{subsec:resolution}

Fig. \ref{fig:abla_resolution_mvtec_visa} illustrates how the input image resolution $\beta$ (Section \ref{subsec:feat}) affects the performance of GATE-AD across the MVTec AD and VisA benchmarks. From the presented results it can be seen that resolutions of $512$ and $768$ pixels exhibit comparable performance. On the contrary, a resolution of $256$ pixels results in noticeably lower I-AUROC and PRO scores across all shot counts. Additionally, while $768$ pixels occasionally introduce slightly higher PRO scores in certain settings, the improvement over $512$ pixels is marginal. To this end, a resolution of 512 pixels is selected for the DINOv3 configurations, providing a balance between detection accuracy and computational cost.

\subsection{Backbone}
\label{subsec:backbone_robustness}

Fig. \ref{fig:ablation_backbone_mvtec_visa_mpdd} illustrates how the backbone network (Section \ref{subsec:feat}) affects the performance of GATE-AD across the MVTec AD, VisA, and MPDD benchmarks. From the presented results it can be seen that no single backbone or distillation size is proven to be universally optimal across all datasets. For MVTec AD, DINOv2 ViT-L/14 generally yields the best performance. For VisA, DINOv3 ViT-B/16 is proven to be the most effective. For MPDD, DINOv3 ViT-L/16 exhibits the highest accuracy.

\subsection{Graph neighborhood size}
\label{subsec:graph_neighborhood}

Fig.~\ref{fig:abla_graph_neighborhood} illustrates how the graph neighborhood
size $K=|\mathcal{N}(i)|$ (Section~\ref{subsec:GAT}) affects the performance
of GATE-AD across the MVTec AD, VisA, and MPDD benchmarks. From the presented
results it can be seen that GATE-AD remains generally robust across the
considered neighborhood sizes. In particular, $K=8$ consistently achieves the
highest AUPRO across all datasets and shot settings, while also yielding the
highest I-AUROC across all shot settings on the more challenging MPDD
benchmark. On the contrary, $K=4$ excludes diagonal neighbors and therefore
restricts the available local context, while increasing the neighborhood to
$K=24$ does not provide a consistent performance improvement and approximately
triples the number of graph edges and message-passing operations with respect
to $K=8$. To this end, $K=8$ is selected as the default graph connectivity.

\begin{figure}[t!]
\centering
\resizebox{0.95\textwidth}{!}{%
\begin{tikzpicture}

\pgfplotsset{
  k4/.style={fill=blue,            draw=black, line width=0.2pt},
  k8/.style={fill=red,             draw=black, line width=0.2pt},
  k24/.style={fill=green!70!black, draw=black, line width=0.2pt},
}

\newcommand{\kitem}[2]{%
  \tikz[baseline=-0.6ex]\filldraw[fill=#1,draw=black]
  (0,0) rectangle (1.8mm,1.2mm);\,\scriptsize #2%
}

\begin{groupplot}[
    group style={
        group size=3 by 2,
        horizontal sep=0.045\textwidth,
        vertical sep=0.055\textwidth
    },
    width=0.35\textwidth,
    height=0.25\textwidth,
    ybar,
    /pgf/bar width=2.8pt,
    enlarge x limits=0.30,
    xmode=log,
    log basis x=2,
    xtick={1,2,4,8},
    xticklabels={1,2,4,8},
    xmin=0.9, xmax=8.5,
    ymajorgrids=true,
    xmajorgrids=true,
    tick label style={font=\scriptsize},
    label style={font=\scriptsize},
    title style={font=\scriptsize},
]

\nextgroupplot[
  title={MVTec AD},
  ylabel={I-AUROC (\%)},
  axis background/.style={fill=green!6},
  ymin=97.5, ymax=98.35
]
\addplot+[k4] coordinates {
  (1,97.870) (2,98.077) (4,98.030) (8,98.122)
};
\addplot+[k8] coordinates {
  (1,97.700) (2,98.000) (4,97.900) (8,98.200)
};
\addplot+[k24] coordinates {
  (1,97.888) (2,98.063) (4,98.040) (8,98.111)
};

\nextgroupplot[
  title={VisA},
  axis background/.style={fill=blue!6},
  ymin=92.5, ymax=96.0
]
\addplot+[k4] coordinates {
  (1,93.490) (2,94.010) (4,95.324) (8,95.709)
};
\addplot+[k8] coordinates {
  (1,93.500) (2,92.800) (4,94.400) (8,95.200)
};
\addplot+[k24] coordinates {
  (1,93.504) (2,94.033) (4,95.292) (8,95.681)
};

\nextgroupplot[
  title={MPDD},
  axis background/.style={fill=orange!6},
  ymin=82.0, ymax=89.0
]
\addplot+[k4] coordinates {
  (1,83.436) (2,83.131) (4,83.086) (8,87.040)
};
\addplot+[k8] coordinates {
  (1,84.200) (2,84.300) (4,84.100) (8,87.800)
};
\addplot+[k24] coordinates {
  (1,83.261) (2,82.938) (4,83.135) (8,86.977)
};

\nextgroupplot[
  ylabel={AUPRO (\%)},
  xlabel={Shots},
  axis background/.style={fill=green!6},
  ymin=92.4, ymax=94.0
]
\addplot+[k4] coordinates {
  (1,92.625) (2,93.116) (4,93.310) (8,93.524)
};
\addplot+[k8] coordinates {
  (1,93.000) (2,93.400) (4,93.500) (8,93.700)
};
\addplot+[k24] coordinates {
  (1,92.621) (2,93.124) (4,93.303) (8,93.517)
};

\nextgroupplot[
  xlabel={Shots},
  axis background/.style={fill=blue!6},
  ymin=92.4, ymax=94.9
]
\addplot+[k4] coordinates {
  (1,92.735) (2,93.410) (4,94.337) (8,94.353)
};
\addplot+[k8] coordinates {
  (1,93.200) (2,93.500) (4,94.500) (8,94.600)
};
\addplot+[k24] coordinates {
  (1,92.681) (2,93.389) (4,94.305) (8,94.315)
};

\nextgroupplot[
  xlabel={Shots},
  axis background/.style={fill=orange!6},
  ymin=93.0, ymax=94.7
]
\addplot+[k4] coordinates {
  (1,93.264) (2,93.320) (4,93.582) (8,94.362)
};
\addplot+[k8] coordinates {
  (1,93.400) (2,93.400) (4,93.700) (8,94.400)
};
\addplot+[k24] coordinates {
  (1,93.274) (2,93.297) (4,93.587) (8,94.382)
};

\end{groupplot}

\node[anchor=south, yshift=2mm]
at (current bounding box.north) {%
\scriptsize
Graph neighborhood size:\,
\kitem{blue}{$K=4$}\quad
\kitem{red}{$K=8$}\quad
\kitem{green!70!black}{$K=24$}
};

\end{tikzpicture}%
}

\caption{Effect of the graph neighborhood size on the MVTec AD, VisA, and MPDD benchmarks.}
\label{fig:abla_graph_neighborhood}
\end{figure}

\subsection{Number of last transformer layers}
\label{subsec:last_layers}

\begin{figure}[t!]
\centering
\resizebox{0.80\textwidth}{!}{%
\begin{tikzpicture}
\pgfplotsset{compat=1.18}
\pgfplotsset{
  l1/.style={fill=blue!75!black,   draw=black, line width=0.2pt},
  l4/.style={fill=orange!85!black, draw=black, line width=0.2pt},
  l6/.style={fill=green!70!black,  draw=black, line width=0.2pt},
  l8/.style={fill=purple!80!black, draw=black, line width=0.2pt},
  l12/.style={fill=black,          draw=black, line width=0.2pt},
}
\newcommand{\xx}{nan}

\newcommand{\layitem}[2]{%
  \tikz[baseline=-0.6ex]\filldraw[fill=#1,draw=black] (0,0) rectangle (1.8mm,1.2mm);\,\scriptsize #2%
}

\begin{groupplot}[
  group style={group size=2 by 1, horizontal sep=0.05\textwidth},
  width=0.56\textwidth,
  height=0.35\textwidth,
  ybar,
  /pgf/bar width=2.2pt,
  enlarge x limits=0.30,
  xmode=log,
  log basis x=2,
  xtick={1,2,4,8},
  xticklabels={1,2,4,8},
  xmin=0.9, xmax=8.5,
  ymajorgrids=true,
  xmajorgrids=true,
  tick label style={font=\scriptsize},
  label style={font=\scriptsize},
  title style={font=\scriptsize},
  xlabel={Shots},
]

\nextgroupplot[
  title={MVTec},
  ylabel={Image AUROC (\%)},
  axis background/.style={fill=green!6},
  ymin=92, ymax=99
]
\addplot+[l1]  coordinates {(1,92.8933) (2,94.0631) (4,95.1205) (8,95.5072)};
\addplot+[l4]  coordinates {(1,96.9365) (2,97.3560) (4,97.2775) (8,97.6200)};
\addplot+[l6]  coordinates {(1,97.6112) (2,98.0004) (4,97.9224) (8,98.1509)};
\addplot+[l8]  coordinates {(1,97.7897) (2,98.1260) (4,97.9742) (8,98.1950)};
\addplot+[l12] coordinates {(1,97.7340) (2,98.2166) (4,98.1802) (8,98.2333)};

\nextgroupplot[
  title={VisA},
  ylabel={},
  axis background/.style={fill=blue!6},
  ymin=82, ymax=96
]
\addplot+[l1]  coordinates {(1,83.6986) (2,84.8078) (4,87.2790) (8,89.4214)};
\addplot+[l4]  coordinates {(1,91.2707) (2,90.5306) (4,92.6628) (8,93.9411)};
\addplot+[l6]  coordinates {(1,93.4412) (2,92.5995) (4,94.1372) (8,95.1938)};
\addplot+[l8]  coordinates {(1,93.8020) (2,93.6020) (4,94.8424) (8,95.6382)};
\addplot+[l12] coordinates {(1,91.8214) (2,92.4590) (4,94.1891) (8,94.7107)};

\end{groupplot}

\node[anchor=south, yshift=2mm] at (current bounding box.north) {%
\scriptsize
Number of last transformer layers:\,
\layitem{blue!75!black}{1}\quad
\layitem{orange!85!black}{4}\quad
\layitem{green!70!black}{6}\quad
\layitem{purple!80!black}{8}\quad
\layitem{black}{12}
};

\end{tikzpicture}%
}
\caption{Effect of the number of last transformer layers on the MVTec and VisA benchmarks.}
\label{fig:lastlayer_ablation_mvtec_visa_mpdd}
\end{figure}
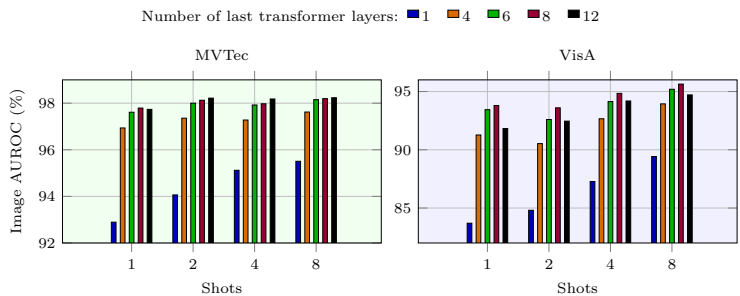

Fig. \ref{fig:lastlayer_ablation_mvtec_visa_mpdd} illustrates how the number of the last transformer layers $\zeta$ (Section \ref{subsec:feat}) affects the performance of GATE-AD across the MVTec AD and VisA benchmarks. From the presented results it can be seen that increasing the number of layers generally improves Image AUROC, until it stabilizes or slightly declines when including too many layers. In particular, averaging patch tokens from the last $\zeta=8$ layer leads to the highest overall performance.

\begin{figure}[t]
\centering
\resizebox{0.75\textwidth}{!}{%
\begin{tikzpicture}
\pgfplotsset{
  c1/.style={fill=blue,           draw=black, line width=0.2pt},
  c2/.style={fill=red,            draw=black, line width=0.2pt},
  c3/.style={fill=green!60!black, draw=black, line width=0.2pt},
}

\newcommand{\resitem}[2]{%
  \tikz[baseline=-0.6ex]\filldraw[fill=#1,draw=black] (0,0) rectangle (1.8mm,1.2mm);\,\scriptsize #2%
}

\begin{groupplot}[
    group style={group size=2 by 2, horizontal sep=0.06\textwidth, vertical sep=0.06\textwidth},
    width=0.49\textwidth,
    height=0.30\textwidth,
    ybar,
    /pgf/bar width=2.2pt,
    enlarge x limits=0.30,
    xmode=log,
    log basis x=2,
    xtick={1,2,4,8},
    xticklabels={1,2,4,8},
    xmin=0.9, xmax=8.5,
    ymajorgrids=true,
    xmajorgrids=true,
    tick label style={font=\scriptsize},
    label style={font=\scriptsize},
    title style={font=\scriptsize},
]

\nextgroupplot[
  title={MVTec AD},
  ylabel={I-AUROC (\%)},
  axis background/.style={fill=green!6},
  ymin=92.5, ymax=98.2
]
\addplot+[c1] coordinates {(1,92.97) (2,94.75) (4,94.81) (8,95.22)};
\addplot+[c2] coordinates {(1,97.34) (2,97.72) (4,97.75) (8,97.86)};
\addplot+[c3] coordinates {(1,97.58) (2,97.56) (4,97.64) (8,97.72)};

\nextgroupplot[
  title={VisA},
  axis background/.style={fill=blue!6},
  ymin=88.0, ymax=96.0
]
\addplot+[c1] coordinates {(1,88.72) (2,90.46) (4,91.84) (8,93.59)};
\addplot+[c2] coordinates {(1,93.48) (2,92.87) (4,94.27) (8,95.10)};
\addplot+[c3] coordinates {(1,93.44) (2,92.92) (4,94.02) (8,95.00)};

\nextgroupplot[
  ylabel={PRO (\%)},
  xlabel={Shots},
  axis background/.style={fill=green!6},
  ymin=91.0, ymax=95.2
]
\addplot+[c1] coordinates {(1,91.70) (2,92.40) (4,92.50) (8,92.58)};
\addplot+[c2] coordinates {(1,93.70) (2,94.19) (4,94.45) (8,94.60)};
\addplot+[c3] coordinates {(1,94.02) (2,94.51) (4,94.81) (8,94.92)};

\nextgroupplot[
  xlabel={Shots},
  axis background/.style={fill=blue!6},
  ymin=88.5, ymax=95.5
]
\addplot+[c1] coordinates {(1,88.90) (2,89.46) (4,90.75) (8,91.22)};
\addplot+[c2] coordinates {(1,93.58) (2,93.83) (4,94.63) (8,94.70)};
\addplot+[c3] coordinates {(1,94.03) (2,94.35) (4,95.07) (8,95.08)};

\end{groupplot}

\node[anchor=south, yshift=2mm] at (current bounding box.north) {%
\scriptsize
Input image resolution:\,
\resitem{blue}{256}\quad
\resitem{red}{512}\quad
\resitem{green!60!black}{768}
};

\end{tikzpicture}%
}
\caption{Effect of the input image resolution on the MVTec AD and VisA benchmarks.}
\label{fig:abla_resolution_mvtec_visa}
\end{figure}
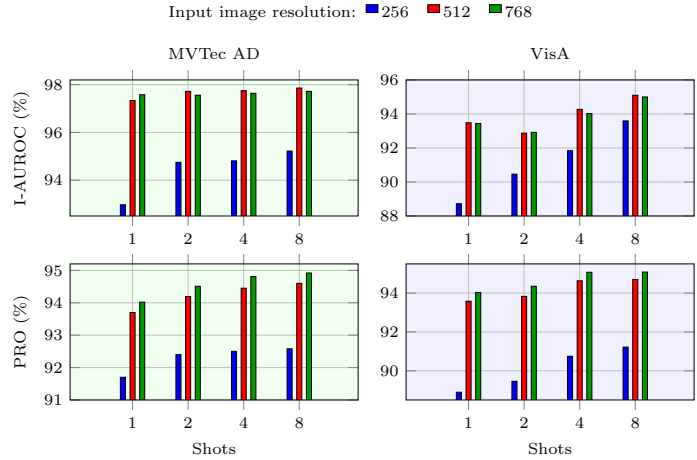

\begin{figure}[h!]
\centering
\resizebox{0.80\textwidth}{!}{%
\begin{tikzpicture}
\pgfplotsset{
  d3s/.style={fill=orange!85!black, draw=black, line width=0.2pt},
  d3b/.style={fill=purple!80!black, draw=black, line width=0.2pt},
  d3l/.style={fill=black,            draw=black, line width=0.2pt},
  d2s/.style={fill=blue,             draw=black, line width=0.2pt},
  d2b/.style={fill=red,              draw=black, line width=0.2pt},
  d2l/.style={fill=green!70!black,   draw=black, line width=0.2pt},
}

\newcommand{\bbitem}[2]{%
  \tikz[baseline=-0.6ex]\filldraw[fill=#1,draw=black] (0,0) rectangle (1.8mm,1.2mm);\,\scriptsize #2%
}

\begin{groupplot}[
    group style={group size=2 by 1, horizontal sep=0.05\textwidth},
    width=0.56\textwidth,
    height=0.35\textwidth,
    ybar,
    /pgf/bar width=2.2pt,
    enlarge x limits=0.30,
    xmode=log,
    log basis x=2,
    xtick={1,2,4,8},
    xticklabels={1,2,4,8},
    xmin=0.9, xmax=8.5,
    ymajorgrids=true,
    xmajorgrids=true,
    tick label style={font=\scriptsize},
    label style={font=\scriptsize},
    title style={font=\scriptsize},
    xlabel={Shots},
]

\nextgroupplot[
  title={MVTec AD},
  ylabel={Image AUROC (\%)},
  axis background/.style={fill=green!6},
  ymin=95, ymax=98.4
]
\addplot+[d3s] coordinates {(1,96.7319) (2,97.1778) (4,97.4971) (8,97.4969)};
\addplot+[d3b] coordinates {(1,97.2082) (2,97.7372) (4,97.7802) (8,97.9470)};
\addplot+[d3l] coordinates {(1,97.1049) (2,97.0880) (4,97.1624) (8,98.0683)};
\addplot+[d2s] coordinates {(1,95.1078) (2,95.9236) (4,96.4497) (8,96.2108)};
\addplot+[d2b] coordinates {(1,96.3884) (2,97.2075) (4,97.1488) (8,97.2930)};
\addplot+[d2l] coordinates {(1,97.5815) (2,97.8962) (4,97.8150) (8,98.2182)};

\nextgroupplot[
  title={VisA},
  ylabel={},
  axis background/.style={fill=blue!6},
  ymin=85, ymax=96
]
\addplot+[d3s] coordinates {(1,90.3814) (2,90.1897) (4,91.7445) (8,93.2999)};
\addplot+[d3b] coordinates {(1,93.6860) (2,93.4395) (4,94.6816) (8,95.4514)};
\addplot+[d3l] coordinates {(1,88.9881) (2,88.4548) (4,91.4216) (8,94.3743)};
\addplot+[d2s] coordinates {(1,90.3423) (2,89.8410) (4,91.6051) (8,92.7553)};
\addplot+[d2b] coordinates {(1,92.8739) (2,92.7504) (4,94.1622) (8,94.8315)};
\addplot+[d2l] coordinates {(1,88.7729) (2,88.7139) (4,90.9438) (8,92.6064)};

\end{groupplot}

\node[anchor=south, yshift=2mm, align=center] at (current bounding box.north) {%
\scriptsize
Backbone:\,
\bbitem{orange!85!black}{DINOv3-S/16}\quad
\bbitem{purple!80!black}{DINOv3-B/16}\quad
\bbitem{black}{DINOv3-L/16}\\[1mm]
\bbitem{blue}{DINOv2-S/14}\quad
\bbitem{red}{DINOv2-B/14}\quad
\bbitem{green!70!black}{DINOv2-L/14}
};

\begin{axis}[
  at={($(group c1r1.south)!0.5!(group c2r1.south)+(0,-0.08\textwidth)$)},
  anchor=north,
  width=0.56\textwidth,
  height=0.35\textwidth,
  ybar,
  /pgf/bar width=2.2pt,
  enlarge x limits=0.30,
  xmode=log,
  log basis x=2,
  xtick={1,2,4,8},
  xticklabels={1,2,4,8},
  xmin=0.9, xmax=8.5,
  ymajorgrids=true,
  xmajorgrids=true,
  tick label style={font=\scriptsize},
  label style={font=\scriptsize},
  title style={font=\scriptsize},
  title={MPDD},
  ylabel={Image AUROC (\%)},
  xlabel={Shots},
  axis background/.style={fill=orange!6},
  ymin=57, ymax=90
]
\addplot+[d3s] coordinates {(1,67.1015) (2,68.3683) (4,65.7423) (8,70.5862)};
\addplot+[d3b] coordinates {(1,70.7991) (2,69.6523) (4,71.4986) (8,78.4820)};
\addplot+[d3l] coordinates {(1,84.6001) (2,84.6247) (4,84.8851) (8,88.2185)};
\addplot+[d2s] coordinates {(1,74.2408) (2,72.5877) (4,71.1910) (8,74.1177)};
\addplot+[d2b] coordinates {(1,73.2965) (2,72.2120) (4,73.0143) (8,80.2310)};
\addplot+[d2l] coordinates {(1,76.8619) (2,76.0243) (4,77.7238) (8,84.4003)};
\end{axis}

\end{tikzpicture}%
}
\caption{Effect of the backbone on the MVTec AD, VisA, and MPDD benchmarks.}
\label{fig:ablation_backbone_mvtec_visa_mpdd}
\end{figure}
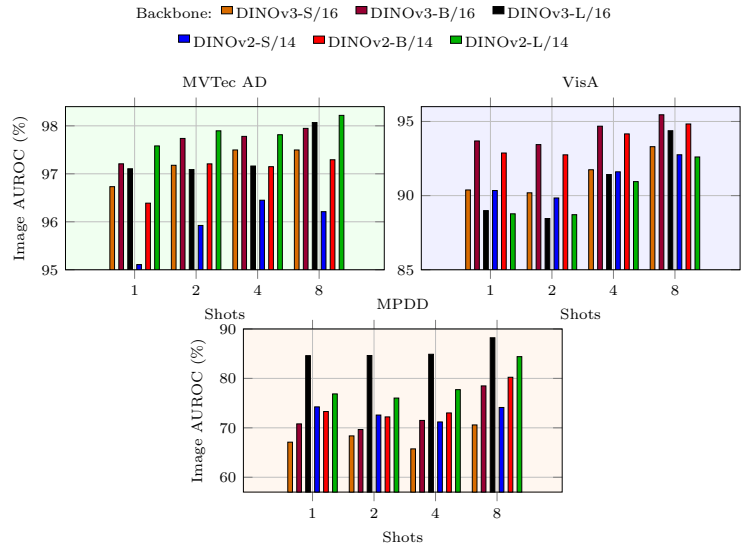

\subsection{Number of attentional layers}
\label{subsec:attn_layers}

Fig. \ref{fig:abla_layers_mvtec_visa} illustrates how the number of the number of attentional layers $R$ (Section \ref{subsec:GAT}) affects the performance of GATE-AD across the MVTec AD and VisA benchmarks. From the presented results it can be seen that setting $R=2$ or $R=3$ results in similar performance, though using $3$ layers is slightly better in terms of achieving the highest I-AUROC and PRO scores. Further increasing the network depth ($R=4$) does not necessarily yield better results and, in some cases, leads to slight performance decrease. Based on the above observations, $R=3$ is used to ensure robust encoding of local relations.

\begin{figure}[h!]
\centering
\resizebox{0.80\textwidth}{!}{%
\begin{tikzpicture}
\pgfplotsset{
  c1/.style={fill=blue,           draw=black, line width=0.2pt},
  c2/.style={fill=red,            draw=black, line width=0.2pt},
  c3/.style={fill=green!70!black, draw=black, line width=0.2pt},
  c4/.style={fill=black,          draw=black, line width=0.2pt},
}

\newcommand{\layeritem}[2]{%
  \tikz[baseline=-0.6ex]\filldraw[fill=#1,draw=black] (0,0) rectangle (1.8mm,1.2mm);\,\scriptsize #2%
}

\begin{groupplot}[
    group style={group size=2 by 2, horizontal sep=0.06\textwidth, vertical sep=0.06\textwidth},
    width=0.49\textwidth,
    height=0.30\textwidth,
    ybar,
    /pgf/bar width=2.2pt,
    enlarge x limits=0.30,
    xmode=log,
    log basis x=2,
    xtick={1,2,4,8},
    xticklabels={1,2,4,8},
    xmin=0.9, xmax=8.5,
    ymajorgrids=true,
    xmajorgrids=true,
    tick label style={font=\scriptsize},
    label style={font=\scriptsize},
    title style={font=\scriptsize},
]

\nextgroupplot[
  title={MVTec AD},
  ylabel={I-AUROC (\%)},
  axis background/.style={fill=green!6},
  ymin=96.8, ymax=98.05
]
\addplot+[c1] coordinates {(1,97.2) (2,97.6) (4,97.7) (8,97.7)};
\addplot+[c2] coordinates {(1,97.2) (2,97.7) (4,97.7) (8,97.8)};
\addplot+[c3] coordinates {(1,97.2) (2,97.8) (4,97.8) (8,97.8)};
\addplot+[c4] coordinates {(1,97.0) (2,97.6) (4,97.6) (8,97.5)};

\nextgroupplot[
  title={VisA},
  axis background/.style={fill=blue!6},
  ymin=92.0, ymax=95.7
]
\addplot+[c1] coordinates {(1,93.1) (2,92.3) (4,94.0) (8,94.9)};
\addplot+[c2] coordinates {(1,93.7) (2,93.4) (4,94.6) (8,95.4)};
\addplot+[c3] coordinates {(1,93.5) (2,93.2) (4,94.6) (8,95.3)};
\addplot+[c4] coordinates {(1,93.4) (2,92.9) (4,94.4) (8,95.3)};

\nextgroupplot[
  ylabel={PRO (\%)},
  xlabel={Shots},
  axis background/.style={fill=green!6},
  ymin=93.4, ymax=94.8
]
\addplot+[c1] coordinates {(1,93.6) (2,94.1) (4,94.3) (8,94.5)};
\addplot+[c2] coordinates {(1,93.6) (2,94.2) (4,94.4) (8,94.6)};
\addplot+[c3] coordinates {(1,93.7) (2,94.2) (4,94.5) (8,94.6)};
\addplot+[c4] coordinates {(1,93.5) (2,94.1) (4,94.4) (8,94.5)};

\nextgroupplot[
  xlabel={Shots},
  axis background/.style={fill=blue!6},
  ymin=92.6, ymax=95.0
]
\addplot+[c1] coordinates {(1,92.7) (2,93.1) (4,94.0) (8,94.1)};
\addplot+[c2] coordinates {(1,93.4) (2,93.6) (4,94.4) (8,94.5)};
\addplot+[c3] coordinates {(1,93.2) (2,93.6) (4,94.6) (8,94.7)};
\addplot+[c4] coordinates {(1,93.0) (2,93.4) (4,94.3) (8,94.4)};

\end{groupplot}

\node[anchor=south, yshift=2mm] at (current bounding box.north) {%
\scriptsize
Number of attentional layers:\,
\layeritem{blue}{1}\quad
\layeritem{red}{2}\quad
\layeritem{green!70!black}{3}\quad
\layeritem{black}{4}
};

\end{tikzpicture}%
}
\caption{Effect of the number of attentional layers on the MVTec AD and VisA benchmarks.}
\label{fig:abla_layers_mvtec_visa}
\end{figure}
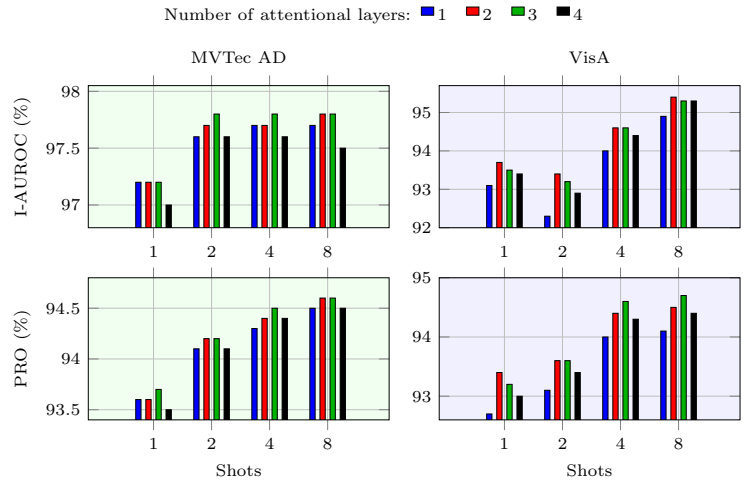

\begin{figure}[h!]
\centering
\resizebox{0.80\textwidth}{!}{%
\begin{tikzpicture}
\pgfplotsset{
  c1/.style={fill=red,             draw=black, line width=0.2pt},      
  c2/.style={fill=green!70!black,  draw=black, line width=0.2pt},      
  c3/.style={fill=orange!85!black, draw=black, line width=0.2pt},      
  c4/.style={fill=black,           draw=black, line width=0.2pt},      
}

\newcommand{\dimitem}[2]{%
  \tikz[baseline=-0.6ex]\filldraw[fill=#1,draw=black] (0,0) rectangle (1.8mm,1.2mm);\,\scriptsize #2%
}

\begin{groupplot}[
    group style={group size=2 by 2, horizontal sep=0.06\textwidth, vertical sep=0.06\textwidth},
    width=0.49\textwidth,
    height=0.28\textwidth,
    ybar,
    /pgf/bar width=2.2pt,
    enlarge x limits=0.30,
    xmode=log,
    log basis x=2,
    xtick={1,2,4,8},
    xticklabels={1,2,4,8},
    xmin=0.9, xmax=8.5,
    ymajorgrids=true,
    xmajorgrids=true,
    tick label style={font=\scriptsize},
    label style={font=\scriptsize},
    title style={font=\scriptsize},
]

\nextgroupplot[
  title={MVTec AD},
  ylabel={I-AUROC (\%)},
  axis background/.style={fill=green!6},
  ymin=96.8, ymax=98.0
]
\addplot+[c1] coordinates {(1,97.22) (2,97.66) (4,97.76) (8,97.89)}; 
\addplot+[c2] coordinates {(1,96.99) (2,97.52) (4,97.52) (8,97.71)}; 
\addplot+[c3] coordinates {(1,96.94) (2,97.47) (4,97.56) (8,97.58)}; 
\addplot+[c4] coordinates {(1,97.21) (2,97.53) (4,97.53) (8,97.77)}; 

\nextgroupplot[
  title={VisA},
  axis background/.style={fill=blue!6},
  ymin=92.0, ymax=95.4
]
\addplot+[c1] coordinates {(1,93.52) (2,92.80) (4,94.28) (8,95.14)}; 
\addplot+[c2] coordinates {(1,93.47) (2,92.33) (4,93.90) (8,94.78)}; 
\addplot+[c3] coordinates {(1,93.17) (2,92.64) (4,94.31) (8,95.23)}; 
\addplot+[c4] coordinates {(1,93.70) (2,92.80) (4,94.14) (8,94.98)}; 

\nextgroupplot[
  ylabel={PRO (\%)},
  xlabel={Shots},
  axis background/.style={fill=green!6},
  ymin=93.5, ymax=94.85
]
\addplot+[c1] coordinates {(1,93.68) (2,94.18) (4,94.46) (8,94.60)}; 
\addplot+[c2] coordinates {(1,93.83) (2,94.32) (4,94.53) (8,94.67)}; 
\addplot+[c3] coordinates {(1,93.81) (2,94.29) (4,94.49) (8,94.60)}; 
\addplot+[c4] coordinates {(1,93.78) (2,94.29) (4,94.50) (8,94.66)}; 

\nextgroupplot[
  xlabel={Shots},
  axis background/.style={fill=blue!6},
  ymin=93.2, ymax=95.0
]
\addplot+[c1] coordinates {(1,93.53) (2,93.79) (4,94.59) (8,94.68)}; 
\addplot+[c2] coordinates {(1,93.46) (2,93.68) (4,94.48) (8,94.55)}; 
\addplot+[c3] coordinates {(1,93.51) (2,93.72) (4,94.61) (8,94.70)}; 
\addplot+[c4] coordinates {(1,93.28) (2,93.58) (4,94.45) (8,94.53)}; 

\end{groupplot}

\node[anchor=south, yshift=2mm] at (current bounding box.north) {%
\scriptsize
Cardinality of GAT layers:\,
\dimitem{red}{256}\quad
\dimitem{green!70!black}{384}\quad
\dimitem{orange!85!black}{512}\quad
\dimitem{black}{768}
};

\end{tikzpicture}%
}
\caption{Effect of the cardinality of the GAT layers on the MVTec AD and VisA benchmarks.}
\label{fig:abla_cardinality_layers}
\end{figure}
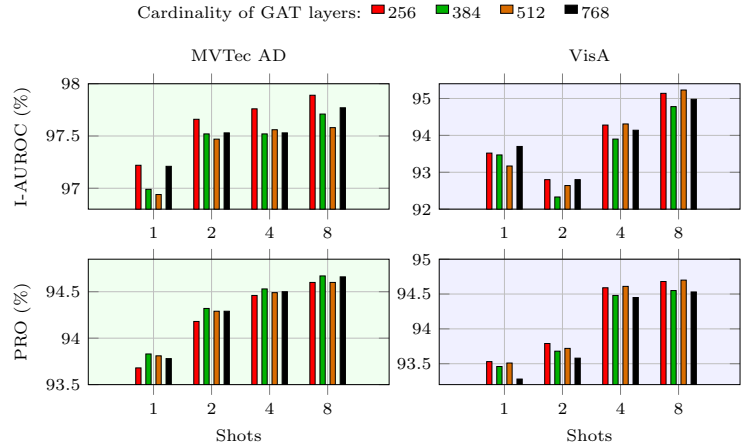

\subsection{Cardinality of GAT layers}
\label{subsec:cardinality_only}

Fig. \ref{fig:abla_cardinality_layers} illustrates how the cardinality of the GAT layers $F$ (Section \ref{subsec:GAT}) affects the performance of GATE-AD across the MVTec AD and VisA benchmarks. From the presented results it can be seen that $F=256$ is sufficient for reaching high recognition performance. Further increasing the cardinality (beyond $256$) results into negligible or inconsistent improvements in Image AUROC and PRO scores. $F=256$ is selected, since it maintains high detection accuracy, while ensuring low inference latency and reduced computational cost.

\subsection{Cardinality of latent representation}
\label{subsec:latent_and_cardinality}

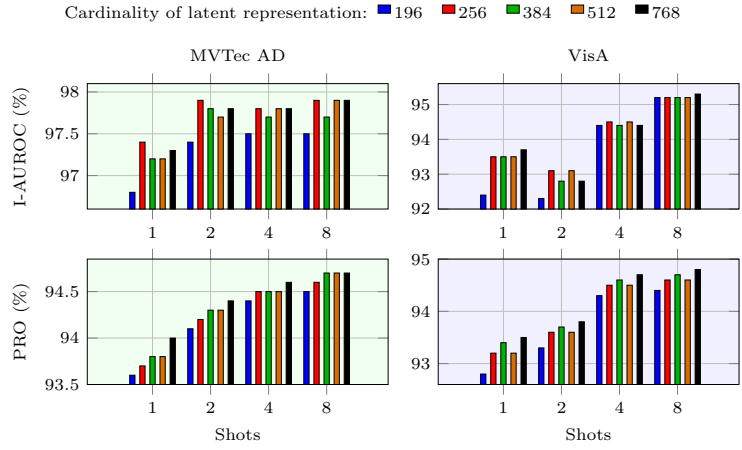
\begin{figure}[t!]
\centering
\resizebox{0.80\textwidth}{!}{%
\begin{tikzpicture}
\pgfplotsset{
  c1/.style={fill=blue,            draw=black, line width=0.2pt},
  c2/.style={fill=red,             draw=black, line width=0.2pt},
  c3/.style={fill=green!70!black,  draw=black, line width=0.2pt},
  c4/.style={fill=orange!85!black, draw=black, line width=0.2pt},
  c5/.style={fill=black,           draw=black, line width=0.2pt},
}

\newcommand{\dimitem}[2]{%
  \tikz[baseline=-0.6ex]\filldraw[fill=#1,draw=black] (0,0) rectangle (1.8mm,1.2mm);\,\scriptsize #2%
}

\begin{groupplot}[
    group style={group size=2 by 2, horizontal sep=0.06\textwidth, vertical sep=0.06\textwidth},
    width=0.49\textwidth,
    height=0.28\textwidth,
    ybar,
    /pgf/bar width=2.2pt,
    enlarge x limits=0.30,
    xmode=log,
    log basis x=2,
    xtick={1,2,4,8},
    xticklabels={1,2,4,8},
    xmin=0.9, xmax=8.5,
    ymajorgrids=true,
    xmajorgrids=true,
    tick label style={font=\scriptsize},
    label style={font=\scriptsize},
    title style={font=\scriptsize},
]

\nextgroupplot[
  title={MVTec AD},
  ylabel={I-AUROC (\%)},
  axis background/.style={fill=green!6},
  ymin=96.6, ymax=98.1
]
\addplot+[c1] coordinates {(1,96.8) (2,97.4) (4,97.5) (8,97.5)};
\addplot+[c2] coordinates {(1,97.4) (2,97.9) (4,97.8) (8,97.9)};
\addplot+[c3] coordinates {(1,97.2) (2,97.8) (4,97.7) (8,97.7)};
\addplot+[c4] coordinates {(1,97.2) (2,97.7) (4,97.8) (8,97.9)};
\addplot+[c5] coordinates {(1,97.3) (2,97.8) (4,97.8) (8,97.9)};

\nextgroupplot[
  title={VisA},
  axis background/.style={fill=blue!6},
  ymin=92.0, ymax=95.6
]
\addplot+[c1] coordinates {(1,92.4) (2,92.3) (4,94.4) (8,95.2)};
\addplot+[c2] coordinates {(1,93.5) (2,93.1) (4,94.5) (8,95.2)};
\addplot+[c3] coordinates {(1,93.5) (2,92.8) (4,94.4) (8,95.2)};
\addplot+[c4] coordinates {(1,93.5) (2,93.1) (4,94.5) (8,95.2)};
\addplot+[c5] coordinates {(1,93.7) (2,92.8) (4,94.4) (8,95.3)};

\nextgroupplot[
  ylabel={PRO (\%)},
  xlabel={Shots},
  axis background/.style={fill=green!6},
  ymin=93.5, ymax=94.85
]
\addplot+[c1] coordinates {(1,93.6) (2,94.1) (4,94.4) (8,94.5)};
\addplot+[c2] coordinates {(1,93.7) (2,94.2) (4,94.5) (8,94.6)};
\addplot+[c3] coordinates {(1,93.8) (2,94.3) (4,94.5) (8,94.7)};
\addplot+[c4] coordinates {(1,93.8) (2,94.3) (4,94.5) (8,94.7)};
\addplot+[c5] coordinates {(1,94.0) (2,94.4) (4,94.6) (8,94.7)};

\nextgroupplot[
  xlabel={Shots},
  axis background/.style={fill=blue!6},
  ymin=92.6, ymax=95.0
]
\addplot+[c1] coordinates {(1,92.8) (2,93.3) (4,94.3) (8,94.4)};
\addplot+[c2] coordinates {(1,93.2) (2,93.6) (4,94.5) (8,94.6)};
\addplot+[c3] coordinates {(1,93.4) (2,93.7) (4,94.6) (8,94.7)};
\addplot+[c4] coordinates {(1,93.2) (2,93.6) (4,94.5) (8,94.6)};
\addplot+[c5] coordinates {(1,93.5) (2,93.8) (4,94.7) (8,94.8)};

\end{groupplot}

\node[anchor=south, yshift=2mm] at (current bounding box.north) {%
\scriptsize
Cardinality of latent representation:\,
\dimitem{blue}{196}\quad
\dimitem{red}{256}\quad
\dimitem{green!70!black}{384}\quad
\dimitem{orange!85!black}{512}\quad
\dimitem{black}{768}
};

\end{tikzpicture}%
}
\caption{Effect of the cardinality of the latent representation on the MVTec AD and VisA benchmarks.}
\label{fig:abla_latentdim_mvtec_visa}
\end{figure}

Fig. \ref{fig:abla_latentdim_mvtec_visa} illustrates how the cardinality of the latent representation $f$ (Section \ref{subsec:GAT}) affects the performance of GATE-AD across the MVTec AD and VisA benchmarks. From the presented results it can be seen that setting $f=256$ or higher leads to negligible variance in image AUROC. To this end, $f$ is selected equal to $256$ for achieving a low computational cost, without sacrificing recognition performance.

\subsection{Objective function}
\label{subsec:objective}

\begin{figure}[t!]
\centering
\resizebox{0.80\textwidth}{!}{%
\begin{tikzpicture}
\pgfplotsset{
  mse/.style={fill=blue,            draw=black, line width=0.2pt},
  cosine/.style={fill=red,          draw=black, line width=0.2pt},
  scea15/.style={fill=green!70!black,  draw=black, line width=0.2pt},
  scea20/.style={fill=orange!85!black, draw=black, line width=0.2pt},
  scea25/.style={fill=black,        draw=black, line width=0.2pt},
}

\newcommand{\lossitem}[2]{%
  \tikz[baseline=-0.6ex]\filldraw[fill=#1,draw=black] (0,0) rectangle (1.8mm,1.2mm);\,\scriptsize #2%
}

\begin{groupplot}[
    group style={group size=2 by 2, horizontal sep=0.06\textwidth, vertical sep=0.06\textwidth},
    width=0.49\textwidth,
    height=0.30\textwidth,
    ybar,
    /pgf/bar width=2.2pt,
    enlarge x limits=0.30,
    xmode=log,
    log basis x=2,
    xtick={1,2,4,8},
    xticklabels={1,2,4,8},
    xmin=0.9, xmax=8.5,
    ymajorgrids=true,
    xmajorgrids=true,
    tick label style={font=\scriptsize},
    label style={font=\scriptsize},
    title style={font=\scriptsize},
]

\nextgroupplot[
  title={MVTec AD},
  ylabel={I-AUROC (\%)},
  axis background/.style={fill=green!6},
  ymin=93.0, ymax=98.1
]
\addplot+[mse]    coordinates {(1,95.11) (2,95.46) (4,95.04) (8,95.33)};
\addplot+[cosine] coordinates {(1,96.91) (2,95.64) (4,94.54) (8,93.17)};
\addplot+[scea15] coordinates {(1,97.36) (2,97.64) (4,97.60) (8,97.67)};
\addplot+[scea20] coordinates {(1,97.27) (2,97.73) (4,97.75) (8,97.88)};
\addplot+[scea25] coordinates {(1,96.67) (2,97.40) (4,97.70) (8,97.84)};

\nextgroupplot[
  title={VisA},
  axis background/.style={fill=blue!6},
  ymin=88.5, ymax=95.5
]
\addplot+[mse]    coordinates {(1,90.90) (2,89.19) (4,92.23) (8,91.86)};
\addplot+[cosine] coordinates {(1,93.20) (2,90.35) (4,92.52) (8,88.93)};
\addplot+[scea15] coordinates {(1,93.41) (2,92.48) (4,94.08) (8,95.18)};
\addplot+[scea20] coordinates {(1,93.47) (2,92.64) (4,94.21) (8,95.13)};
\addplot+[scea25] coordinates {(1,93.19) (2,92.85) (4,94.48) (8,95.30)};

\nextgroupplot[
  ylabel={PRO (\%)},
  xlabel={Shots},
  axis background/.style={fill=green!6},
  ymin=86.0, ymax=95.0
]
\addplot+[mse]    coordinates {(1,93.58) (2,94.00) (4,94.00) (8,92.97)};
\addplot+[cosine] coordinates {(1,93.85) (2,91.94) (4,88.81) (8,86.51)};
\addplot+[scea15] coordinates {(1,93.90) (2,94.32) (4,94.49) (8,94.62)};
\addplot+[scea20] coordinates {(1,93.68) (2,94.17) (4,94.45) (8,94.60)};
\addplot+[scea25] coordinates {(1,93.21) (2,93.86) (4,94.22) (8,94.35)};

\nextgroupplot[
  xlabel={Shots},
  axis background/.style={fill=blue!6},
  ymin=83.0, ymax=95.5
]
\addplot+[mse]    coordinates {(1,92.87) (2,91.22) (4,94.31) (8,92.23)};
\addplot+[cosine] coordinates {(1,93.50) (2,92.61) (4,91.57) (8,83.19)};
\addplot+[scea15] coordinates {(1,94.01) (2,94.06) (4,94.78) (8,94.90)};
\addplot+[scea20] coordinates {(1,93.52) (2,93.78) (4,94.62) (8,94.70)};
\addplot+[scea25] coordinates {(1,92.74) (2,93.20) (4,94.14) (8,94.17)};

\end{groupplot}

\node[anchor=south, yshift=2mm] at (current bounding box.north) {%
\scriptsize
Objective function:\,
\lossitem{blue}{MSE}\quad
\lossitem{red}{Cosine}\quad
\lossitem{green!70!black}{SCE ($\gamma$=1.5)}\quad
\lossitem{orange!85!black}{SCE ($\gamma$=2.0)}\quad
\lossitem{black}{SCE ($\gamma$=2.5)}
};

\end{tikzpicture}%
}

\caption{Effect of the objective function on the MVTec AD and VisA benchmarks.}
\label{fig:abla_loss_mse_cosine_sce15_mvtec_visa}
\end{figure}
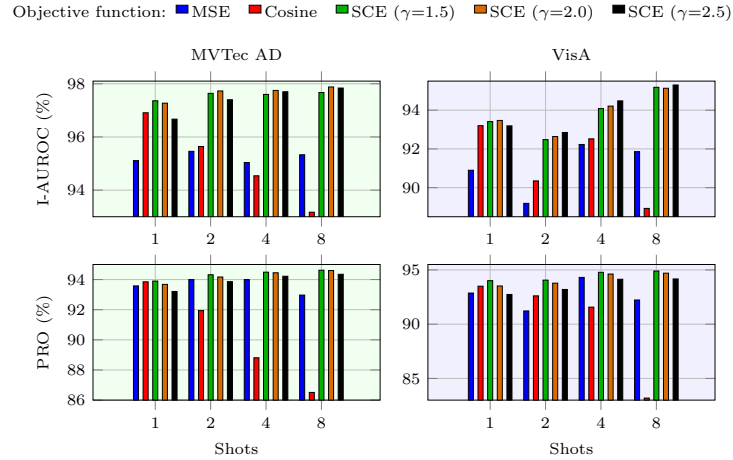

Fig. \ref{fig:abla_loss_mse_cosine_sce15_mvtec_visa} illustrates how the objective function (Section \ref{subsec:GAT}) affects the performance of GATE-AD across the MVTec AD and VisA benchmarks. From the presented results it can be seen that SCE is clearly advantageous over both MSE and cosine error. Concerning the SCE function, setting the scaling factor $\gamma=2$ leads to the most consistent results.

\subsection{Masking ratio}
\label{subsec:mask_ratio}

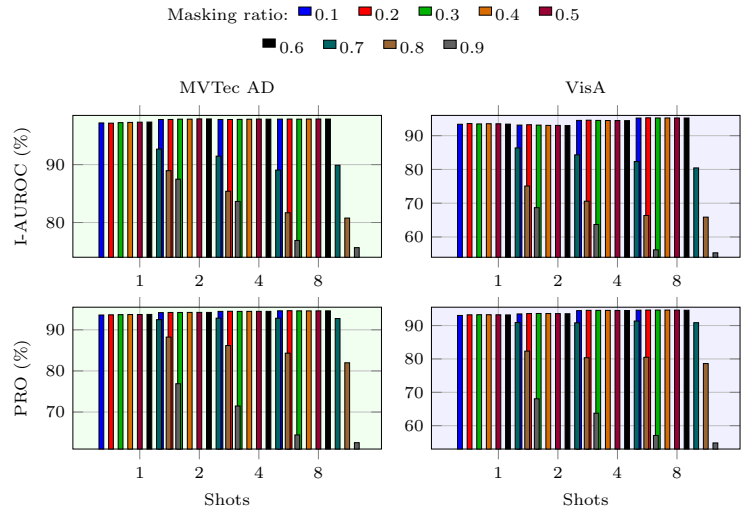
\begin{figure}[t!]
\centering
\resizebox{0.80\textwidth}{!}{%
\begin{tikzpicture}
\pgfplotsset{
  c01/.style={fill=blue,             draw=black, line width=0.2pt},   
  c02/.style={fill=red,              draw=black, line width=0.2pt},   
  c03/.style={fill=green!70!black,   draw=black, line width=0.2pt},   
  c04/.style={fill=orange!85!black,  draw=black, line width=0.2pt},   
  c05/.style={fill=purple!80!black,  draw=black, line width=0.2pt},   
  c06/.style={fill=black,            draw=black, line width=0.2pt},   
  c07/.style={fill=teal!80!black,    draw=black, line width=0.2pt},   
  c08/.style={fill=brown!80!black,   draw=black, line width=0.2pt},   
  c09/.style={fill=gray!75!black,    draw=black, line width=0.2pt},   
}

\newcommand{\maskitem}[2]{%
  \tikz[baseline=-0.6ex]\filldraw[fill=#1,draw=black] (0,0) rectangle (1.8mm,1.2mm);\,\scriptsize #2%
}

\begin{groupplot}[
    group style={group size=2 by 2, horizontal sep=0.06\textwidth, vertical sep=0.06\textwidth},
    width=0.50\textwidth,
    height=0.30\textwidth,
    ybar,
    /pgf/bar width=2.0pt,
    enlarge x limits=0.30,
    xmode=log,
    log basis x=2,
    xtick={1,2,4,8},
    xticklabels={1,2,4,8},
    xmin=0.9, xmax=8.5,
    ymajorgrids=true,
    xmajorgrids=true,
    tick label style={font=\scriptsize},
    label style={font=\scriptsize},
    title style={font=\scriptsize},
]

\nextgroupplot[
  title={MVTec AD},
  ylabel={I-AUROC (\%)},
  axis background/.style={fill=green!6},
  ymin=74.0, ymax=98.5
]
\addplot+[c01] coordinates {(1,97.20) (2,97.79) (4,97.80) (8,97.86)};
\addplot+[c02] coordinates {(1,97.16) (2,97.80) (4,97.79) (8,97.87)};
\addplot+[c03] coordinates {(1,97.25) (2,97.85) (4,97.81) (8,97.85)};
\addplot+[c04] coordinates {(1,97.29) (2,97.85) (4,97.84) (8,97.87)};
\addplot+[c05] coordinates {(1,97.33) (2,97.90) (4,97.87) (8,97.88)};
\addplot+[c06] coordinates {(1,97.35) (2,97.89) (4,97.85) (8,97.87)};
\addplot+[c07] coordinates {(1,92.68) (2,91.46) (4,89.06) (8,89.90)};
\addplot+[c08] coordinates {(1,88.95) (2,85.40) (4,81.68) (8,80.77)};
\addplot+[c09] coordinates {(1,87.48) (2,83.64) (4,76.90) (8,75.66)};

\nextgroupplot[
  title={VisA},
  axis background/.style={fill=blue!6},
  ymin=54.0, ymax=96.0
]
\addplot+[c01] coordinates {(1,93.38) (2,93.13) (4,94.52) (8,95.21)};
\addplot+[c02] coordinates {(1,93.59) (2,93.24) (4,94.58) (8,95.28)};
\addplot+[c03] coordinates {(1,93.47) (2,93.12) (4,94.52) (8,95.24)};
\addplot+[c04] coordinates {(1,93.54) (2,93.04) (4,94.48) (8,95.24)};
\addplot+[c05] coordinates {(1,93.52) (2,93.04) (4,94.47) (8,95.24)};
\addplot+[c06] coordinates {(1,93.42) (2,92.98) (4,94.45) (8,95.23)};
\addplot+[c07] coordinates {(1,86.37) (2,84.32) (4,82.32) (8,80.45)};
\addplot+[c08] coordinates {(1,75.09) (2,70.59) (4,66.37) (8,65.88)};
\addplot+[c09] coordinates {(1,68.68) (2,63.70) (4,56.19) (8,55.31)};

\nextgroupplot[
  ylabel={PRO (\%)},
  xlabel={Shots},
  axis background/.style={fill=green!6},
  ymin=61.0, ymax=95.5
]
\addplot+[c01] coordinates {(1,93.60) (2,94.18) (4,94.48) (8,94.62)};
\addplot+[c02] coordinates {(1,93.65) (2,94.22) (4,94.51) (8,94.65)};
\addplot+[c03] coordinates {(1,93.69) (2,94.22) (4,94.48) (8,94.61)};
\addplot+[c04] coordinates {(1,93.71) (2,94.23) (4,94.49) (8,94.60)};
\addplot+[c05] coordinates {(1,93.72) (2,94.24) (4,94.49) (8,94.61)};
\addplot+[c06] coordinates {(1,93.74) (2,94.23) (4,94.49) (8,94.61)};
\addplot+[c07] coordinates {(1,92.47) (2,92.83) (4,92.82) (8,92.75)};
\addplot+[c08] coordinates {(1,88.23) (2,86.17) (4,84.32) (8,81.98)};
\addplot+[c09] coordinates {(1,76.89) (2,71.47) (4,64.43) (8,62.55)};

\nextgroupplot[
  xlabel={Shots},
  axis background/.style={fill=blue!6},
  ymin=53.0, ymax=95.5
]
\addplot+[c01] coordinates {(1,93.03) (2,93.48) (4,94.48) (8,94.60)};
\addplot+[c02] coordinates {(1,93.22) (2,93.59) (4,94.56) (8,94.64)};
\addplot+[c03] coordinates {(1,93.25) (2,93.62) (4,94.55) (8,94.63)};
\addplot+[c04] coordinates {(1,93.25) (2,93.60) (4,94.55) (8,94.64)};
\addplot+[c05] coordinates {(1,93.23) (2,93.58) (4,94.53) (8,94.62)};
\addplot+[c06] coordinates {(1,93.19) (2,93.54) (4,94.51) (8,94.59)};
\addplot+[c07] coordinates {(1,90.91) (2,90.82) (4,91.36) (8,90.86)};
\addplot+[c08] coordinates {(1,82.34) (2,80.36) (4,80.50) (8,78.61)};
\addplot+[c09] coordinates {(1,68.06) (2,63.74) (4,57.07) (8,54.83)};

\end{groupplot}

\node[anchor=south, yshift=2mm, align=center] at (current bounding box.north) {%
\scriptsize
Masking ratio:\,
\maskitem{blue}{0.1}\quad
\maskitem{red}{0.2}\quad
\maskitem{green!70!black}{0.3}\quad
\maskitem{orange!85!black}{0.4}\quad
\maskitem{purple!80!black}{0.5}\\[1mm]
\maskitem{black}{0.6}\quad
\maskitem{teal!80!black}{0.7}\quad
\maskitem{brown!80!black}{0.8}\quad
\maskitem{gray!75!black}{0.9}
};

\end{tikzpicture}%
}
\caption{Effect of the masking ratio on the MVTec and VisA benchmarks.}
\label{fig:abla_maskratio}
\end{figure}

Fig. \ref{fig:abla_maskratio} illustrates how the masking ratio $\mu$ (Section \ref{subsec:GAT}) affects the performance of GATE-AD across the MVTec AD and VisA benchmarks. From the presented results it can be seen that $\mu \in [0.1,0.6]$ results into negligible performance variations, where selecting $\mu = 0.2$ leads to the best overall results.

\subsection{Dropout rate}
\label{subsec:dropout}

\begin{figure}[t!]
\centering
\resizebox{0.80\textwidth}{!}{%
\begin{tikzpicture}
\pgfplotsset{
  c00/.style={fill=blue,            draw=black, line width=0.2pt},
  c01/.style={fill=red,             draw=black, line width=0.2pt},
  c02/.style={fill=green!70!black,  draw=black, line width=0.2pt},
  c03/.style={fill=orange!85!black, draw=black, line width=0.2pt},
  c04/.style={fill=purple!80!black, draw=black, line width=0.2pt},
  c05/.style={fill=brown!80!black,  draw=black, line width=0.2pt},
  c06/.style={fill=cyan!70!black,   draw=black, line width=0.2pt},
  c07/.style={fill=black,           draw=black, line width=0.2pt},
}

\newcommand{\dropitem}[2]{%
  \tikz[baseline=-0.6ex]\filldraw[fill=#1,draw=black] (0,0) rectangle (1.8mm,1.2mm);\,\scriptsize #2%
}

\begin{groupplot}[
    group style={group size=2 by 2, horizontal sep=0.06\textwidth, vertical sep=0.06\textwidth},
    width=0.50\textwidth,
    height=0.28\textwidth,
    ybar,
    /pgf/bar width=2.0pt,
    enlarge x limits=0.30,
    xmode=log,
    log basis x=2,
    xtick={1,2,4,8},
    xticklabels={1,2,4,8},
    xmin=0.9, xmax=8.5,
    ymajorgrids=true,
    xmajorgrids=true,
    tick label style={font=\scriptsize},
    label style={font=\scriptsize},
    title style={font=\scriptsize},
]

\nextgroupplot[
  title={MVTec AD},
  ylabel={I-AUROC (\%)},
  axis background/.style={fill=green!6},
  ymin=96.0, ymax=98.1
]
\addplot+[c00] coordinates {(1,96.1) (2,96.9) (4,97.3) (8,97.4)};
\addplot+[c01] coordinates {(1,97.1) (2,97.6) (4,97.7) (8,97.9)};
\addplot+[c02] coordinates {(1,97.4) (2,97.6) (4,97.7) (8,97.9)};
\addplot+[c03] coordinates {(1,97.4) (2,97.9) (4,97.8) (8,97.9)};
\addplot+[c04] coordinates {(1,97.1) (2,97.6) (4,97.7) (8,97.9)};
\addplot+[c05] coordinates {(1,97.1) (2,97.6) (4,97.7) (8,97.8)};
\addplot+[c06] coordinates {(1,97.0) (2,97.4) (4,97.5) (8,97.6)};
\addplot+[c07] coordinates {(1,96.8) (2,97.3) (4,97.4) (8,97.5)};

\nextgroupplot[
  title={VisA},
  axis background/.style={fill=blue!6},
  ymin=91.5, ymax=96.0
]
\addplot+[c00] coordinates {(1,91.9) (2,92.4) (4,94.2) (8,95.1)};
\addplot+[c01] coordinates {(1,93.9) (2,93.7) (4,94.7) (8,95.6)};
\addplot+[c02] coordinates {(1,94.0) (2,93.7) (4,94.7) (8,95.6)};
\addplot+[c03] coordinates {(1,93.5) (2,93.1) (4,94.5) (8,95.2)};
\addplot+[c04] coordinates {(1,94.0) (2,93.6) (4,94.6) (8,95.5)};
\addplot+[c05] coordinates {(1,93.9) (2,93.7) (4,94.7) (8,95.5)};
\addplot+[c06] coordinates {(1,93.7) (2,93.4) (4,94.4) (8,95.3)};
\addplot+[c07] coordinates {(1,93.6) (2,93.1) (4,94.4) (8,95.1)};

\nextgroupplot[
  ylabel={PRO (\%)},
  xlabel={Shots},
  axis background/.style={fill=green!6},
  ymin=92.6, ymax=95.0
]
\addplot+[c00] coordinates {(1,92.7) (2,93.4) (4,94.0) (8,94.1)};
\addplot+[c01] coordinates {(1,93.8) (2,94.4) (4,94.6) (8,94.7)};
\addplot+[c02] coordinates {(1,93.8) (2,94.4) (4,94.6) (8,94.7)};
\addplot+[c03] coordinates {(1,93.7) (2,94.2) (4,94.5) (8,94.6)};
\addplot+[c04] coordinates {(1,93.8) (2,94.3) (4,94.6) (8,94.7)};
\addplot+[c05] coordinates {(1,93.7) (2,94.3) (4,94.5) (8,94.6)};
\addplot+[c06] coordinates {(1,93.6) (2,94.1) (4,94.3) (8,94.4)};
\addplot+[c07] coordinates {(1,93.4) (2,93.9) (4,94.1) (8,94.2)};

\nextgroupplot[
  xlabel={Shots},
  axis background/.style={fill=blue!6},
  ymin=91.0, ymax=95.0
]
\addplot+[c00] coordinates {(1,91.3) (2,92.2) (4,93.4) (8,93.8)};
\addplot+[c01] coordinates {(1,93.2) (2,93.7) (4,94.5) (8,94.6)};
\addplot+[c02] coordinates {(1,93.3) (2,93.7) (4,94.5) (8,94.5)};
\addplot+[c03] coordinates {(1,93.2) (2,93.6) (4,94.5) (8,94.6)};
\addplot+[c04] coordinates {(1,93.5) (2,93.8) (4,94.6) (8,94.6)};
\addplot+[c05] coordinates {(1,93.5) (2,93.8) (4,94.5) (8,94.5)};
\addplot+[c06] coordinates {(1,93.1) (2,93.5) (4,94.7) (8,94.7)};
\addplot+[c07] coordinates {(1,93.1) (2,93.5) (4,94.3) (8,94.3)};

\end{groupplot}

\node[anchor=south, yshift=2mm] at (current bounding box.north) {%
\scriptsize
Dropout rate::\,
\dropitem{blue}{0.0}\quad
\dropitem{red}{0.1}\quad
\dropitem{green!70!black}{0.2}\quad
\dropitem{orange!85!black}{0.3}\quad
\dropitem{purple!80!black}{0.4}\quad
\dropitem{brown!80!black}{0.5}\quad
\dropitem{cyan!70!black}{0.6}\quad
\dropitem{black}{0.7}
};
\end{tikzpicture}%
}
\caption{Effect of the dropout rate ablation on the MVTec and VisA benchmarks.}
\label{fig:abla_dropout}
\end{figure}
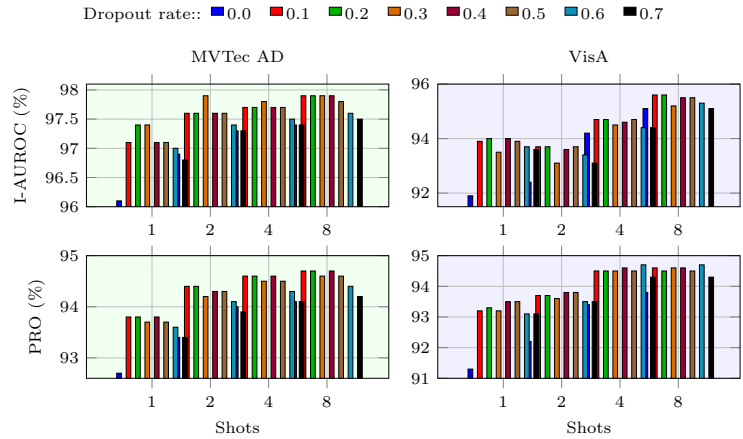

Fig. \ref{fig:abla_dropout} illustrates how the dropout rate $\delta$ (Section \ref{subsec:GAT}) affects the performance of GATE-AD across the MVTec AD and VisA benchmarks. From the presented results it can be seen that $\delta \in [0.1,0.4]$ results in the highest detection performance, where $\delta=0.3$ being the optimal value.

\subsection{Anomalous patches ratio}
\label{subsec:topk}

\begin{figure}[t!]
\centering
\resizebox{0.80\textwidth}{!}{%
\begin{tikzpicture}
\pgfplotsset{
  c1/.style={fill=blue,            draw=black, line width=0.2pt},
  c2/.style={fill=teal!70!black,   draw=black, line width=0.2pt},
  c3/.style={fill=orange!85!black, draw=black, line width=0.2pt},
  c4/.style={fill=red,             draw=black, line width=0.2pt},
  c5/.style={fill=green!60!black,  draw=black, line width=0.2pt},
  c6/.style={fill=purple!80!black, draw=black, line width=0.2pt},
}

\newcommand{\topkitem}[2]{%
  \tikz[baseline=-0.6ex]\filldraw[fill=#1,draw=black] (0,0) rectangle (1.8mm,1.2mm);\,\scriptsize #2%
}

\begin{groupplot}[
    group style={group size=2 by 1, horizontal sep=0.06\textwidth},
    width=0.49\textwidth,
    height=0.30\textwidth,
    ybar,
    /pgf/bar width=2.2pt,
    enlarge x limits=0.30,
    xmode=log,
    log basis x=2,
    xtick={1,2,4,8},
    xticklabels={1,2,4,8},
    xmin=0.9, xmax=8.5,
    ymajorgrids=true,
    xmajorgrids=true,
    tick label style={font=\scriptsize},
    label style={font=\scriptsize},
    title style={font=\scriptsize},
]

\nextgroupplot[
  title={MVTec AD},
  ylabel={I-AUROC (\%)},
  axis background/.style={fill=green!6},
  ymin=96.5, ymax=97.9
]
\addplot+[c1] coordinates {(1,96.64) (2,97.41) (4,97.61) (8,97.74)};
\addplot+[c2] coordinates {(1,97.09) (2,97.66) (4,97.73) (8,97.78)};
\addplot+[c3] coordinates {(1,97.33) (2,97.64) (4,97.67) (8,97.77)};
\addplot+[c4] coordinates {(1,97.16) (2,97.61) (4,97.66) (8,97.71)};
\addplot+[c5] coordinates {(1,97.20) (2,97.55) (4,97.63) (8,97.70)};
\addplot+[c6] coordinates {(1,97.17) (2,97.51) (4,97.58) (8,97.72)};

\nextgroupplot[
  title={VisA},
  axis background/.style={fill=blue!6},
  ymin=92.9, ymax=95.9
]
\addplot+[c1] coordinates {(1,94.20) (2,93.96) (4,94.95) (8,95.75)};
\addplot+[c2] coordinates {(1,94.15) (2,93.75) (4,94.80) (8,95.61)};
\addplot+[c3] coordinates {(1,94.08) (2,93.60) (4,94.75) (8,95.61)};
\addplot+[c4] coordinates {(1,93.98) (2,93.49) (4,94.69) (8,95.62)};
\addplot+[c5] coordinates {(1,93.80) (2,93.27) (4,94.58) (8,95.60)};
\addplot+[c6] coordinates {(1,93.65) (2,93.09) (4,94.47) (8,95.53)};

\end{groupplot}

\node[anchor=south, yshift=2mm] at (current bounding box.north) {%
\scriptsize
Anomalous patches ratio:\,
\topkitem{blue}{0.01}\quad
\topkitem{teal!70!black}{0.02}\quad
\topkitem{orange!85!black}{0.025}\quad
\topkitem{red}{0.03}\quad
\topkitem{green!60!black}{0.04}\quad
\topkitem{purple!80!black}{0.05}
};

\end{tikzpicture}%
}
\caption{Effect of the anomalous patches ratio on the MVTec AD and VisA benchmarks.}
\label{fig:abla_topk_mvtec_visa}
\end{figure}
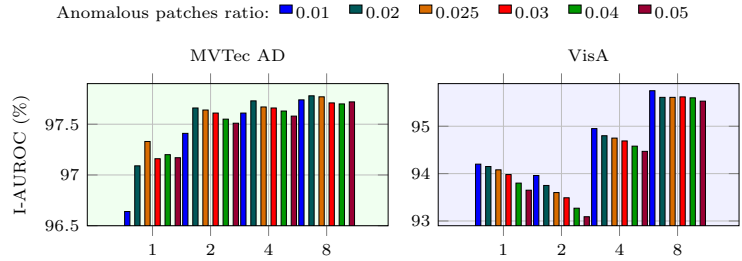

Fig.~\ref{fig:abla_topk_mvtec_visa} illustrates how the anomalous patches ratio $\epsilon$ (Section~\ref{subsec:inference}) affects the image-level detection performance of GATE-AD on MVTec AD and VisA. A ratio of $\epsilon=0.025$ on MVTec AD and $\epsilon=0.01$ on VisA provides the best overall performance, while deviations from these values result in only moderate performance variations. These dataset-specific values are therefore retained for the within-method analyses of Section~\ref{ssec:ablation}; the comparative evaluation follows the unified protocol defined in Section~\ref{ssec:setup}.

\subsection{Training budget and early stopping}
\label{subsec:epoch_budget}

The maximum training budget of $2{,}000$ epochs is employed as an early-stopped upper optimization bound rather than as a fixed training duration. Despite the limited number of normal support images in FS-IVAD, the unit of supervision in GATE-AD is the patch node rather than the image. Hence, even a single support image provides a large number of patch-level training instances. Moreover, at every epoch, a newly sampled subset of nodes is masked, such that successive epochs correspond to different reconstruction problems rather than repeated optimization over an identical input configuration. This stochastic corruption acts as an additional regularization mechanism against memorization of the available support images. In practice, early stopping terminates optimization well before the maximum budget in many cases, with an average training duration of approximately $1{,}569$ epochs across all datasets, shot settings, categories, and support-set seeds. To further examine the effect of the training duration, a fixed-budget ablation is conducted on MVTec AD under the $1$-shot setting, which represents the regime most susceptible to overfitting. For this experiment, early stopping is disabled and the optimization budget is varied from $100$ to $2{,}000$ epochs. As shown in Table~\ref{tab:epoch_budget}, detection and localization performance progressively improve as the training budget increases and largely stabilize after approximately $1{,}000$ epochs. Increasing the budget to $2{,}000$ epochs introduces only minor fluctuations, with no indication of performance collapse. These results show that competitive performance is reached well before the maximum optimization budget and support the use of $2{,}000$ epochs as an early-stopped upper bound.

\begin{table}[h]
\centering
\small
\setlength{\tabcolsep}{5.0pt}
\renewcommand{\arraystretch}{1.05}
\caption{Effect of the fixed training budget on MVTec AD under the $1$-shot setting, with early stopping disabled (\%).}
\label{tab:epoch_budget}
\begin{tabular}{@{}ccccc@{}}
\toprule
\textbf{Epochs} &
\textbf{I-AUROC} &
\textbf{I-AP} &
\textbf{P-AUROC} &
\textbf{AUPRO} \\
\midrule
100     & 96.21 & 97.77 & 95.04 & 91.54 \\
200     & 97.03 & 98.26 & 95.48 & 92.12 \\
500     & 97.25 & 98.48 & 95.88 & 92.55 \\
1,000   & 97.40 & 98.54 & 96.11 & 92.70 \\
1,500   & \textbf{97.65} & \textbf{98.81} & \textbf{96.27} & \textbf{93.00} \\
2,000   & 97.33 & 98.57 & 96.18 & 92.75 \\
\bottomrule
\end{tabular}
\end{table}

\section{Additional experimental analysis}
\label{sec:perclass}

This section provides a detailed characterization of GATE-AD, complementary to the controlled comparative evaluation of Section~\ref{ssec:comparative_evaluation}. Unless otherwise stated, the supplementary GATE-AD analyses use the dataset-specific configurations identified through the hyper-parameter analysis of Appendix~\ref{sec:hparam}. Object-level quantitative results, product-agnostic defect-type analysis, qualitative results, query-conditioned representation analysis, and representative failure cases are examined.

\subsection{Quantitative evaluation results}
\label{subsec:quantitative}

Tables \ref{tab:object_level_mvtec}-\ref{tab:object_level_mpdd} illustrate the object-level detection performance of GATE-AD and AnomalyDINO \citep{damm2025anomalydino}
across the MVTec AD, VisA, and MPDD benchmarks. In order to ensure a fair and consistent comparison, both methods utilize the standardized sample selection process. Key insights on a per dataset basis are provided in the following.

\noindent
\underline{MVTec AD (Table \ref{tab:object_level_mvtec})}:
\begin{itemize}
    \item GATE-AD achieves high detection scores even in the $1$-shot setting, accomplishing an $100.0\%$ image AUROC for classes `Bottle' and `Leather'.
    \item GATE-AD performs robustly for challenging classes, like `Cable' ($94.7\%$ image AUROC in the $1$-shot setting) and `Capsule' ($93.1\%$ image AUROC in the $1$-shot case), significantly outperforming AnomalyDINO.
    \item GATE-AD maintains high localization accuracy in most cases, exhibiting up to $98.0\%$ pixel PRO for class `Leather' in the 1-shot setting.
\end{itemize}

\begin{table}[t!]
\centering
\scriptsize
\setlength{\tabcolsep}{3.2pt}
\renewcommand{\arraystretch}{1.15}
\caption{FS-IVAD object-level detection results for GATE-AD and AnomalyDINO on the MVTec AD benchmark.}
\label{tab:object_level_mvtec}

\resizebox{\textwidth}{!}{%
\begin{tabular}{l cc cc cc cc}
\toprule
\multirow{2}{*}{Object}
& \multicolumn{2}{c}{1-shot} & \multicolumn{2}{c}{2-shot}
& \multicolumn{2}{c}{4-shot} & \multicolumn{2}{c}{8-shot} \\
\cmidrule(lr){2-3}\cmidrule(lr){4-5}\cmidrule(lr){6-7}\cmidrule(lr){8-9}
& GATE-AD & AnomalyDINO
& GATE-AD & AnomalyDINO
& GATE-AD & AnomalyDINO
& GATE-AD & AnomalyDINO \\
\midrule
\multicolumn{9}{c}{\textbf{Image AUROC (\%)}} \\
\midrule

Bottle      & \textbf{\mstd{100.0}{0.0}} & \mstd{99.7}{0.2} & \mstd{99.8}{0.1} & \textbf{\mstd{99.9}{0.1}} & \textbf{\mstd{100.0}{0.0}} & \mstd{99.9}{0.1} & \textbf{\mstd{100.0}{0.0}} & \textbf{\mstd{100.0}{0.0}} \\
Cable       & \textbf{\mstd{94.7}{0.1}} & \mstd{92.7}{0.8} & \textbf{\mstd{95.5}{0.1}} & \mstd{92.4}{1.1} & \textbf{\mstd{94.3}{0.1}} & \mstd{93.8}{0.9} & \mstd{95.0}{0.4} & \textbf{\mstd{95.2}{0.3}} \\
Capsule     & \textbf{\mstd{93.1}{0.1}} & \mstd{90.2}{5.5} & \textbf{\mstd{94.3}{0.7}} & \mstd{89.2}{7.9} & \mstd{95.0}{0.9} & \textbf{\mstd{95.8}{0.5}} & \textbf{\mstd{95.8}{0.8}} & \mstd{95.6}{0.5} \\
Carpet      & \mstd{99.8}{0.0} & \textbf{\mstd{100.0}{0.0}} & \mstd{99.9}{0.0} & \textbf{\mstd{100.0}{0.0}} & \mstd{99.8}{0.0} & \textbf{\mstd{100.0}{0.0}} & \mstd{99.9}{0.0} & \textbf{\mstd{100.0}{0.0}} \\
Grid        & \textbf{\mstd{99.6}{0.0}} & \mstd{99.1}{0.2} & \textbf{\mstd{100.0}{0.0}} & \mstd{99.2}{0.4} & \textbf{\mstd{100.0}{0.0}} & \mstd{99.5}{0.3} & \textbf{\mstd{100.0}{0.0}} & \mstd{99.5}{0.1} \\
Hazelnut    & \textbf{\mstd{97.7}{0.9}} & \mstd{97.5}{2.6} & \mstd{98.9}{0.4} & \textbf{\mstd{99.6}{0.5}} & \mstd{99.6}{0.3} & \textbf{\mstd{99.8}{0.1}} & \mstd{99.9}{0.1} & \textbf{\mstd{100.0}{0.0}} \\
Leather     & \textbf{\mstd{100.0}{0.0}} & \textbf{\mstd{100.0}{0.0}} & \textbf{\mstd{100.0}{0.0}} & \textbf{\mstd{100.0}{0.0}} & \textbf{\mstd{100.0}{0.0}} & \textbf{\mstd{100.0}{0.0}} & \textbf{\mstd{100.0}{0.0}} & \textbf{\mstd{100.0}{0.0}} \\
Metal nut   & \mstd{99.4}{0.3} & \textbf{\mstd{99.9}{0.1}} & \mstd{99.2}{0.3} & \textbf{\mstd{100.0}{0.0}} & \mstd{99.7}{0.1} & \textbf{\mstd{100.0}{0.0}} & \mstd{99.7}{0.0} & \textbf{\mstd{100.0}{0.0}} \\
Pill        & \textbf{\mstd{96.3}{1.0}} & \mstd{93.7}{0.9} & \textbf{\mstd{97.3}{0.6}} & \mstd{95.4}{0.7} & \textbf{\mstd{96.5}{0.5}} & \mstd{96.0}{0.2} & \textbf{\mstd{97.7}{0.3}} & \mstd{97.2}{0.2} \\
Screw       & \textbf{\mstd{94.6}{0.2}} & \mstd{93.2}{0.3} & \textbf{\mstd{95.1}{0.2}} & \mstd{93.5}{0.8} & \textbf{\mstd{94.5}{0.0}} & \mstd{92.7}{2.3} & \textbf{\mstd{94.5}{0.2}} & \mstd{93.5}{1.1} \\
Tile        & \mstd{99.5}{0.1} & \textbf{\mstd{100.0}{0.0}} & \mstd{99.7}{0.0} & \textbf{\mstd{100.0}{0.0}} & \mstd{99.9}{0.0} & \textbf{\mstd{100.0}{0.0}} & \textbf{\mstd{100.0}{0.0}} & \textbf{\mstd{100.0}{0.0}} \\
Toothbrush  & \textbf{\mstd{98.1}{0.4}} & \mstd{97.4}{0.5} & \mstd{97.5}{0.0} & \textbf{\mstd{98.1}{1.0}} & \mstd{96.4}{0.0} & \textbf{\mstd{97.5}{0.6}} & \mstd{96.7}{0.4} & \textbf{\mstd{97.7}{1.6}} \\
Transistor  & \textbf{\mstd{93.0}{0.4}} & \mstd{90.9}{1.2} & \textbf{\mstd{93.1}{0.7}} & \mstd{89.4}{4.6} & \textbf{\mstd{93.3}{0.4}} & \mstd{93.2}{2.2} & \mstd{94.6}{0.1} & \textbf{\mstd{96.2}{1.3}} \\
Wood        & \textbf{\mstd{99.8}{0.0}} & \mstd{98.0}{0.2} & \textbf{\mstd{100.0}{0.0}} & \mstd{98.0}{0.1} & \textbf{\mstd{100.0}{0.0}} & \mstd{97.9}{0.2} & \textbf{\mstd{100.0}{0.0}} & \mstd{98.3}{0.4} \\
Zipper      & \textbf{\mstd{99.8}{0.2}} & \mstd{97.4}{0.9} & \textbf{\mstd{99.9}{0.1}} & \mstd{98.9}{0.4} & \textbf{\mstd{99.8}{0.1}} & \mstd{99.0}{0.4} & \textbf{\mstd{99.9}{0.1}} & \mstd{99.6}{0.1} \\
\midrule
Mean        & \textbf{\mstd{97.7}{0.2}} & \mstd{96.6}{0.4} & \textbf{\mstd{98.0}{0.2}} & \mstd{96.9}{0.7} & \textbf{\mstd{97.9}{0.2}} & \mstd{97.7}{0.2} & \textbf{\mstd{98.2}{0.1}} & \textbf{\mstd{98.2}{0.2}} \\

\midrule
\multicolumn{9}{c}{\textbf{Pixel PRO (\%)}} \\
\midrule
Bottle      & \mstd{95.7}{0.1} & \textbf{\mstd{95.9}{0.3}} & \mstd{95.7}{0.1} & \textbf{\mstd{96.3}{0.0}} & \mstd{95.9}{0.1} & \textbf{\mstd{96.7}{0.1}} & \mstd{96.0}{0.1} & \textbf{\mstd{96.5}{0.4}} \\
Cable       & \textbf{\mstd{89.8}{0.4}} & \mstd{89.4}{0.4} & \textbf{\mstd{90.4}{0.1}} & \mstd{89.5}{0.4} & \mstd{90.1}{0.1} & \textbf{\mstd{90.4}{0.3}} & \textbf{\mstd{90.5}{0.1}} & \textbf{\mstd{90.5}{0.2}} \\
Capsule     & \mstd{96.6}{0.1} & \textbf{\mstd{97.1}{0.3}} & \mstd{96.7}{0.1} & \textbf{\mstd{97.3}{0.6}} & \mstd{97.0}{0.1} & \textbf{\mstd{97.9}{0.1}} & \mstd{97.1}{0.1} & \textbf{\mstd{97.9}{0.1}} \\
Carpet      & \mstd{96.7}{0.0} & \textbf{\mstd{97.8}{0.0}} & \mstd{96.8}{0.0} & \textbf{\mstd{97.9}{0.0}} & \mstd{96.7}{0.1} & \textbf{\mstd{97.8}{0.0}} & \mstd{96.7}{0.1} & \textbf{\mstd{97.8}{0.0}} \\
Grid        & \mstd{96.5}{0.2} & \textbf{\mstd{97.2}{0.1}} & \mstd{96.6}{0.2} & \textbf{\mstd{97.2}{0.1}} & \mstd{96.6}{0.1} & \textbf{\mstd{97.2}{0.1}} & \mstd{96.6}{0.1} & \textbf{\mstd{97.2}{0.0}} \\
Hazelnut    & \mstd{96.7}{0.1} & \textbf{\mstd{97.4}{0.4}} & \mstd{97.1}{0.0} & \textbf{\mstd{98.0}{0.3}} & \mstd{97.3}{0.1} & \textbf{\mstd{98.0}{0.1}} & \mstd{97.3}{0.1} & \textbf{\mstd{98.1}{0.1}} \\
Leather     & \textbf{\mstd{98.0}{0.2}} & \mstd{97.9}{0.1} & \textbf{\mstd{97.9}{0.2}} & \mstd{97.8}{0.0} & \textbf{\mstd{97.9}{0.1}} & \mstd{97.6}{0.1} & \textbf{\mstd{97.8}{0.2}} & \mstd{97.6}{0.1} \\
Metal nut   & \mstd{92.2}{0.1} & \textbf{\mstd{94.2}{0.0}} & \mstd{92.7}{0.2} & \textbf{\mstd{94.6}{0.2}} & \mstd{93.4}{0.0} & \textbf{\mstd{95.3}{0.1}} & \mstd{93.8}{0.2} & \textbf{\mstd{95.4}{0.3}} \\
Pill        & \mstd{96.6}{0.1} & \textbf{\mstd{97.3}{0.1}} & \mstd{96.9}{0.1} & \textbf{\mstd{97.5}{0.1}} & \mstd{96.7}{0.1} & \textbf{\mstd{97.6}{0.1}} & \mstd{96.9}{0.1} & \textbf{\mstd{97.7}{0.1}} \\
Screw       & \textbf{\mstd{96.1}{0.1}} & \mstd{93.4}{0.4} & \textbf{\mstd{96.6}{0.1}} & \mstd{94.3}{0.4} & \textbf{\mstd{96.6}{0.1}} & \mstd{94.5}{0.9} & \textbf{\mstd{96.8}{0.1}} & \mstd{95.2}{0.5} \\
Tile        & \textbf{\mstd{89.7}{0.4}} & \mstd{88.0}{0.2} & \textbf{\mstd{90.1}{0.5}} & \mstd{87.6}{0.4} & \textbf{\mstd{90.0}{0.4}} & \mstd{87.1}{0.4} & \textbf{\mstd{89.9}{0.4}} & \mstd{86.7}{0.2} \\
Toothbrush  & \textbf{\mstd{96.0}{0.0}} & \mstd{94.0}{0.8} & \textbf{\mstd{95.9}{0.1}} & \mstd{94.7}{0.3} & \textbf{\mstd{95.5}{0.0}} & \mstd{94.7}{0.3} & \textbf{\mstd{95.7}{0.0}} & \mstd{95.1}{0.8} \\
Transistor  & \textbf{\mstd{68.2}{0.3}} & \mstd{67.3}{2.1} & \textbf{\mstd{70.5}{0.1}} & \mstd{68.4}{3.1} & \textbf{\mstd{70.9}{0.1}} & \mstd{70.6}{1.4} & \mstd{71.4}{0.1} & \textbf{\mstd{75.3}{1.9}} \\
Wood        & \mstd{93.4}{0.1} & \textbf{\mstd{94.7}{0.1}} & \mstd{93.6}{0.0} & \textbf{\mstd{94.6}{0.0}} & \mstd{93.4}{0.0} & \textbf{\mstd{94.6}{0.1}} & \mstd{93.9}{0.1} & \textbf{\mstd{94.4}{0.3}} \\
Zipper      & \textbf{\mstd{93.2}{0.1}} & \mstd{89.2}{1.2} & \textbf{\mstd{94.2}{0.4}} & \mstd{90.2}{0.4} & \textbf{\mstd{94.1}{0.3}} & \mstd{91.2}{0.3} & \textbf{\mstd{94.8}{0.3}} & \mstd{91.1}{0.4} \\
\midrule
Mean        & \textbf{\mstd{93.0}{0.2}} & \mstd{92.7}{0.1} & \textbf{\mstd{93.4}{0.1}} & \mstd{93.1}{0.2} & \textbf{\mstd{93.5}{0.1}} & \mstd{93.4}{0.1} & \mstd{93.7}{0.1} & \textbf{\mstd{93.8}{0.1}} \\

\bottomrule
\end{tabular}%
}
\end{table}

\begin{table}[t!]
\centering
\scriptsize
\setlength{\tabcolsep}{3.2pt}
\renewcommand{\arraystretch}{1.15}
\caption{FS-IVAD object-level detection results for GATE-AD and AnomalyDINO on the VisA benchmark.}
\label{tab:object_level_visa}
\resizebox{\textwidth}{!}{%
\begin{tabular}{l cc cc cc cc}
\toprule
\multirow{2}{*}{Object}
& \multicolumn{2}{c}{1-shot} & \multicolumn{2}{c}{2-shot}
& \multicolumn{2}{c}{4-shot} & \multicolumn{2}{c}{8-shot} \\
\cmidrule(lr){2-3}\cmidrule(lr){4-5}\cmidrule(lr){6-7}\cmidrule(lr){8-9}
& GATE-AD & AnomalyDINO
& GATE-AD & AnomalyDINO
& GATE-AD & AnomalyDINO
& GATE-AD & AnomalyDINO \\
\midrule
\multicolumn{9}{c}{\textbf{Image AUROC (\%)}} \\
\midrule

Candle      & \textbf{\mstd{97.8}{0.4}} & \mstd{87.9}{0.3} & \textbf{\mstd{97.6}{0.3}} & \mstd{89.4}{3.0} & \textbf{\mstd{97.5}{0.3}} & \mstd{91.3}{2.9} & \textbf{\mstd{97.7}{0.2}} & \mstd{93.5}{1.2} \\
Capsules    & \textbf{\mstd{99.3}{0.3}} & \mstd{98.4}{0.5} & \textbf{\mstd{99.5}{0.2}} & \mstd{98.9}{0.1} & \textbf{\mstd{99.6}{0.2}} & \mstd{99.2}{0.1} & \textbf{\mstd{99.6}{0.2}} & \mstd{99.2}{0.1} \\
Cashew      & \textbf{\mstd{95.5}{0.4}} & \mstd{86.1}{3.6} & \textbf{\mstd{96.1}{0.2}} & \mstd{89.4}{3.8} & \textbf{\mstd{98.7}{0.2}} & \mstd{94.5}{0.7} & \textbf{\mstd{98.6}{0.2}} & \mstd{95.3}{0.6} \\
Chewing gum  & \textbf{\mstd{98.8}{0.2}} & \mstd{98.0}{0.4} & \textbf{\mstd{99.1}{0.1}} & \mstd{98.6}{0.4} & \textbf{\mstd{99.3}{0.1}} & \mstd{98.8}{0.2} & \textbf{\mstd{99.4}{0.1}} & \mstd{98.8}{0.2} \\
Fryum       & \textbf{\mstd{97.4}{0.4}} & \mstd{94.8}{0.5} & \textbf{\mstd{98.4}{0.1}} & \mstd{96.5}{0.2} & \textbf{\mstd{98.7}{0.2}} & \mstd{97.0}{0.1} & \textbf{\mstd{98.6}{0.2}} & \mstd{97.6}{0.4} \\
Macaroni1   & \textbf{\mstd{96.0}{0.4}} & \mstd{87.5}{1.1} & \textbf{\mstd{95.4}{0.5}} & \mstd{87.5}{0.9} & \textbf{\mstd{95.9}{0.3}} & \mstd{89.5}{1.4} & \textbf{\mstd{96.9}{0.3}} & \mstd{90.1}{1.7} \\
Macaroni2   & \textbf{\mstd{75.9}{1.4}} & \mstd{62.2}{4.3} & \textbf{\mstd{73.3}{1.2}} & \mstd{66.9}{1.9} & \textbf{\mstd{72.6}{1.0}} & \mstd{70.0}{1.7} & \textbf{\mstd{75.7}{0.7}} & \mstd{74.9}{0.4} \\
PCB1        & \textbf{\mstd{98.0}{0.3}} & \mstd{91.5}{2.0} & \textbf{\mstd{97.9}{0.4}} & \mstd{91.2}{2.7} & \textbf{\mstd{97.6}{0.4}} & \mstd{94.0}{2.1} & \textbf{\mstd{97.6}{0.3}} & \mstd{95.5}{0.5} \\
PCB2        & \textbf{\mstd{92.0}{0.5}} & \mstd{84.8}{1.2} & \textbf{\mstd{91.5}{0.6}} & \mstd{88.1}{2.5} & \textbf{\mstd{91.4}{0.7}} & \mstd{91.1}{1.7} & \mstd{92.1}{0.3} & \textbf{\mstd{92.6}{0.3}} \\
PCB3        & \textbf{\mstd{85.7}{0.3}} & \mstd{84.9}{3.3} & \textbf{\mstd{91.1}{0.2}} & \mstd{89.4}{3.8} & \mstd{94.0}{0.2} & \textbf{\mstd{94.3}{0.4}} & \mstd{93.6}{0.4} & \textbf{\mstd{95.6}{0.2}} \\
PCB4        & \textbf{\mstd{88.3}{0.8}} & \mstd{79.9}{13.7}& \mstd{78.2}{2.0} & \textbf{\mstd{87.4}{11.3}}& \mstd{91.6}{1.0} & \textbf{\mstd{96.2}{2.6}} & \mstd{96.2}{0.3} & \textbf{\mstd{98.0}{0.3}} \\
Pipe fryum  & \textbf{\mstd{96.9}{0.6}} & \mstd{92.7}{2.7} & \textbf{\mstd{95.9}{0.7}} & \mstd{93.3}{1.2} & \textbf{\mstd{96.1}{0.5}} & \mstd{94.6}{1.9} & \textbf{\mstd{96.7}{0.5}} & \mstd{95.0}{1.6} \\
\midrule
Mean        & \textbf{\mstd{93.5}{0.2}} & \mstd{87.4}{1.2} & \textbf{\mstd{92.8}{0.3}} & \mstd{89.7}{1.3} & \textbf{\mstd{94.4}{0.2}} & \mstd{92.6}{0.9} & \textbf{\mstd{95.2}{0.2}} & \mstd{93.8}{0.3} \\

\midrule
\multicolumn{9}{c}{\textbf{Pixel PRO (\%)}} \\
\midrule
Candle      & \textbf{\mstd{97.0}{0.2}} & \mstd{96.8}{0.4} & \textbf{\mstd{97.1}{0.2}} & \mstd{97.0}{0.2} & \textbf{\mstd{97.3}{0.2}} & \mstd{97.2}{0.1} & \textbf{\mstd{97.4}{0.2}} & \mstd{97.3}{0.2} \\
Capsules    & \textbf{\mstd{97.0}{0.3}} & \mstd{95.1}{0.7} & \textbf{\mstd{97.1}{0.3}} & \mstd{95.5}{0.2} & \textbf{\mstd{97.4}{0.3}} & \mstd{96.3}{0.4} & \textbf{\mstd{97.7}{0.3}} & \mstd{96.7}{0.2} \\
Cashew      & \textbf{\mstd{96.4}{0.2}} & \mstd{96.1}{0.9} & \textbf{\mstd{96.8}{0.3}} & \mstd{96.7}{0.7} & \mstd{96.9}{0.4} & \textbf{\mstd{97.4}{0.5}} & \mstd{96.9}{0.3} & \textbf{\mstd{97.3}{0.2}} \\
Chewing gum  & \textbf{\mstd{93.0}{1.0}} & \mstd{92.0}{1.0} & \textbf{\mstd{93.2}{0.9}} & \mstd{92.9}{0.3} & \textbf{\mstd{93.7}{0.7}} & \mstd{93.0}{0.1} & \textbf{\mstd{93.9}{0.5}} & \mstd{93.1}{0.3} \\
Fryum       & \mstd{92.8}{0.3} & \textbf{\mstd{93.2}{0.2}} & \mstd{93.8}{0.4} & \textbf{\mstd{93.9}{0.3}} & \mstd{94.0}{0.3} & \textbf{\mstd{94.5}{0.4}} & \mstd{93.9}{0.3} & \textbf{\mstd{94.9}{0.3}} \\
Macaroni1   & \mstd{97.1}{0.3} & \textbf{\mstd{97.5}{0.3}} & \mstd{97.3}{0.3} & \textbf{\mstd{97.9}{0.3}} & \mstd{97.6}{0.3} & \textbf{\mstd{98.3}{0.2}} & \mstd{97.9}{0.2} & \textbf{\mstd{98.6}{0.2}} \\
Macaroni2   & \textbf{\mstd{96.1}{0.3}} & \mstd{92.0}{0.7} & \textbf{\mstd{96.1}{0.3}} & \mstd{93.0}{0.4} & \textbf{\mstd{96.4}{0.3}} & \mstd{93.9}{0.8} & \textbf{\mstd{96.8}{0.3}} & \mstd{95.0}{0.6} \\
PCB1        & \textbf{\mstd{93.5}{0.5}} & \mstd{92.6}{0.2} & \textbf{\mstd{94.2}{0.4}} & \mstd{92.5}{0.5} & \textbf{\mstd{94.4}{0.3}} & \mstd{93.3}{0.5} & \textbf{\mstd{94.2}{0.3}} & \mstd{93.9}{0.2} \\
PCB2        & \mstd{88.8}{0.5} & \textbf{\mstd{89.9}{0.2}} & \mstd{88.9}{0.4} & \textbf{\mstd{90.7}{0.3}} & \mstd{89.5}{0.3} & \textbf{\mstd{91.4}{0.2}} & \mstd{89.6}{0.3} & \textbf{\mstd{92.0}{0.1}} \\
PCB3        & \mstd{84.8}{0.2} & \textbf{\mstd{88.5}{1.3}} & \mstd{86.9}{0.2} & \textbf{\mstd{90.8}{0.5}} & \mstd{89.8}{0.3} & \textbf{\mstd{91.7}{0.4}} & \mstd{88.7}{0.3} & \textbf{\mstd{93.1}{0.3}} \\
PCB4        & \textbf{\mstd{84.1}{0.5}} & \mstd{78.5}{6.8} & \textbf{\mstd{82.9}{0.7}} & \mstd{82.0}{6.1} & \textbf{\mstd{88.9}{0.5}} & \mstd{84.1}{1.5} & \textbf{\mstd{90.0}{0.6}} & \mstd{87.9}{2.3} \\
Pipe fryum  & \mstd{97.7}{0.1} & \textbf{\mstd{98.0}{0.0}} & \mstd{97.8}{0.1} & \textbf{\mstd{97.9}{0.2}} & \textbf{\mstd{97.9}{0.1}} & \mstd{97.8}{0.1} & \textbf{\mstd{97.8}{0.1}} & \mstd{97.6}{0.1} \\
\midrule
Mean        & \textbf{\mstd{93.2}{0.3}} & \mstd{92.5}{0.5} & \textbf{\mstd{93.5}{0.3}} & \mstd{93.4}{0.6} & \textbf{\mstd{94.5}{0.2}} & \mstd{94.1}{0.1} & \mstd{94.6}{0.3} & \textbf{\mstd{94.8}{0.2}} \\
\bottomrule
\end{tabular}%
}
\end{table}

\begin{table}[t!]
\centering
\scriptsize
\setlength{\tabcolsep}{3.2pt}
\renewcommand{\arraystretch}{1.15}
\caption{FS-IVAD object-level detection results for GATE-AD and AnomalyDINO on the MPDD benchmark.}
\label{tab:object_level_mpdd}
\resizebox{\textwidth}{!}{%
\begin{tabular}{l cc cc cc cc}
\toprule
\multirow{2}{*}{Object}
& \multicolumn{2}{c}{1-shot} & \multicolumn{2}{c}{2-shot}
& \multicolumn{2}{c}{4-shot} & \multicolumn{2}{c}{8-shot} \\
\cmidrule(lr){2-3}\cmidrule(lr){4-5}\cmidrule(lr){6-7}\cmidrule(lr){8-9}
& GATE-AD & AnomalyDINO
& GATE-AD & AnomalyDINO
& GATE-AD & AnomalyDINO
& GATE-AD & AnomalyDINO \\
\midrule
\multicolumn{9}{c}{\textbf{Image AUROC (\%)}} \\
\midrule
Bracket black & \textbf{\mstd{83.7}{3.0}} & \mstd{51.0}{6.5} & \textbf{\mstd{83.4}{2.0}} & \mstd{59.4}{8.0} & \textbf{\mstd{80.9}{2.0}} & \mstd{64.2}{8.1} & \textbf{\mstd{85.7}{0.6}} & \mstd{69.5}{2.7} \\
Bracket brown & \mstd{49.7}{1.9} & \textbf{\mstd{52.0}{4.6}} & \mstd{51.7}{1.3} & \textbf{\mstd{56.8}{4.4}} & \mstd{56.1}{2.1} & \textbf{\mstd{58.6}{1.2}} & \mstd{57.0}{2.0} & \textbf{\mstd{61.3}{2.8}} \\
Bracket white & \textbf{\mstd{80.8}{3.3}} & \mstd{70.8}{2.3} & \textbf{\mstd{81.1}{2.2}} & \mstd{72.0}{2.8} & \textbf{\mstd{81.0}{1.9}} & \mstd{65.7}{16.7} & \textbf{\mstd{93.3}{1.7}} & \mstd{72.5}{20.1} \\
Connector    & \textbf{\mstd{93.8}{1.0}} & \mstd{80.6}{4.1} & \textbf{\mstd{91.8}{0.8}} & \mstd{83.1}{3.8} & \textbf{\mstd{89.4}{1.0}} & \mstd{83.7}{1.7} & \textbf{\mstd{93.6}{1.3}} & \mstd{87.9}{3.1} \\
Metal plate  & \mstd{99.3}{0.6} & \textbf{\mstd{100.0}{0.0}} & \mstd{99.7}{0.3} & \textbf{\mstd{100.0}{0.0}} & \mstd{99.8}{0.3} & \textbf{\mstd{100.0}{0.0}} & \mstd{99.5}{0.5} & \textbf{\mstd{100.0}{0.0}} \\
Tubes        & \textbf{\mstd{97.6}{0.8}} & \mstd{89.2}{1.4} & \textbf{\mstd{98.0}{0.6}} & \mstd{88.8}{1.5} & \textbf{\mstd{97.6}{0.7}} & \mstd{91.6}{1.9} & \textbf{\mstd{97.5}{0.7}} & \mstd{91.6}{2.1} \\
\midrule
Mean        & \textbf{\mstd{84.2}{0.9}} & \mstd{73.9}{1.5} & \textbf{\mstd{84.3}{0.5}} & \mstd{76.7}{2.5} & \textbf{\mstd{84.1}{0.9}} & \mstd{77.3}{2.4} & \textbf{\mstd{87.8}{0.5}} & \mstd{80.5}{3.0} \\

\midrule
\multicolumn{9}{c}{\textbf{Pixel PRO (\%)}} \\
\midrule
Bracket black & \textbf{\mstd{97.4}{0.5}} & \mstd{89.1}{0.6} & \textbf{\mstd{97.2}{0.5}} & \mstd{90.3}{0.2} & \textbf{\mstd{97.3}{0.3}} & \mstd{91.5}{1.7} & \textbf{\mstd{98.3}{0.0}} & \mstd{93.4}{0.9} \\
Bracket brown & \textbf{\mstd{85.8}{0.3}} & \mstd{80.6}{1.2} & \textbf{\mstd{85.7}{0.3}} & \mstd{83.7}{1.7} & \textbf{\mstd{86.6}{0.3}} & \mstd{84.9}{1.7} & \textbf{\mstd{86.9}{0.6}} & \mstd{85.3}{2.6} \\
Bracket white & \textbf{\mstd{95.4}{0.2}} & \mstd{95.4}{0.7} & \mstd{95.2}{0.3} & \textbf{\mstd{96.3}{1.6}} & \mstd{95.4}{0.3} & \textbf{\mstd{96.3}{1.6}} & \textbf{\mstd{97.6}{0.2}} & \mstd{95.5}{2.0} \\
Connector    & \textbf{\mstd{90.2}{0.5}} & \mstd{84.4}{2.4} & \textbf{\mstd{89.9}{0.5}} & \mstd{86.2}{2.8} & \textbf{\mstd{90.2}{0.4}} & \mstd{88.0}{2.8} & \textbf{\mstd{91.2}{0.3}} & \mstd{89.0}{0.7} \\
Metal plate  & \textbf{\mstd{92.8}{0.2}} & \mstd{89.0}{0.3} & \textbf{\mstd{93.4}{0.2}} & \mstd{89.4}{0.9} & \textbf{\mstd{93.8}{0.1}} & \mstd{89.7}{0.6} & \textbf{\mstd{93.6}{0.1}} & \mstd{90.5}{0.8} \\
Tubes        & \textbf{\mstd{98.7}{0.2}} & \mstd{86.1}{1.4} & \textbf{\mstd{98.7}{0.2}} & \mstd{87.6}{1.4} & \textbf{\mstd{98.6}{0.2}} & \mstd{88.5}{1.1} & \textbf{\mstd{98.6}{0.2}} & \mstd{88.8}{0.9} \\
\midrule
Mean        & \textbf{\mstd{93.4}{0.2}} & \mstd{87.4}{0.6} & \textbf{\mstd{93.4}{0.1}} & \mstd{88.9}{0.9} & \textbf{\mstd{93.7}{0.1}} & \mstd{89.8}{1.1} & \textbf{\mstd{94.4}{0.0}} & \mstd{90.4}{0.5} \\
\bottomrule
\end{tabular}%
}
\end{table}

\noindent
\underline{VisA (Table \ref{tab:object_level_visa})}:
\begin{itemize}
    \item GATE-AD exhibits high performance for most classes across all settings, e.g., `Capsules' ($99.3\%$ image AUROC in the $1$-shot case) and `Chewing gum' ($98.8\%$ image AUROC in the $1$-shot case), clearly surpassing the respective AnomalyDINO metrics.
    \item For complex electronic components (i.e., printed circuit boards), GATE-AD maintains high detection accuracy: `PCB1' ($98.0\%$ image AUROC in the $1$-shot setting), `PCB2' ($92.0\%$ image AUROC in the $1$-shot case), and `PCB4' ($88.3\%$ image AUROC in the $1$-shot setting).
    \item For food and organic shapes, GATE-AD exhibits increased localization performance, e.g., `Capsules' ($99.3\%$ image AUROC in the $1$-shot setting) and `Chewing gum'($98.8\%$ image AUROC in the $1$-shot case).
\end{itemize}

\noindent
\underline{MPDD (Table \ref{tab:object_level_mpdd})}:
\begin{itemize}
    \item GATE-AD achieves high detection scores even in the $1$-shot setting, accomplishing an $99.3\%$ image AUROC for class `Metal plate' and a corresponding $97.6\%$ for category `Tubes'.
    \item GATE-AD demonstrates consistent localization, achieving $98.7\%$ pixel PRO in the $1$-shot setting for category `Tubes' and a corresponding $97.4\%$ for class `Bracket black'.
    \item Especially for the localization task, GATE-AD demonstrates higher pixel PRO rates for the vast majority of classes across all settings.
\end{itemize}

\begin{figure}[t!]
  \centering
  \includegraphics[width=0.95\textwidth]{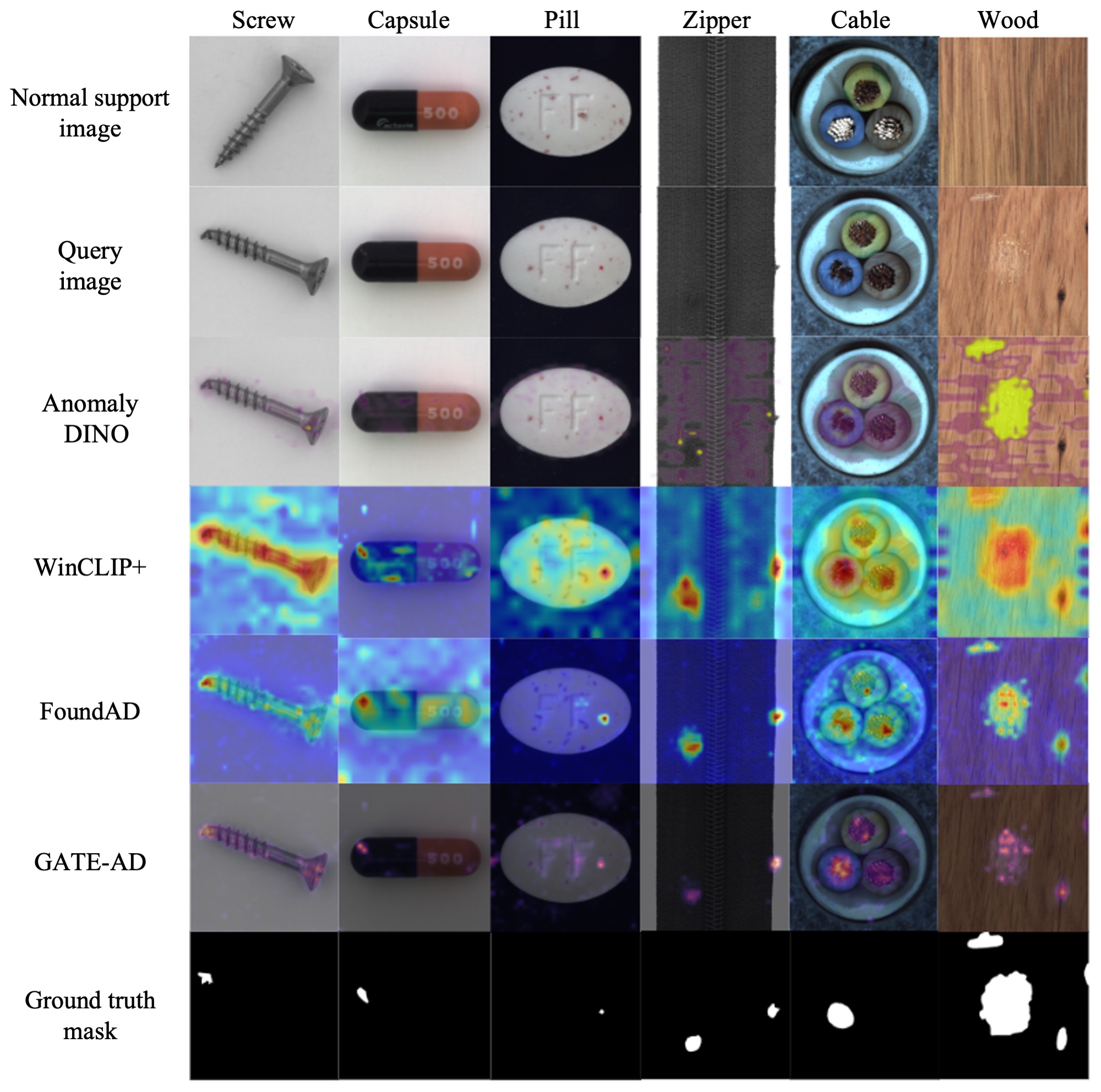}
  \caption{Indicative $1$-shot defect detection results on the MVTec AD benchmark.}
  \label{fig:qualitative_examples_MVTec}
\end{figure}

\begin{figure}[t!]
  \centering
  \includegraphics[width=0.95\textwidth]{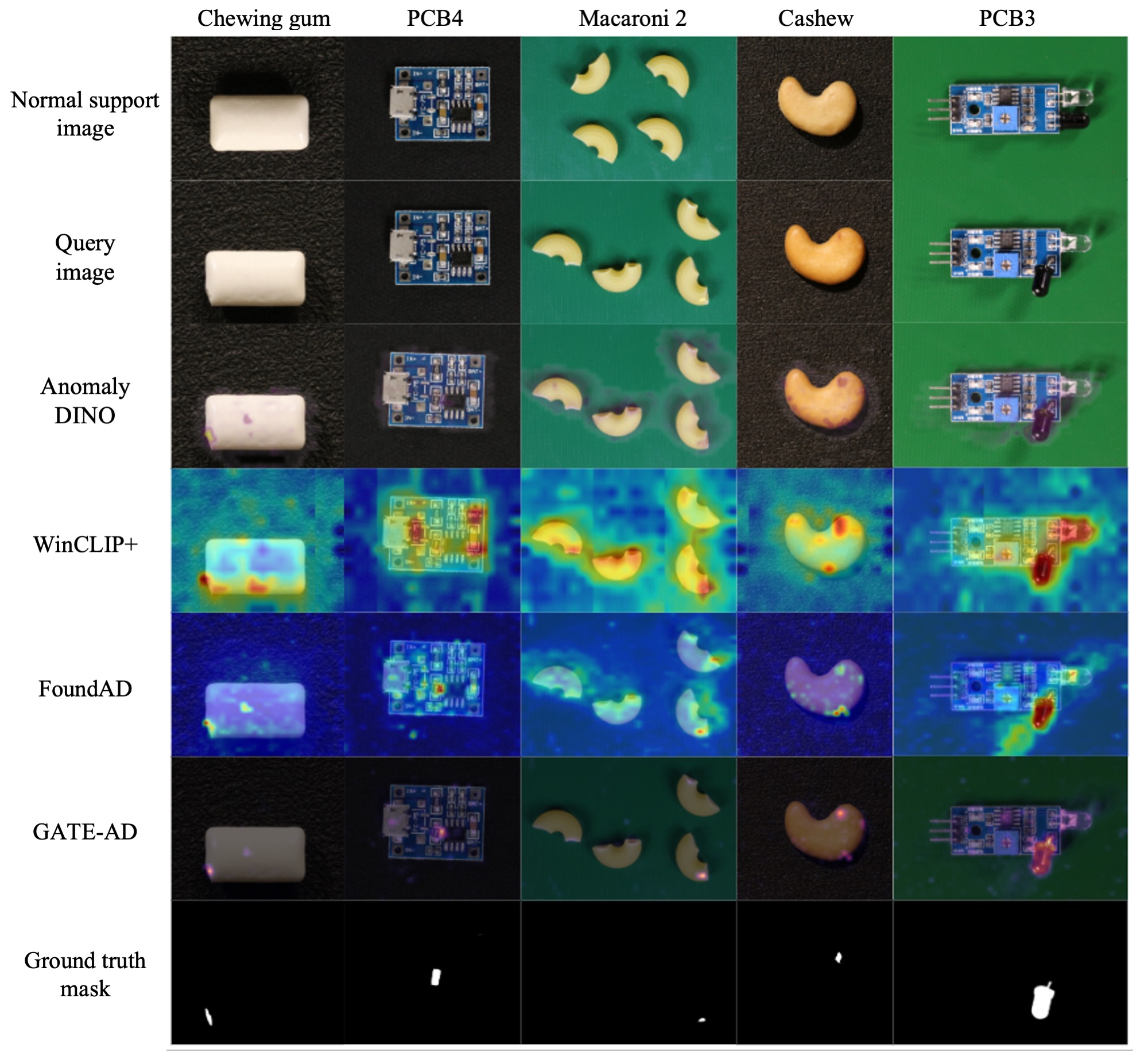}
  \caption{Indicative $1$-shot defect detection results on the VisA benchmark.}
  \label{fig:qualitative_examples_VisA}
\end{figure}

\begin{figure}[t!]
  \centering
  \includegraphics[width=0.95\textwidth]{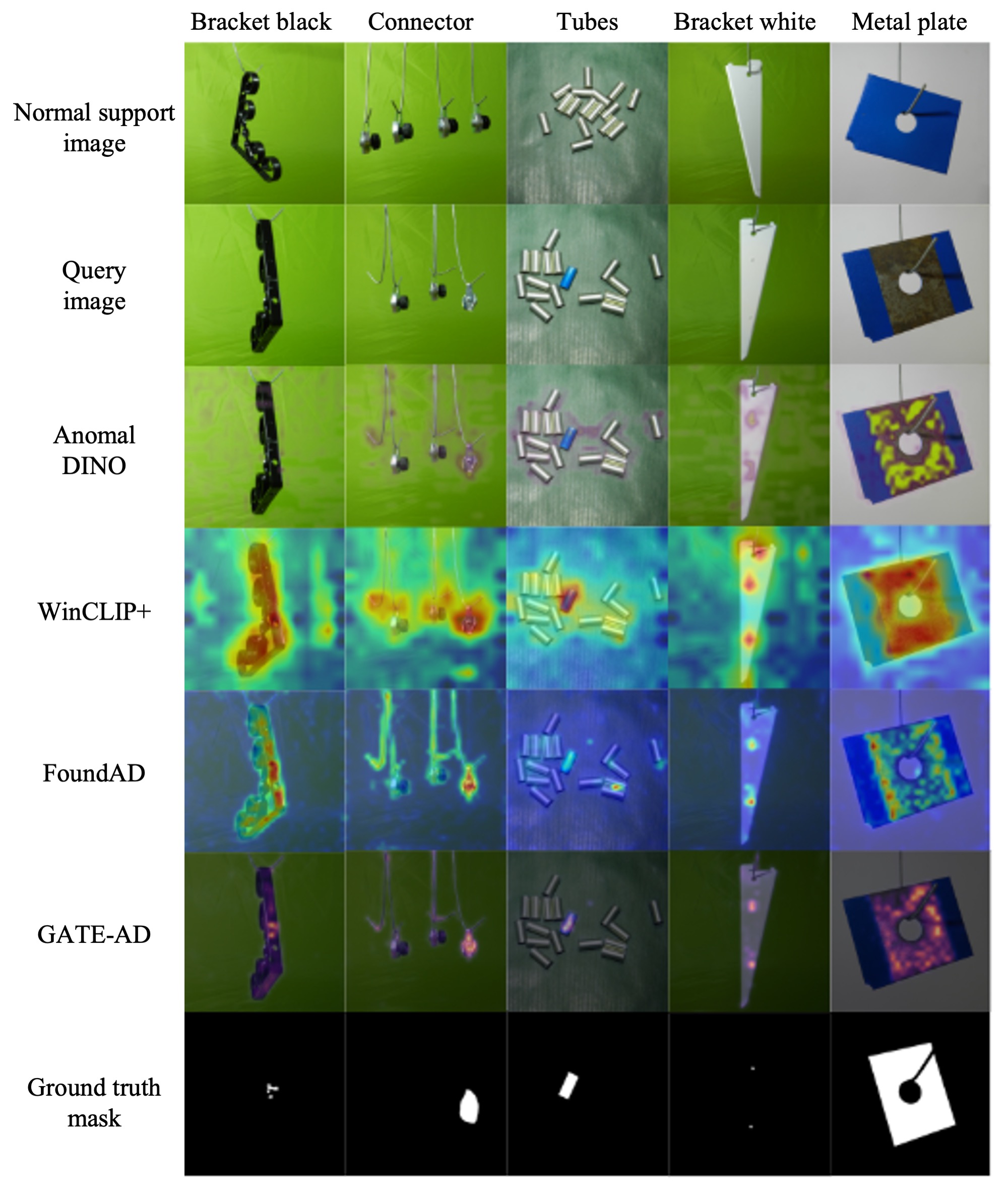}
  \caption{Indicative $1$-shot defect detection results on the MPDD benchmark.}
  \label{fig:qualitative_examples_MPDD}
\end{figure}

\subsection{Product-agnostic defect-type analysis}
\label{sec:defect_type_analysis}

To further examine the performance of GATE-AD across different defect types, a stratified evaluation is conducted on MVTec AD using the product-agnostic defect annotations of Gangopadhyay et al.~\cite{gangopadhyay2023benchmarking}, grouping anomalous test images into surface (local, e.g., scratches and dents), structural (broader-scale, e.g., deformation, holes, damage, and missing parts), contamination, and combined defects. Since each subgroup contains only anomalous images, image-level AUROC is not meaningful; therefore, pixel-level P-AUROC and AUPRO are reported in Table~\ref{tab:defect_type_analysis} across the 1-, 2-, 4-, and 8-shot settings. GATE-AD performs strongly across all considered groups, with its clearest performance on local surface defects; at 8-shot, surface defects achieve $98.3\%$ P-AUROC and $96.0\%$ AUPRO, compared with $96.8\%$ and $91.1\%$, respectively, for broader-scale structural defects, with the difference being most evident in AUPRO and remaining consistent across the considered shot settings. This behavior is consistent with the design of GATE-AD, which operates on a patch-level graph over a fixed local neighborhood and detects anomalies through local reconstruction inconsistency, making the formulation particularly well suited to localized defects. Nevertheless, strong localization performance is also maintained for structural, contamination, and combined defects, while extending the current formulation toward multi-scale or global graph reasoning constitutes a promising direction for broader structural anomalies.

\begin{table}[h]
\centering
\caption{Defect-type localization results on MVTec AD (P-AUROC / AUPRO, \%, mean$\pm$std). ``Mean'' denotes the unweighted average over the four defect groups.}
\label{tab:defect_type_analysis}

\small
\setlength{\tabcolsep}{2.5pt}
\renewcommand{\arraystretch}{1.10}

\newcommand{\std}[1]{{\scriptstyle \pm #1}}

\resizebox{\textwidth}{!}{%
\begin{tabular}{@{}lcccc@{}}
\toprule
\textbf{Defect group} &
\textbf{1-shot} &
\textbf{2-shot} &
\textbf{4-shot} &
\textbf{8-shot} \\
\midrule

Surface (local)
& $97.9\std{0.0}$ / $95.0\std{0.0}$
& $98.1\std{0.0}$ / $95.5\std{0.0}$
& $98.2\std{0.0}$ / $95.8\std{0.0}$
& $98.3\std{0.0}$ / $96.0\std{0.1}$ \\

Structural (broader-scale)
& $96.4\std{0.1}$ / $90.5\std{0.1}$
& $96.7\std{0.0}$ / $90.9\std{0.1}$
& $96.8\std{0.0}$ / $91.0\std{0.1}$
& $96.8\std{0.0}$ / $91.2\std{0.1}$ \\

Contamination
& $97.3\std{0.1}$ / $94.3\std{0.0}$
& $97.6\std{0.1}$ / $94.7\std{0.1}$
& $97.8\std{0.1}$ / $94.8\std{0.1}$
& $97.8\std{0.1}$ / $94.8\std{0.1}$ \\

Combined
& $97.5\std{0.0}$ / $91.2\std{0.2}$
& $97.6\std{0.0}$ / $91.5\std{0.1}$
& $97.5\std{0.0}$ / $91.7\std{0.1}$
& $97.6\std{0.0}$ / $91.9\std{0.1}$ \\

\midrule

\textbf{Mean}
& $\mathbf{97.3}$ / $\mathbf{92.7}$
& $\mathbf{97.5}$ / $\mathbf{93.2}$
& $\mathbf{97.6}$ / $\mathbf{93.3}$
& $\mathbf{97.6}$ / $\mathbf{93.5}$ \\

\bottomrule
\end{tabular}%
}
\end{table}

\subsection{Qualitative evaluation results}
\label{subsec:qualitative}

Figs. \ref{fig:qualitative_examples_MVTec}-\ref{fig:qualitative_examples_MPDD} illustrate indicative $1$-shot defect detection results from the application of GATE-AD, AnomalyDINO \citep{damm2025anomalydino}, WinCLIP+ \citep{jeong2023winclip}, and FoundAD \citep{zhai2025foundation} across the MVTec AD, VisA, and MPDD benchmarks. In order to ensure a fair comparison, all baseline methods are evaluated using their default (open source) configuration, while the same single normal support image is employed. Overall, the presented visualizations reinforce the above quantitative analysis results (Section \ref{subsec:quantitative}), demonstrating that GATE-AD localizes defects more accurately and robustly than other baseline methods, across multiple pose, placement, and abnormality type settings. Key insights on a per dataset basis are provided in the following.

\noindent
\underline{MVTec AD (Fig. \ref{fig:qualitative_examples_MVTec})}:
\begin{itemize}
    \item GATE-AD demonstrates superior localization for both structural defects (e.g., in `Screw' and `Cable') and surface anomalies (e.g., in `Pill' and `Wood').
    \item While baseline methods (e.g., WinCLIP+) often produce diffuse or noisy heatmaps that cover large, non-defective areas, GATE-AD produces focused, high-intensity responses that align closely with the ground truth masks.
    \item GATE-AD effectively distinguishes between natural texture variations and genuine defects in categories such as `Wood'.
\end{itemize}

\noindent
\underline{VisA (Fig. \ref{fig:qualitative_examples_VisA})}:
\begin{itemize}
    \item GATE-AD excels in scenes containing multiple objects or complex electronic layouts, such as `PCB4' and `PCB3'.
    \item For organic or irregular shapes (e.g., `Macaroni 2' and `Cashew'), GATE-AD successfully detects fine-grained anomalies that other reconstruction-based methods (e.g., FoundAD) may fail to distinguish.
    \item GATE-AD-generated anomaly maps exhibit minimal activation in background regions, in sharp contradistinction to most baseline methods.
\end{itemize}

\noindent
\underline{MPDD (Fig. \ref{fig:qualitative_examples_MPDD})}:
\begin{itemize}
    \item GATE-AD demonstrates significant robustness in the case of varied poses and lighting conditions (e.g., `Bracket black' and `Tubes' classes).
    \item GATE-AD is also shown to accurately identify missing components or surface failures (e.g., `Bracket black' and `Connector' categories).
    \item Even when the query image has a different orientation than the single support image, GATE-AD detects local defect occurrences (e.g., `Metal plate' class).
\end{itemize}

\subsection{Query-conditioned representation analysis}
\label{subsec:representation_analysis}

\begin{figure*}[h]
    \centering
    \includegraphics[width=\textwidth]{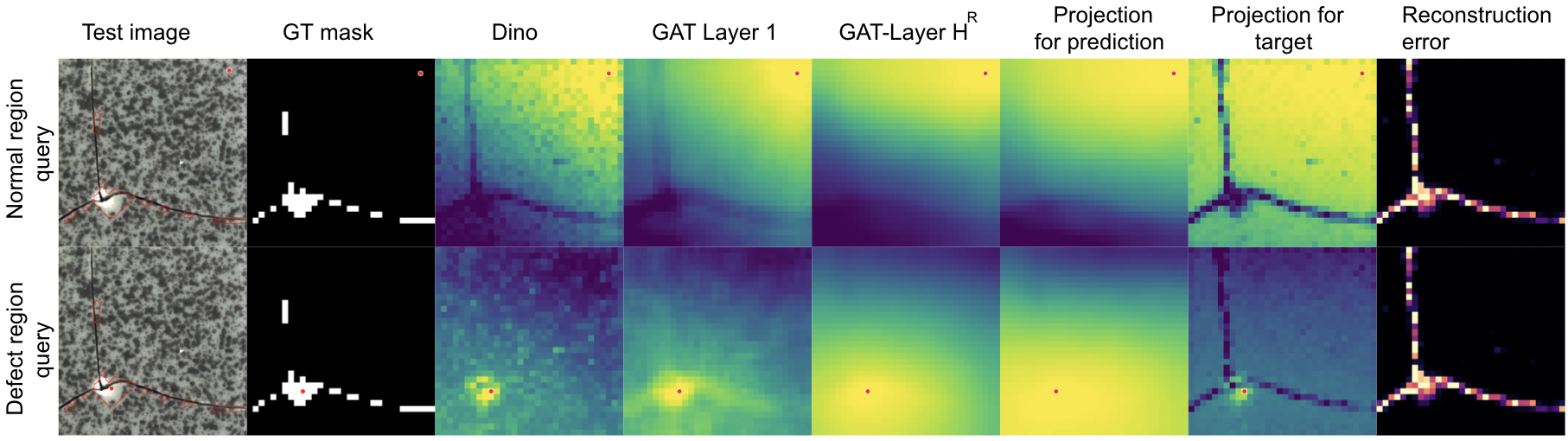}
    \caption{Query-conditioned representation analysis of GATE-AD. The red point denotes a query patch selected from a normal region (top) or from the ground-truth defect (bottom). The heatmaps show query-to-patch similarity in the original DINO representation, after the first and final GAT layers, and after the prediction and target projections, followed by the corresponding reconstruction error. For the normal query, the contextual prediction remains consistent with the observed target, producing a low reconstruction error. For the defective query, graph message passing progressively forms a smoother, context-conditioned prediction from the surrounding spatial context, while the target projection retains and enhances the distinction of the observed defect. Their disagreement produces the localized reconstruction error used by GATE-AD.}
    \label{fig:representation_analysis}
\end{figure*}

To further examine how relational reconstruction contributes to anomaly localization, Fig.~\ref{fig:representation_analysis} traces the representation of a query patch selected from either a normal region or a ground-truth defect through the different stages of GATE-AD. At each stage, the heatmap illustrates the attention between the selected query representation and all other patches in the image. For the normal query, the representation remains consistent with its surrounding context through the GAT layers, and the projected prediction closely agrees with the corresponding target, resulting in a low reconstruction error. In contrast, the defective query is initially distinguishable in the DINO representation, while successive GAT layers progressively form a smoother, context-conditioned representation from the surrounding neighborhood. The prediction projection therefore reflects the locally expected representation, whereas the target projection preserves and further separates the observed defect from normal regions. Their disagreement produces a strong localized reconstruction error at the anomalous patch. This analysis illustrates the complementary roles of the two components: the GAT reconstructs the expected patch representation from its local relational context, while representation alignment provides a more discriminative space in which the context-conditioned prediction is compared with the observed target.

\subsection{Failure cases}
\label{subsec:failure}

\begin{figure}[t!]
\centering
\includegraphics[width=0.95\textwidth]{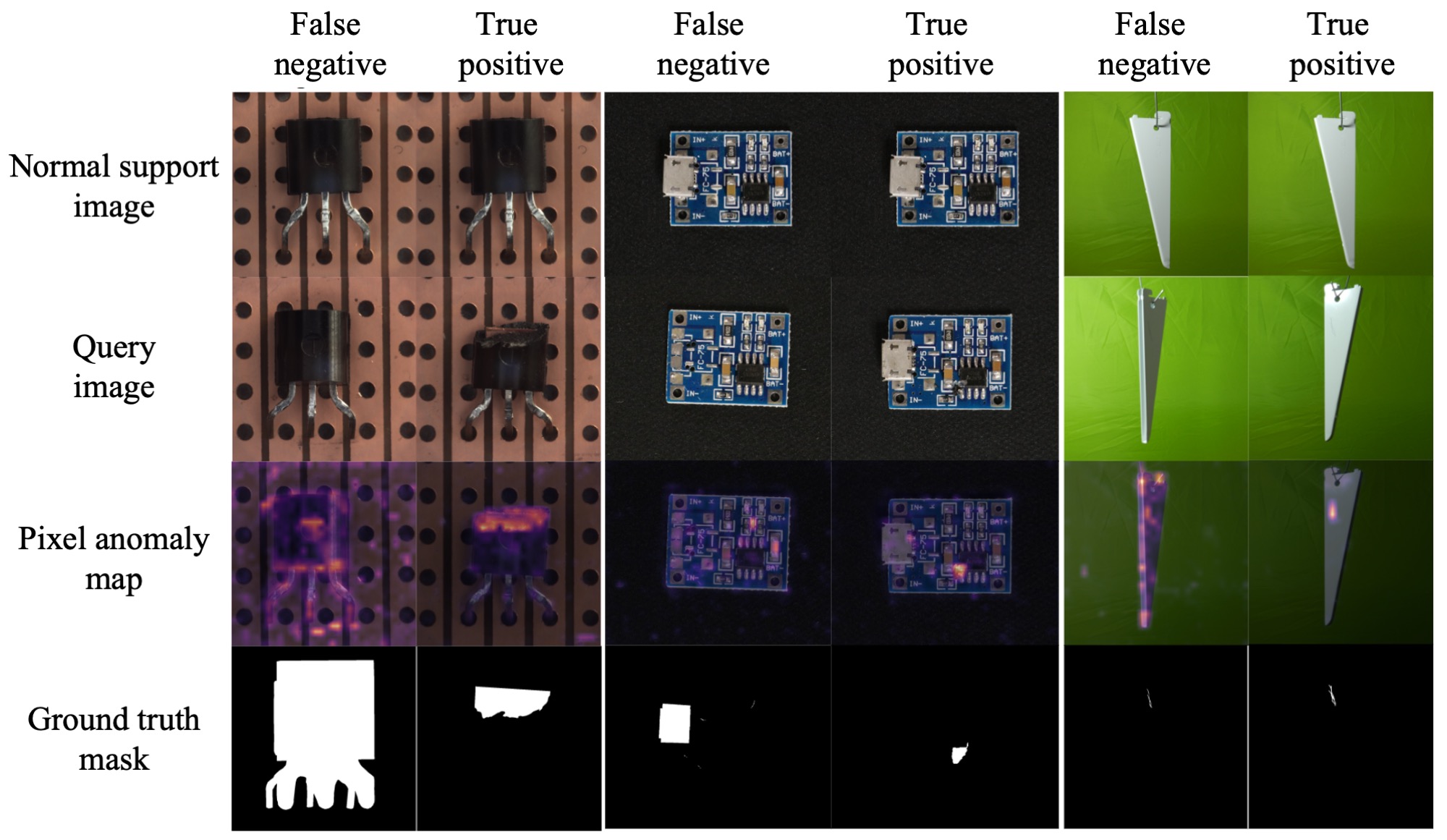}
\caption{Indicative examples of defect detection failure cases across the MVTec AD, VisA, and MPDD benchmarks.}
\label{fig:failure_cases}
\end{figure}

Fig. \ref{fig:failure_cases} illustrates indicative scenarios where GATE-AD may fail to detect or accurately localize anomalies across the MVTec AD, VisA, and MPDD benchmarks. Key insights extracted from the provided results are as follows:
\begin{itemize}
    \item Primary source of failure comprises when anomalies arise from global or semantic changes, rather than local surface defects. This includes instances of rotated or misplaced components.
    \item If a query image contains a pose variation that was not present in the available (small) normal support set, GATE-AD may struggle to distinguish this natural variation from a genuine defect.
    \item Since GATE-AD is grounded on local-level reconstruction of patch features, it may be sensitive to distinct pixel-level mismatches, but less effective in capturing high-level structural anomalies that do not significantly disrupt local feature consistency.
\end{itemize}

\section{Computational analysis}
\label{sec:efficiency}

This section complements the architectural analysis of Section~\ref{ssec:ablation} by examining the computational requirements of the considered relational encoders and reconstruction formulations. End-to-end inference efficiency against existing FS-IVAD methods is reported separately in Table~\ref{tab:runtime_same_gpu}. Table~\ref{tab:architectural_complexity} reports the trainable parameter count, operator-counted FLOPs, and training time of GATE-AD and the matched architectural variants. The reported training time corresponds to the optimization of each object-class model after frozen-backbone feature extraction and is summarized across the evaluated shot settings. GAT and GCN exhibit comparable computational requirements, whereas the global Transformer Encoder (TE) substantially increases the parameter count, FLOPs, and training time. In particular, TE requires approximately an order of magnitude longer training than GAT across all three datasets. Regarding the reconstruction formulation, GATE-AD is also computationally lighter than the explicit autoencoding alternatives, particularly MAE. Together with the accuracy results of Table~\ref{tab:architectural_ablation}, these results support the adopted GAT-based formulation as providing a favorable trade-off between accuracy and efficiency.

\begin{table}[h]
\centering
\small
\setlength{\tabcolsep}{4.5pt}
\renewcommand{\arraystretch}{1.10}

\newcommand{\tstd}[2]{#1\,{\scriptsize $\pm$#2}}

\caption{Computational analysis of the architectural variants.}
\label{tab:architectural_complexity}

\resizebox{\textwidth}{!}{%
\begin{tabular}{lccccc}
\toprule
& \multicolumn{2}{c}{\textbf{Complexity}}
& \multicolumn{3}{c}{\textbf{Training time (s/class)}} \\
\cmidrule(lr){2-3}
\cmidrule(lr){4-6}

\textbf{Method}
& \textbf{Params (M)}
& \textbf{FLOPs (G)}
& \textbf{MVTec AD}
& \textbf{VisA}
& \textbf{MPDD} \\
\midrule

GATE-AD (GAT)
& \textbf{0.99}
& \textbf{177.48}
& \textbf{\tstd{6.7}{0.9}}
& \textbf{\tstd{7.8}{0.9}}
& \tstd{4.8}{0.6} \\

GATE-AD (GCN)
& 1.32
& 178.15
& \tstd{9.0}{3.5}
& \tstd{10.4}{3.4}
& \textbf{\tstd{4.4}{0.3}} \\

Transformer Encoder (TE)
& 22.84
& 232.37
& \tstd{62.9}{16.1}
& \tstd{83.1}{23.4}
& \tstd{46.6}{10.0} \\

GATE-AD AE
& 4.14
& 182.04
& \tstd{16.1}{6.5}
& \tstd{19.0}{6.7}
& \tstd{18.6}{1.3} \\

MAE
& 28.36
& 232.64
& \tstd{52.4}{11.4}
& \tstd{64.5}{6.8}
& \tstd{75.4}{1.5} \\

\bottomrule
\end{tabular}%
}
\end{table}

\newpage
\section*{NeurIPS Paper Checklist}

\begin{enumerate}

\item {\bf Claims}
    \item[] Question: Do the main claims made in the abstract and introduction accurately reflect the paper's contributions and scope?
    \item[] Answer: \answerYes{}.
    \item[] Justification: The abstract and introduction section focus on the actual contribution and scope of the paper, which relates to the introduction of a new method, termed GATE-AD, for Few-Shot Industrial Visual Anomaly Detection (FS-IVAD).

\item {\bf Limitations}
    \item[] Question: Does the paper discuss the limitations of the work performed by the authors?
    \item[] Answer: \answerYes.
    \item[] Justification: Limitations of the current approach are in principle experimentally/empirically assessed, by reporting specific fail cases of the proposed method and discussing their possible root causes (Appendix \ref{subsec:failure}).

\item {\bf Theory assumptions and proofs}
    \item[] Question: For each theoretical result, does the paper provide the full set of assumptions and a complete (and correct) proof?
    \item[] Answer: \answerNA{}.
    \item[] Justification: The paper does not include theoretical results.

    \item {\bf Experimental result reproducibility}
    \item[] Question: Does the paper fully disclose all the information needed to reproduce the main experimental results of the paper to the extent that it affects the main claims and/or conclusions of the paper (regardless of whether the code and data are provided or not)?
    \item[] Answer: \answerYes{}.
    \item[] Justification: Towards the direction of method reproducibility, the method's technical description (Section \ref{sec:method}) and hyper-parameters (Section \ref{ssec:setup}) are described in detail, while extensive ablation studies regarding hyper-parameter tuning (Appendix \ref{sec:hparam} and the method's source code (https://github.com/gthpapadopoulos/GATE-AD) are also provided.
    
\item {\bf Open access to data and code}
    \item[] Question: Does the paper provide open access to the data and code, with sufficient instructions to faithfully reproduce the main experimental results, as described in supplemental material?
    \item[] Answer: \answerYes{}.
    \item[] Justification: The source code of the developed methods as well as detailed execution instructions are provided (https://github.com/gthpapadopoulos/GATE-AD).

\item {\bf Experimental setting/details}
    \item[] Question: Does the paper specify all the training and test details (e.g., data splits, hyperparameters, how they were chosen, type of optimizer) necessary to understand the results?
    \item[] Answer: \answerYes{}.
    \item[] Justification: The paper reports all required implementation details (Section \ref{ssec:setup}).

\item {\bf Experiment statistical significance}
    \item[] Question: Does the paper report error bars suitably and correctly defined or other appropriate information about the statistical significance of the experiments?
    \item[] Answer: \answerYes{}.
    \item[] Justification:  The reported experiments are repeated three times with fixed seeds, and the estimated mean and standard deviation are provided (Section \ref{ssec:comparative_evaluation}, and Appendix \ref{sec:perclass}).

\item {\bf Experiments compute resources}
    \item[] Question: For each experiment, does the paper provide sufficient information on the computer resources (type of compute workers, memory, time of execution) needed to reproduce the experiments?
    \item[] Answer: \answerYes{}.
    \item[] Justification: The utilized compute resources are described in detail (Section \ref{ssec:setup}).

\item {\bf Code of ethics}
    \item[] Question: Does the research conducted in the paper conform, in every respect, with the NeurIPS Code of Ethics \url{https://neurips.cc/public/EthicsGuidelines}?
    \item[] Answer: \answerYes{}.
    \item[] Justification: The research conducted in the paper conforms, in every respect, with the NeurIPS Code of Ethics.

\item {\bf Broader impacts}
    \item[] Question: Does the paper discuss both potential positive societal impacts and negative societal impacts of the work performed?
    \item[] Answer: \answerNo{}.
    \item[] Justification: The paper discusses the results of foundational research on industrial anomaly detection and the authors cannot reasonably estimate any negative societal impact.
    
\item {\bf Safeguards}
    \item[] Question: Does the paper describe safeguards that have been put in place for responsible release of data or models that have a high risk for misuse (e.g., pre-trained language models, image generators, or scraped datasets)?
    \item[] Answer: \answerNo{}.
    \item[] Justification: The developed model does not have a high risk for misuse or dual-use.

\item {\bf Licenses for existing assets}
    \item[] Question: Are the creators or original owners of assets (e.g., code, data, models), used in the paper, properly credited and are the license and terms of use explicitly mentioned and properly respected?
    \item[] Answer: \answerYes{}.
    \item[] Justification: All public code and datasets utilized in this study are appropriately credited (Section \ref{sec:experiments}).

\item {\bf New assets}
    \item[] Question: Are new assets introduced in the paper well documented and is the documentation provided alongside the assets?
    \item[] Answer: \answerYes{}.
    \item[] Justification: Details of the method's code are appropriately provided (supplemental material).

\item {\bf Crowdsourcing and research with human subjects}
    \item[] Question: For crowdsourcing experiments and research with human subjects, does the paper include the full text of instructions given to participants and screenshots, if applicable, as well as details about compensation (if any)? 
    \item[] Answer: \answerNA{}.
    \item[] Justification: The paper does not involve crowdsourcing nor research with human subjects.

\item {\bf Institutional review board (IRB) approvals or equivalent for research with human subjects}
    \item[] Question: Does the paper describe potential risks incurred by study participants, whether such risks were disclosed to the subjects, and whether Institutional Review Board (IRB) approvals (or an equivalent approval/review based on the requirements of your country or institution) were obtained?
    \item[] Answer: \answerNA{}.
    \item[] Justification: The paper does not involve crowdsourcing nor research with human subjects.

\item {\bf Declaration of LLM usage}
    \item[] Question: Does the paper describe the usage of LLMs if it is an important, original, or non-standard component of the core methods in this research? Note that if the LLM is used only for writing, editing, or formatting purposes and does \emph{not} impact the core methodology, scientific rigor, or originality of the research, declaration is not required.
    \item[] Answer: \answerNA{}.
    \item[] Justification: The core method development in this research does not involve LLMs as any important, original, or non-standard components.

\end{enumerate}

\end{document}